\documentclass[preprint,12pt,authoryear]{elsarticle}
\usepackage[colorlinks=true, linkcolor=red, anchorcolor=red, citecolor=blue, urlcolor=red]{hyperref}
\usepackage{algorithmic}
\usepackage{algorithm}
\usepackage{xcolor}
\usepackage{graphicx}
\usepackage{colortbl}
\usepackage{pifont} % For special symbols
\usepackage{multirow}
\usepackage{rotating}

\definecolor{class-1}{RGB}{255, 0, 0}
\definecolor{class-2}{RGB}{0, 255, 0}
\definecolor{class-3}{RGB}{0, 0, 255}
\definecolor{class-4}{RGB}{255, 255, 0}
\definecolor{class-5}{RGB}{255, 0, 255}
\definecolor{class-6}{RGB}{0, 255, 255}
\definecolor{class-7}{RGB}{128, 0, 0}
\definecolor{class-8}{RGB}{0, 128, 0}
\definecolor{class-9}{RGB}{0, 0, 128}
\definecolor{class-10}{RGB}{128, 128, 0}
\definecolor{class-11}{RGB}{128, 0, 128}
\definecolor{class-12}{RGB}{0, 128, 128}
\definecolor{class-13}{RGB}{192, 192, 192}
\definecolor{class-14}{RGB}{64, 64, 64}
\definecolor{class-15}{RGB}{255, 128, 0}
\definecolor{class-16}{RGB}{128, 128, 255}
\usepackage{amssymb}
\usepackage{amsmath}
\usepackage{bm}
\usepackage{lineno}

\journal{}

\begin{document}

\begin{frontmatter}



\title{When Segmentation Meets Hyperspectral Image: \\ New Paradigm for Hyperspectral Image Classification}

\author[label1]{Weilian Zhou}
\author[label1]{Weixuan Xie}
\author[label1]{Sei-ichiro Kamata\corref{cor1}} 
\author[label3]{Man Sing Wong}
\author[label4]{Huiying (Cynthia) Hou}
\author[label5]{Haipeng Wang}

%% Provide the corresponding author footnote
\cortext[cor1]{Corresponding author: Sei-ichiro Kamata (email: \href{kam@waseda.jp}{kam@waseda.jp})}

\affiliation[label1]{organization={Graduate School of Information, Production and Systems, Waseda University},%Department and Organization
            city={Kitakyushu},
            country={Japan}}

\affiliation[label3]{organization={Department of Land Surveying and Geo-Informatics, Hong Kong Polytechnic University},%Department and Organization
            city={Hong Kong},
            country={China}}
\affiliation[label4]{organization={Department of Building Environment and Energy Engineering, Hong Kong Polytechnic University},%Department and Organization
            city={Hong Kong},
            country={China}}
\affiliation[label5]{organization={Key Laboratory of Electromagnetic Waves Information, Fudan University},%Department and Organization
            city={Shanghai},
            country={China}}

%% Abstract
\begin{abstract}
Hyperspectral image (HSI) classification is a cornerstone of remote sensing, enabling precise material and land-cover identification through rich spectral information. While deep learning has driven significant progress in this task, small patch-based classifiers, which account for over 90\% of the progress, face limitations: (1) the small patch (e.g., $7 \times 7$, $9 \times 9$, etc)-based sampling approach considers a limited receptive field, resulting in insufficient spatial structural information critical for object-level identification, and noise-like misclassifications even within uniform regions, (2) undefined optimal patch sizes lead to coarse label predictions, which degrade performance, and (3) a lack of multi-shape awareness around objects. To address these challenges, we draw inspiration from large-scale image segmentation techniques, which excel at handling object boundaries, and —a capability essential for semantic labeling in HSI classification. However, their application remains under-explored in this task due to (1) the prevailing notion that larger patch sizes degrade performance, (2) the extensive unlabeled regions in HSI ground-truth, and (3) the misalignment of input shapes between HSI data and segmentation models. Thus, in this study, we propose a novel paradigm and baseline, \textbf{HSIseg}, for HSI classification that leverages segmentation techniques combined with a novel Dynamic Shifted Regional Transformer (DSRT) to overcome these challenges. We also introduce an intuitive progressive learning framework with adaptive pseudo-labeling to iteratively incorporate unlabeled regions into the training process, thereby advancing the application of segmentation techniques. Additionally, we incorporate auxiliary data through multi-source data collaboration, promoting better feature interaction. Validated on five public HSI datasets, our proposal outperforms state-of-the-art methods, demonstrating the feasibility of segmentation techniques for HSI classification and breaking through small patch size limitations. This study sets a new benchmark and opens a promising research direction. Code will be available at \url{https://github.com/zhouweilian1904/HSI_Segmentation}.

\end{abstract}

% \includegraphics[scale=0.405]{graphical_abstract.jpg}

% \begin{highlights}
% \item New Paradigm and Baseline for Hyperspectral Image Classification.
% \item Extensible Single-Branch and Multi-Branch Framework.
% \item Dynamic Shifted Regional Transformer.
% \item Progressive Learning with Adaptive Pseudo Labeling.
% \item Break through the Small Patch Size Limitation. 
% \item Satisfying Performance on Five Datasets with Tiny Samples.
% \end{highlights}

%% Keywords

\end{frontmatter}

% \linenumbers


\section{Introduction}
\label{introduction}
Hyperspectral imaging (HSI) is an advanced remote sensing (RS) technology that captures images across numerous densely sampled spectral bands, covering the electromagnetic spectrum from visible to infrared wavelengths. Typically deployed via satellites or unmanned aerial vehicles (UAVs) \cite{c1}, HSI provides rich spectral information for each pixel, enabling precise characterization of materials and land-cover types on the Earth's surface \cite{c2}. This unique capability has led to a wide range of applications, including agriculture \cite{agriculture}, environmental monitoring \cite{environment}, geology \cite{geology}, urban planning \cite{urban_planning}, and defense and security \cite{defense}. Among these applications, HSI classification, which assigns semantic labels to individual pixels, is a critical task. Achieving high accuracy and robustness in HSI classification is essential for improving decision-making in RS applications and remains an active research focus \cite{c7}.

\begin{figure}[t]%% placement specifier
\centering%% For centre alignment of image.
\includegraphics[scale=0.4]{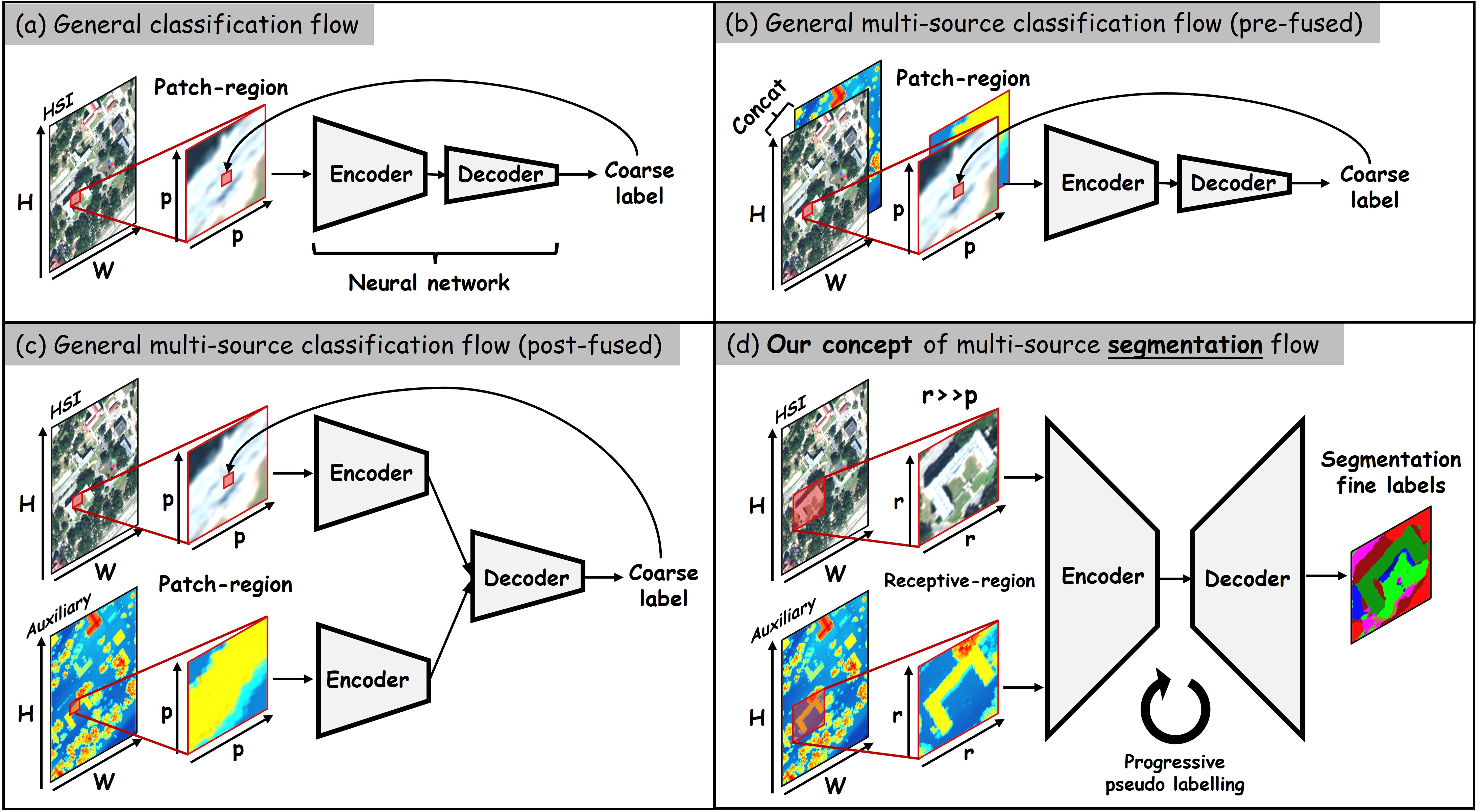}
\caption{Conceptual comparison. (a) General HSI Classification Flow: A small patch region ($p \times p$) is input into a down-sampling-driven encoder-decoder network, which predicts the coarse label of the patch’s central pixel. (b) Multi-Source HSI Classification with Early Fusion: Two data sources are fused before entering the neural network, integrating complementary spectral-spatial information at the input level. (c) Multi-Source HSI Classification with Late Fusion: Two data sources are processed separately in the encoder and then fused at the end of the encoding stage, before classification. (d) Our Multi-Source HSI Segmentation Flow: A much larger receptive field ($r \times r$) is used for both data sources, which are processed through a down-sampling-driven encoder and an up-sampling-driven decoder. Unlike conventional fusion strategies, feature interaction occurs dynamically at both the encoder and decoder stages. Furthermore, our segmentation framework integrates a progressive pseudo-labeling strategy, allowing the model to iteratively refine its predictions and achieve better classification performance.}\label{concept}
\end{figure}

In recent years, deep learning models such as Convolutional Neural Networks (CNNs) \cite{cnn_sota1}, Recurrent Neural Networks (RNNs) \cite{rnn_sota1}, and Transformers \cite{vit0} have been widely adopted in various domains, including HSI classification. In this task, over 90\% of existing deep learning-based approaches employ a patch-wise learning framework, where small patches (e.g., $7 \times 7, 9 \times 9, 11 \times 11$) centered around each HSI pixel serve as input to neural networks as illustrated in Fig.~\ref{concept}(a). While this technique integrates both spectral and spatial features and achieves remarkable results compared to traditional handcrafted methods, it still suffers from several fundamental issues, as described below.

First, their small patch-based sampling strategy has several limitations: (1) Limited Receptive Field: Small patch sizes fail to capture broader spatial structures, leading to incomplete pattern recognition of objects and misclassification near the boundaries of different land-cover types. Additionally, this limitation can introduce noise-like misclassifications even within uniform and smooth land-cover regions, where consistent labeling is expected \cite{hsi_seg_survey}. (2) Uncertainty in Patch Size Selection: The arbitrary choice of patch size introduces a trade-off \cite{patch_decision}. Small patches may exclude essential contextual information, while larger patches may include irrelevant pixels, degrading classification accuracy and causing over-smoothing of object boundaries. (3) Coarse Label Predictions: Patch-wise classifiers often produce coarse labels for the center pixels of cropped patches, failing to preserve fine object details \cite{seg_hsi}, particularly for irregularly shaped land-cover structures. These limitations highlight the need for alternative approaches that leverage larger spatial contexts while maintaining robustness.

Intuitively, large-scale image segmentation techniques, widely used in medical imaging \cite{seg_medical} and autonomous driving \cite{seg_autodriving}, offer a promising solution to the small patch-size limitations in HSI classification. Unlike patch-wise classifiers, segmentation techniques process larger regions, enhancing contextual and structural understanding while improving object boundary delineation. Additionally, they are good at maintaining consistency in uniform and homogeneous regions by promoting spatial coherence and reducing isolated prediction errors. Extending segmentation techniques to HSI could address the small receptive field problem and reduce interference from small patch-wise learning. However, the direct application of segmentation models to HSI remains largely unexplored due to several challenges: (1) High Dimensionality: HSIs contain hundreds of spectral bands, making segmentation computationally expensive; (2) Limited Annotations: Unlike typical segmentation datasets, HSI ground-truth maps contain extensive unlabeled regions, restricting direct and effective pixel-wise training with segmentation techniques \cite{limited_sample}; (3) Assumptions Regarding Patch Limitations: Many studies assume that larger patches degrade classification performance, discouraging segmentation-based exploration; and (4) Input Shape Misalignment: There are differing input shape requirements between HSI classifiers and segmentation models. These challenges necessitate novel segmentation-driven HSI classification frameworks, bridging the gap between segmentation techniques and HSI classification.

Moreover, to improve classification accuracy, some studies have explored multi-source data fusion, integrating LiDAR, SAR, and MSI with HSI to exploit complementary information. Auxiliary modalities provide additional structural and contextual details, enhancing model robustness in challenging classification tasks \cite{multisource1}. However, existing fusion methods face several limitations: (1) Limited Feature Exchange in Fusion Strategies: Most studies adopt either early fusion (e.g., Fig.~\ref{concept}(b)), where auxiliary data is combined with HSI before encoding, or late fusion (e.g., Fig.~\ref{concept}(c)), where features are merged after separate encoders process each modality. Both approaches lack continuous interactive feature exchange, reducing the synergy between modalities; (2) Inadequate Integration in Decoding Stages: Many fusion models (e.g., \cite{mft}\cite{deepsft}, etc.) do not incorporate multi-source feature refinement in the decoder (e.g., relying on simple summation before the fully connected layer), missing opportunities to enhance multi-modal representations during segmentation; and (3) Reliance on Small Patch-Based Learning: Existing multi-source fusion techniques often utilize small patch sizes, inheriting the same spatial limitations as conventional patch-wise HSI classifiers. Therefore, an effective multi-source fusion framework should facilitate modality collaboration across both encoding and decoding stages, particularly within segmentation-based approaches.

Finally, HSI classification is further complicated by the scarcity of labeled data, as pixel-wise annotations require significant domain expertise and field verification. Many HSI datasets contain a large proportion of unlabeled pixels, which are typically ignored during training \cite{unlabel}. While pseudo-labeling techniques offer a potential solution by leveraging unlabeled regions, existing methods exhibit several key limitations: (1) Static Pseudo-Label Assignments: Once pseudo-labels are generated, they remain fixed throughout training, failing to adapt dynamically as the model learns and evolves \cite{seg_hsi}; (2) Fixed Thresholding for Label Confidence: Many approaches rely on predefined confidence thresholds to determine pseudo-label reliability, which may not adequately capture fluctuating model confidence across different regions; and (3) Uniform Treatment of Unlabeled Pixels: Existing methods treat all unlabeled pixels equally, failing to selectively incorporate the most informative samples into the training process. These limitations may hinder the model’s ability to fully leverage unlabeled information and achieve robust classification performance.

In response to four aforementioned scenarios, we propose a new HSI classification paradigm by integrating image segmentation techniques with multi-source data collaboration. Our framework is modular and extensible, supporting both single-branch and multi-branch configurations. To handle extensive unlabeled regions, we propose a progressive learning framework with adaptive pseudo-labeling, iteratively incorporating high-confidence pseudo-labeled regions to enhance model training.

To achieve this paradigm, we first design HSIseg, a single-branch segmentation model utilizing only HSI data in a U-shaped architecture. To refine its performance, we introduce a Discriminative Feature Selection (DFS) module inside, which selects high-confidence features and improves the up-sampling process. Additionally, we propose a Dynamic Shifted Regional Transformer (DSRT), which integrates CNNs for local-global feature aggregation through query-driven multi-head and multi-region self-attention. To enhance feature interaction, we introduce the symmetric Cross Feature Interaction (CFI) module, facilitating effective fusion between high- and low-resolution features. Additionally, for multi-source data collaboration, we extend HSIseg into a multi-branch architecture, where CFI enables cross-modal feature fusion. The HSI branch focuses on classification, while an auxiliary branch supports reconstruction, promoting better multi-source integration. Furthermore, we introduce a progressive pseudo-labeling framework, allowing the trained model to iteratively expand labeled regions, leveraging both labeled and high-confidence unlabeled pixels to improve performance. The conceptual framework of this study is illustrated in Fig.~\ref{concept}(d).

Extensive experiments on five multi-source datasets confirm our framework’s effectiveness, achieving state-of-the-art classification accuracy. This study not only validates the feasibility of segmentation-based HSI classification but also overcomes the limitations of small patch-wise learning, establishing a new benchmark for future research.

The key contributions of this study are summarized as follows:
\begin{itemize} 
\item We propose a novel paradigm for HSI classification based on segmentation techniques, bridging the gap between traditional classification and segmentation methodologies. 
\item We design a modular and extensible segmentation framework for HSI classification, capable of supporting both single-source and multi-source data integration. 
\item We develop a progressive pseudo-labeling framework to address the challenge of limited labeled data, which iteratively incorporates high-confidence unlabeled regions to enhance segmentation performance. 
\item To the best of our knowledge, this is the first study to systematically investigate the impact of segmentation models on HSI classification, providing new insights into their effectiveness for this task. 
\item We conduct extensive experiments on five multi-source datasets, demonstrating the superior performance of our approach and establishing a new benchmark for HSI classification. 
\end{itemize}

The remainder of this paper is organized as follows: Section \ref{related_work} reviews related literature and outlines the key challenges in HSI classification. Section \ref{methodology} describes the proposed segmentation framework and the progressive pseudo-labeling strategy in detail. Section \ref{experiment} presents the experimental results on five multi-source datasets, accompanied by ablation studies. Section \ref{discussion} discusses the limitations of the current study and potential directions for future research. Finally, Section \ref{conclusion} summarizes the main findings and highlights avenues for future work.

\section{Related Work and Problem Statement}\label{related_work}
This section explores four key areas related to the scenarios identified in the Introduction, providing a foundation for the research objectives addressed in this study.
\subsection{Patch-Wise HSI Classification}
HSI classification assigns semantic labels to pixels, but pixel-wise classification is prone to noise, inter-class similarities, and intra-class variabilities. To mitigate these challenges, patch-wise learning frameworks have been widely adopted, incorporating spatial context by using small local patches around each pixel. This spectral-spatial approach has consistently outperformed purely spectral-only or spatial-only methods.

With recent deep learning advancements, patch-wise HSI classification techniques have achieved remarkable results. Early approaches used deep 2D-CNNs \cite{deep-2d-cnn}, later evolving into hybrid 3D-2D CNNs for improved spectral-spatial feature extraction \cite{3d-2d-cnn1}. Attention-enhanced 3D-CNNs further boosted accuracy by emphasizing salient features \cite{attention-cnn}. RNNs were introduced for spectral-spatial modeling, with some methods separately processing spectral and spatial dependencies before fusing them \cite{rnn1}, while others combined CNNs and RNNs for cascaded feature integration \cite{rnn2} or employed multi-directional RNNs for refined representation learning \cite{rnn3}. More recently, ViTs have emerged as state-of-the-art solutions, with approaches like SpectralFormer, which refines Transformer encoders through band-grouping strategies \cite{vit1}, and tokenized ViTs, which integrate 3D-2D CNNs \cite{vit2}. Other architectures include hierarchical Transformers for joint spectral-spatial feature extraction \cite{vit3} and multi-scale attention mechanisms that enhance spatial pattern recognition \cite{vit4}. Additionally, some methods incorporate multi-directional positional embeddings from RNNs to enable purely sequential spectral-spatial processing \cite{vit5}.

Despite these advancements, the reliance on small patch sizes inherently limits the receptive field, failing to capture large-scale spatial structures. This limitation is particularly problematic for complex land-cover regions and boundary areas. While increasing the patch size can provide richer contextual information, it also introduces interfering pixels, leading to feature blending and degraded classification accuracy \cite{seg_hsi}. Additionally, small patch-based methods often suffer from noise-like misclassifications in uniform and homogeneous regions, where inconsistent predictions create isolated errors that disrupt spatial coherence. This issue resembles the "salt-and-pepper" effect, undermining classification reliability in otherwise smooth areas. Furthermore, down-sampling-based decoders struggle to preserve fine details, making optimal patch selection a persistent challenge. Overcoming these limitations and striking the right balance between patch size and contextual coverage remains a promising research direction in this field.

\subsection{Image Segmentation with U-Net}
Image segmentation, which assigns semantic labels to each pixel, plays a crucial role in fields such as medical imaging, RS, autonomous driving, and robotics. Deep learning has significantly advanced segmentation accuracy, with U-Net \cite{unet} emerging as one of the most influential architectures. Originally designed for biomedical tasks, U-Net’s symmetric encoder-decoder structure and skip connections enable the fusion of deep semantic features with fine-grained spatial details, leading to precise pixel-level classification. Segmentation models operate on larger spatial regions, capturing broader context and improving boundary delineation, establishing meaningful spatial relationships.

Building on U-Net’s success, various enhancements have been developed. U-Net++ \cite{unet++} improves skip connections with dense blocks for enhanced feature fusion, while 3D U-Nets \cite{3dunet} handle volumetric data. Attention U-Nets \cite{attention-unet} integrate attention mechanisms for complex scene understanding. More recently, Transformer-based U-Nets have been introduced to capture long-range dependencies. TransUnet \cite{transunet} incorporates a Transformer encoder with a U-shaped decoder for improved global-local feature modeling, while SwinUnet \cite{swinunet} utilizes Swin Transformers \cite{swintrans} to enhance multi-scale feature representation through hierarchical self-attention.

In RS field, segmentation models have been widely applied to large-scale optical aerial and satellite imagery for land-cover mapping. Approaches such as CNN-Transformer hybrid U-Nets \cite{swinunet-rs}, multi-modal fusion networks combining CNN and ViT features \cite{transunet-rs}, and multi-scale convolutional attention with cross-shaped window Transformers \cite{mcatunet} have achieved satisfying results in large-scale RS segmentation.

Applying segmentation frameworks directly to HSI classification presents several challenges. Segmentation models typically assume fully labeled datasets, while HSI ground-truths often contain extensive unlabeled regions with limited annotations, which can mislead model training. Additionally, the traditional patch-wise learning framework struggles to support larger region cropping due to the limited spatial dimensions of HSI data. As a result, most current HSI studies rely on patch-wise or pixel-wise classification, leaving large-scale segmentation largely under-explored. To address these limitations, our study aims to introduce a segmentation-based paradigm for HSI classification, moving beyond traditional small patch-wise approaches to enable larger context-aware feature learning. Given the absence of prior work applying segmentation techniques to HSI classification, we carefully select the well-known SwinUnet as our research benchmark , which we aim to modify and improve to achieve superior results.

\subsection{Multi-Source Data Collaboration}
To enhance HSI classification, numerous studies have incorporated auxiliary data sources such as LiDAR, SAR, and MSI, leveraging their complementary properties to generate more discriminative features. For instance, LiDAR-derived digital surface models (DSM) provide valuable elevation information, improving semantic differentiation in land-cover analysis.

Recent efforts have explored various multi-source fusion techniques. HSI-MSI fusion has been improved using depthwise feature interaction networks with CNNs  \cite{hsi-msi}. HSI-MSI-SAR integration has been addressed through shared and specific feature learning models \cite{hsi-msi-sar}. Meanwhile, symmetric fusion Transformers with local-global mixture modules have been employed for HSI-LiDAR fusion, promoting more robust multi-source learning \cite{hsi-lidar}. Other advanced strategies include masked auto-encoders for multi-source data reconstruction prior to classification \cite{ssmae} and HSI-X networks, such as multistage information-complementary fusion models using flexible mixup \cite{hsi-x-1} and local-to-global cross-modal attention-aware architectures for effective feature alignment across modalities \cite{hsi-x-2}. While these approaches have demonstrated effectiveness, several critical issues remain.

Existing fusion methods typically adopt early fusion (Fig. \ref{concept}(b)), where auxiliary data is combined with HSI before encoding, or late fusion (Fig. \ref{concept}(c)), where features are merged after separate encoders process each modality. Both strategies lack interactive feature exchange between modalities within the encoder and decoder, and they primarily rely on small patch-based learning frameworks. Early fusion risks suboptimal representations, as it fails to preserve modality-specific characteristics (e.g., SAR data affected by speckle noise may degrade fusion quality). Late fusion, in contrast, treats each modality independently, limiting its ability to capture complementary relationships. Moreover, most approaches employ simple down-sampling-based fully connected layers as decoders, which further restricts multi-source feature refinement. Thus, an effective fusion framework should enable modality collaboration across both encoding and decoding stages, particularly within models utilizing larger receptive fields under segmentation techniques. However, this remains an under-explored research direction.

\subsection{Semi-Supervised Learning with Pseudo Labeling}
Acquiring pixel-level annotations for HSI is both costly and time consuming, requiring domain expertise and extensive ground verification. As a result, HSI datasets typically contain a significant number of unlabeled pixels, which are often assigned a generic background label (e.g., 0). Traditional HSI classification methods train models exclusively on labeled regions, evaluating performance on the remaining labeled pixels while ignoring the distribution of unlabeled regions. While this approach may achieve acceptable accuracy, it fails to utilize valuable spectral and spatial information hidden within the unlabeled data, limiting the model’s generalization capability.

Semi-supervised learning (SSL) and pseudo-labeling have emerged as effective strategies to incorporate unlabeled data into training. Some approaches generate pseudo-labels through cluster assignments, leveraging large amounts of unlabeled data to refine feature learning \cite{pseudo-4}. Other methods pre-train deep CNN-RNN models on pseudo-labeled samples before fine-tuning them on limited labeled data \cite{pseudo-3}. Probabilistic frameworks have also been explored, where pseudo-labels are assigned based on Gaussian distributions to maximize feature consistency \cite{pseudo-1}. More advanced techniques focus on reducing pseudo-label noise by filtering out low-confidence samples. Some methods distinguish between reliable and unreliable pseudo-labels, incorporating only high-confidence samples into the training process \cite{pseudo-5}. Others iteratively refine pseudo-labels through multi-scale super-pixel segmentation and spectral-spatial distance analysis \cite{pseudo-2}. Additionally, uncertainty-aware selection strategies based on Bayesian networks have been introduced to iteratively update pseudo-labels while minimizing misclassification risks \cite{pseudo-6}.

Despite these advancements, existing HSI pseudo-labeling strategies face several limitations. Most approaches rely on static pseudo-label assignments, which remain fixed throughout training and fail to adapt as the model evolves. Additionally, fixed thresholding criteria are commonly used to determine pseudo-label reliability, yet these thresholds may not accurately reflect confidence variations or data complexity. Lastly, most methods treat all unlabeled pixels equally, overlooking the potential benefits of selectively incorporating only the most informative samples. These constraints limit the model’s ability to fully leverage unlabeled data, necessitating more adaptive and dynamic pseudo-labeling frameworks for HSI classification.

\section{Methodology\label{methodology}}
\subsection{Preview of Whole Methodology}
We first introduce the data preparation process used in this study in \ref{data prep}. Next, we present the single-branch segmentation model, HSIseg, in \ref{HSIseg}, detailing the design of its encoder, decoder, CFI module, DFS module, and DSRT mechanism. In \ref{multibranch}, we extend HSIseg into a multi-branch architecture to enable multi-modality data collaboration. Finally,  \ref{progressive learning} describes the progressive learning framework with pseudo-labeling, which further refines the model through iterative fine-tuning.

\subsection{Data preparation\label{data prep}}
The original HSI is denoted as:
\begin{equation}
\mathbf{X}_{hsi} \in \mathbb{R}^{\mathcal{H} \times \mathcal{W} \times \mathcal{C}_0},
\end{equation}
and can be defined as:
\begin{equation}
\label{eq:X}
\mathbf{X}_{hsi} = \Big\{\mathbf{x}_{hsi}^{(i,j)} \,\Big|\, i \in [0, \mathcal{H}-1],\, j \in [0, \mathcal{W}-1] \Big\},
\end{equation}
where \(\mathbf{x}_{hsi}^{(i,j)} \in \mathbb{R}^{\mathcal{C}_0}\) represents the spectral signature at spatial position \((i, j)\). Consequently, the full HSI representation is given by:
\begin{equation}
\mathbf{X}_{hsi} =
\begin{pmatrix}
\mathbf{x}_{hsi}^{(0,0)} & \mathbf{x}_{hsi}^{(0,1)} & \dots & \mathbf{x}_{hsi}^{(0,\mathcal{W}-1)} \\
\mathbf{x}_{hsi}^{(1,0)} & \mathbf{x}_{hsi}^{(1,1)} & \dots & \mathbf{x}_{hsi}^{(1,\mathcal{W}-1)} \\
\vdots & \vdots & \ddots & \vdots \\
\mathbf{x}_{hsi}^{(\mathcal{H}-1,0)} & \mathbf{x}_{hsi}^{(\mathcal{H}-1,1)} & \dots & \mathbf{x}_{hsi}^{(\mathcal{H}-1,\mathcal{W}-1)}
\end{pmatrix}.
\end{equation}

Since the proposed framework is patch-based, each input to the neural model is a local region cropped from the original HSI. For a patch \(\mathbf{X}_{hsi}^{(i,j)}\) centered at \(\mathbf{x}^{(i,j)}\) with patch size \(p\), we define:

\begin{equation}
    \mathbf{X}_{hsi}^{(i,j)} = \Big\{\mathbf{x}_{}^{(i+\alpha, j+\beta)} \,\Big|\,
    \alpha, \beta \in \mathbb{Z},\, \alpha, \beta = -\tfrac{p-1}{2}, \ldots, \tfrac{p-1}{2} \Big\},
\end{equation}
where \(\alpha, \beta\) represent the relative spatial offsets within the patch, and \((i+\alpha, j+\beta)\) denote the corresponding spatial coordinates in the original HSI. 

Thus, in the HSI classification task, the data samples consist of a collection of such patches, where a subset is used for training, while the remaining patches are reserved for testing. The original ground truth \(\mathbf{Y}_{hsi} \in \mathbb{R}^{\mathcal{H} \times \mathcal{W}}\) undergoes the same cropping process to generate corresponding patch labels \(\mathbf{Y}_{hsi}^{(i,j)}\).

Before feeding each patch \(\mathbf{X}_{hsi}^{(i,j)}\) into the neural network, we apply Principal Component Analysis (PCA) for dimensionality reduction:
\begin{equation}
    \mathbf{X}_{hsi}^{(i,j)} \leftarrow PCA\bigl(\mathbf{X}_{hsi}^{(i,j)}\bigr),
\end{equation}
retaining the top \(\mathcal{C}_1\) principal components. As a result, the transformed patch \(\mathbf{X}_{hsi}^{(i,j)} \in \mathbb{R}^{p \times p \times \mathcal{C}_1}\) is ready for subsequent processing.

Auxiliary datasets, such as LiDAR (\(\mathbf{X}_{lidar}\)), SAR (\(\mathbf{X}_{sar}\)), and MSI (\(\mathbf{X}_{msi}\)), undergo a similar patching process using the same spatial indices, yielding \(\mathbf{X}_{lidar}^{(i,j)}\), \(\mathbf{X}_{sar}^{(i,j)}\), and \(\mathbf{X}_{msi}^{(i,j)}\), respectively. Any available multi-source data can be incorporated into the framework for collaborative learning.

\subsection{Single-Branch Segmentation Model (Main Branch-HSI Branch)\label{HSIseg}}
\begin{figure}[!t]
    \centering
    \includegraphics[scale=0.385]{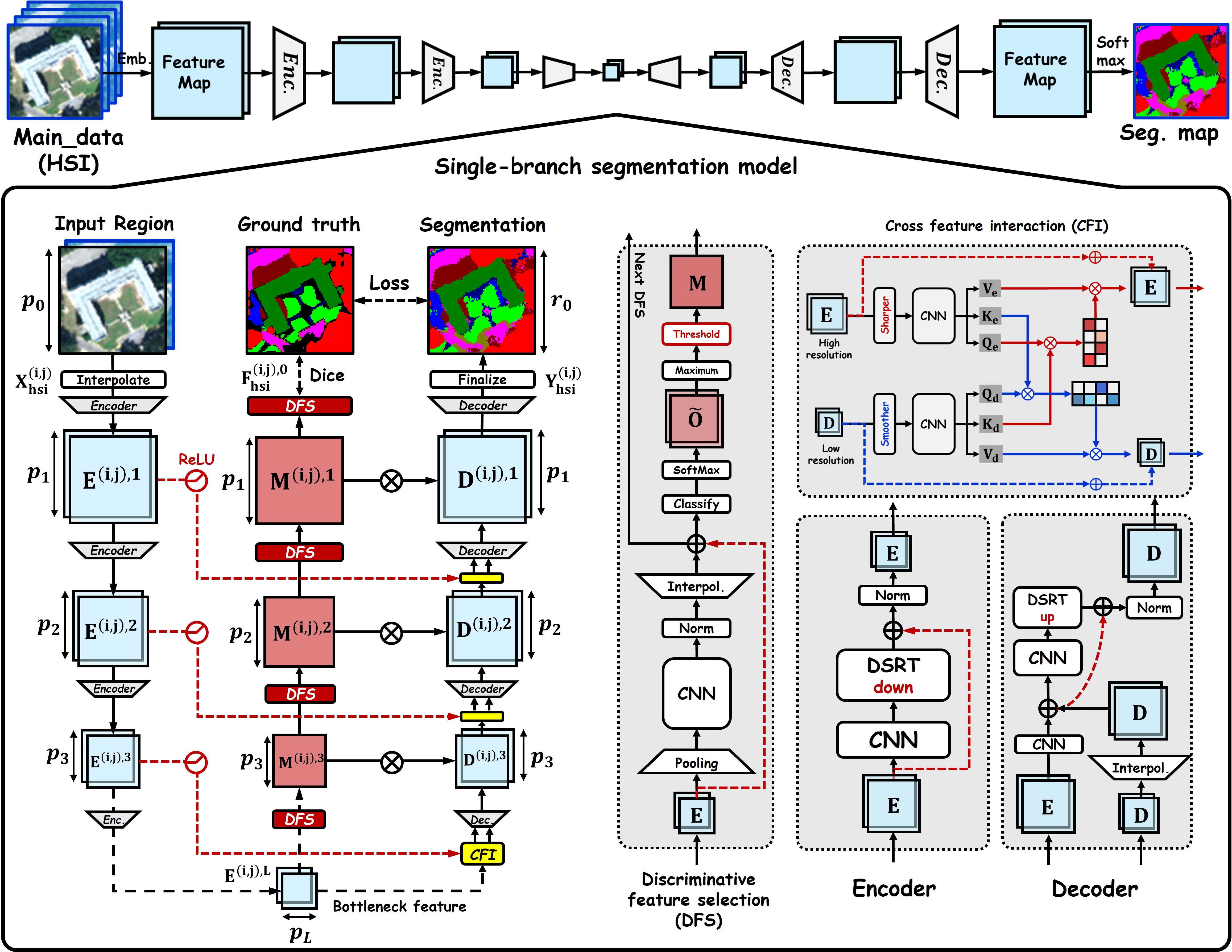}
    \caption{Single-branch segmentation model with HSI data only. U-shape manner with one additional DFS branch (middle branch) is collaborated for discriminative positional selection at decoding stage. The encoder integrates CNNs with the proposed DSRT to effectively capture spectral-spatial features. Additionally, CFI module facilitates cross-symmetric attention, enabling better interaction between high-resolution and low-resolution features. The decoder processes two interacted feature streams using CNNs and DSRT for up-sampling and feature reconstruction. The final segmentation map is optimized using cross-entropy loss to ensure accurate pixel-level label predictions. Furthermore, the discriminative positional feature map is aligned with the ground truth by Dice loss, reinforcing spatial structural consistency.}
    \label{fig:single_branch}
\end{figure}

To establish a baseline study, we modularized the model structure to be extensible and adaptable, allowing modifications as needed. The main branch, which utilizes only HSI data, can be extracted for standalone use. In this section, we describe the single-branch segmentation procedure, as illustrated in Fig.~\ref{fig:single_branch}.

The single-branch model adopts a U-shaped architecture, inspired by the well-known U-Net design for image segmentation. As illustrated in Fig.~\ref{fig:single_branch}, it consists of multiple encoder layers for down-sampling and multiple decoder layers for up-sampling. Unlike conventional U-Net-based segmentation models, our approach introduces several key innovations: (1) DFS module: Fully leverages the latent bottleneck features to selectively refine representations during the decoding process, (2) CFI module: Captures cross-symmetric feature correlations between high-resolution and low-resolution features, replacing conventional summation or concatenation-based fusion, and (3) DSRT: Integrated into both the encoder and decoder for dynamic regional self-attention, employing residual learning for adaptive local and global feature aggregation.

\subsubsection{Down-Sampling Encoder}
To aggregate features from high to low resolution, we first initialize the HSI patch sample by initializing its dimensionality from \(\mathcal{C}_1\) to \(d\).

Each encoder follows a residual learning framework. First, CNN layers perform local feature aggregation (e.g., depth-wise convolution) and dynamic positional embedding (e.g., point-wise convolution). The CNN-extracted features are then fed into the DSRT module to capture adaptive local and global dependencies. The channel attention is captured by the general SE attention \cite{se_attn}. Finally, a residual connection is applied to enhance feature representation before normalization, formulated as:
\begin{equation}
    \mathbf{E}^{(i,j),l} = Norm \Big( SE \Big( DSRT \big( ReLU \big( CNN \big(\mathbf{E}^{(i,j),l-1} \big) \big) \big) \big) + \mathbf{E}^{(i,j),l-1} \Big),
\end{equation}

where \(\mathbf{E}\) denotes the encoded feature, and \(l \geq 1\) represents the encoder stage index. For \(l = 0\), the initial encoded feature \(\mathbf{E}^{(i,j),0}\) corresponds to the input HSI data, \(\mathbf{X}^{(i,j)}\).

\subsubsection{Cross Feature Interaction (CFI)}
Before the decoding stage, effectively integrating multi-resolution features is crucial for maintaining spatial coherence and semantic consistency in the segmentation task \cite{multiscaleseg}. The proposed CFI module applies symmetric cross-attention between two feature maps from different branches: a low-resolution feature map \(\mathbf{D}^{(i,j),l}\) at stage \(l\) and a high-resolution feature map \(\mathbf{E}^{(i,j),l-1}\) at stage \(l-1\), each with a different resolution. 
% By dynamically computing spatial relationships between these feature maps, the CFI module enhances complementary regions, resulting in refined outputs for the decoding stage.

As illustrated in Fig.~\ref{fig:single_branch}, to prepare the feature maps for interaction, a smoothing operation (adaptive average pooling) is applied to the low-resolution feature map \(\mathbf{D}^{(i,j),l}\), while a sharpening operation (adaptive max pooling) is applied to the high-resolution feature map \(\mathbf{E}^{(i,j),l-1}\). These operations emphasize low-frequency information in \(\mathbf{D}^{(i,j),l}\) and high-frequency information in \(\mathbf{E}^{(i,j),l-1}\), ensuring more effective feature interaction. The transformations are formulated as:
\begin{align}
\mathbf{D}^{(i,j),l} &\leftarrow Smoother
\big( \mathbf{D}^{(i,j),l} \big), \\
\mathbf{E}^{(i,j),l-1} &\leftarrow Sharper \big( \mathbf{E}^{(i,j),l-1} \big).
\end{align}

For each feature map, the query (\(\mathbf{Q}\)), key (\(\mathbf{K}\)), and value (\(\mathbf{V}\)) representations are computed using point-wise CNN layers to project feature dimension from $d$ to $3d$. For the high-resolution feature map, these transformations are defined as:

\begin{equation}
    \mathbf{Q}_e, \mathbf{K}_e, \mathbf{V}_e = Split \Big( CNN \big( \mathbf{E}^{(i,j),l-1} \big) \Big).
\end{equation}

Similarly, for the low-resolution feature map:

\begin{equation}
    \mathbf{Q}_d, \mathbf{K}_d, \mathbf{V}_d = Split \Big( CNN \big( \mathbf{D}^{(i,j),l} \big) \Big).
\end{equation}

The CFI module computes symmetric cross-attention weights between two feature maps to identify complementary spatial regions. The attention weights for \(\mathbf{E}^{(i,j),l-1}\) are computed using queries (\(\mathbf{Q}_e\)) from \(\mathbf{D}^{(i,j),l}\) and keys (\(\mathbf{K}_d\)) from \(\mathbf{D}^{(i,j),l}\):

\begin{equation}
\mathbf{A}_e = Softmax \left(\frac{\mathbf{Q}_e \mathbf{K}_d^\top}{\sqrt{d_k}}\right),
\end{equation}

where \(d_k\) represents the key dimension scaling factor.

Similarly, the attention weights for \(\mathbf{D}^{(i,j),l}\) are computed as:
\begin{equation}
\mathbf{A}_d = Softmax \left(\frac{\mathbf{Q}_d \mathbf{K}_e^\top}{\sqrt{d_k}}\right).
\end{equation}

The attention matrices are then used to aggregate the corresponding value representations, \(\mathbf{V}_e\) and \(\mathbf{V}_d\):

\begin{align}
\mathbf{\Bar{E}}^{(i,j),l-1} &= \mathbf{A}_e \mathbf{V}_e, \\
\mathbf{\Bar{D}}^{(i,j),l} &= \mathbf{A}_d \mathbf{V}_d.
\end{align}

The aggregated outputs are activated using the GELU function and integrated with the original inputs via residual connections:

\begin{align}
\mathbf{\Hat{E}}^{(i,j),l-1} &= GELU \Big(\mathbf{\Bar{E}}^{(i,j),l-1} \Big) + \mathbf{E}^{(i,j),l-1}, \\
\mathbf{\Hat{D}}^{(i,j),l} &= GELU \Big(\mathbf{\Bar{D}}^{(i,j),l} \Big) + \mathbf{D}^{(i,j),l}.
\end{align}

The refined feature maps, \(\mathbf{\Hat{E}}^{(i,j),l-1}\) and \(\mathbf{\Hat{D}}^{(i,j),l}\), are returned for further processing in the decoder.

The CFI mechanism ensures complementary information exchange, leading to improved feature representation and better decoding performance.

\subsubsection{Up-Sampling Decoder}
The decoder is designed to seamlessly integrate multi-resolution features and perform adaptive up-sampling, ensuring accurate spatial detail reconstruction.

Given the previously interacted low-resolution feature map \(\mathbf{\Hat{D}}^{(i,j),l}\) from the decoder and the high-resolution feature map \(\mathbf{\Hat{E}}^{(i,j),l-1}\) from the skip connection, the low-resolution feature is first interpolated to match the spatial dimensions of \(\mathbf{E}^{(i,j),l-1}\):

\begin{equation}
\mathbf{\Tilde{D}}^{(i,j),l} = Interpolate \big(\mathbf{D}^{(i,j),l} \big).
\end{equation}

Simultaneously, the shallow channel dimensions of \(\mathbf{E}^{(i,j),l}\) are aligned with \(\mathbf{D}^{(i,j),l-1}\) using a point-wise CNN:

\begin{equation}
\mathbf{\Tilde{E}}^{(i,j),l} = CNN \big(\mathbf{E}^{(i,j),l} \big).
\end{equation}

To effectively fuse the two features, we first concatenate them along the channel dimension:
\begin{equation}
    \mathbf{\Hat{D}}^{(i,j),l} = Concat \big(\mathbf{\Tilde{D}}_{hsi}^{(i,j),l},\, \mathbf{\Tilde{E}}^{(i,j),l} \big).
\end{equation}

The combined feature \(\mathbf{\Hat{D}}^{(i,j),l}\) then serves as the residual base for subsequent local and global feature generation using the CNN and the proposed DSRT module:

\begin{equation}
    \mathbf{D}^{(i,j),l-1} = Norm \Big( SE \Big( DSRT \big( ReLU \big( CNN \big( \mathbf{\Hat{D}}^{(i,j),l} \big) \big) \big) \big) + \mathbf{\Hat{D}}^{(i,j),l} \Big), 
\end{equation}

where \(\mathbf{D}^{(i,j),l-1}\) is the output feature for the next decoder layer, and normalization is applied to ensure stable gradient flow and consistent feature distributions.

The decoder module effectively integrates multi-resolution features while adaptively focusing on spatially complementary regions, ensuring both spatial detail preservation and semantic consistency.

\subsubsection{Discriminative Feature Selection (DFS)}

In most segmentation models, the bottleneck feature provides a compact latent-space representation of the entire image, capturing high-level abstract features. However, its direct contribution to the decoding process remains uncertain. Typically, up-sampling in the decoder restores the image progressively, yet it lacks a mechanism to distinguish between beneficial and irrelevant features.

To overcome this limitation, inspired by \cite{ynet}, we propose the DFS module, which selectively enhances and up-samples discriminative features. By assuming that regions with lower activation probabilities contribute less to meaningful reconstruction, DFS adaptively suppresses less informative features while amplifying critical regions, leading to improved feature refinement in the decoding stage.

As shown in Fig.~\ref{fig:single_branch}, the DFS module begins by processing the bottleneck feature \(\mathbf{E}^{(i,j),L}\). Inspired by Pyramid Scene Parsing (PSP) \cite{psp}, it first applies adaptive pooling to a predefined resolution before passing the feature through CNN layers. This process is formulated as:

\begin{equation}
\label{eq:dfs}
\mathbf{O}^{(i,j),L} = Interpolate \Big( Norm \Big( CNN \Big( Pool \big( \mathbf{E}^{(i,j),L} \big) \Big) \Big) \Big) + \mathbf{E}^{(i,j),L},
\end{equation}
where \(\mathbf{O}^{(i,j),L}\) is one of the DFS module outputs, serving as input for the next DFS stage in the up-sampling process.

On the other hand, \(\mathbf{O}^{(i,j),L}\) is classified at the current scale using a point-wise CNN, which maps the feature dimension to the number of classes. The classification process is defined as:

\begin{equation}
    \mathbf{\Tilde{O}}^{(i,j),L} = Softmax \Big( CNN \big( \mathbf{O}^{(i,j),L} \big) \Big),
\end{equation}
where the softmax function normalizes the probabilities between 0 and 1 for each pixel belonging to each class. The resulting tensor \(\mathbf{\Tilde{O}}^{(i,j),L} \in \mathbb{R}^{p_L \times p_L \times K}\), where \(K\) is the number of classes.  

The maximum probability across the class dimension is extracted at each spatial position:
\begin{equation}
\mathbf{P}^{(i,j),L} = Maximum \big( \mathbf{\Tilde{O}}^{(i,j),L} \big),
\end{equation}
where \(\mathbf{P}^{(i,j),L} \in \mathbb{R}^{p_L \times p_L}\) records the maximum probability at each position.

To adaptively mask less discriminative regions, we first compute the mean and standard deviation of the probability map \(\mathbf{P}^{(i,j),L}\) as:

\begin{equation}
    \mu_1 = \frac{1}{N} \sum_{i=1}^{N} \mathbf{P}^{(i,j),L}(x_i), \quad 
    \sigma_1 = \sqrt{\frac{1}{N} \sum_{i=1}^{N} \left( \mathbf{P}^{(i,j),L}(x_i) - \mu_1 \right)^2},
\end{equation}
where \(N = p_L \times p_L\) is the total number of elements in \(\mathbf{P}^{(i,j),L}\).

Next, we define an adaptive threshold \(T_1\) with an upper bound \(\tau_1\) as:
\begin{equation}
    \tau_1 \geq T_1 \geq \mu_1 + \sigma_1,
\end{equation}
where \(\tau_1 \in [0,1]\) is a hyper-parameter that will be analyzed in later experiments. If \(\tau_1 \leq \mu_1 + \sigma_1\), then \(T_1\) is set to \(\mu_1 + \sigma_1\).

A binary mask is then generated by comparing \(\mathbf{P}^{(i,j),L}\) to the threshold \(T_1\):
\begin{equation}
\mathbf{M}^{(i,j),L}(h, w) = 
\begin{cases} 
1, & \text{if } \mathbf{P}^{(i,j),L}(h, w) \geq T_1, \\
0, & \text{otherwise}.
\end{cases}
\end{equation}
where \((h, w)\) denotes the spatial position in the feature map.

The resulting discriminative mask \(\mathbf{M}^{(i,j),L}\) is applied via element-wise multiplication with the corresponding scale-decoded feature \(\mathbf{D}^{(i,j),L}\), enhancing salient regions in the decoded features.

To obtain the next scale feature \(\mathbf{O}^{(i,j),L-1}\) via up-sampling, we follow Eq.~\ref{eq:dfs}:
\begin{equation}
    \mathbf{O}^{(i,j),L-1} = Interpolate \Big( Norm \Big( CNN \Big( \operatorname{Pool} \big( \mathbf{O}^{(i,j),L} \big) \Big) \Big) \Big) + \mathbf{O}^{(i,j),L}.
\end{equation}

Finally, the initial scale discriminative mask \(\mathbf{F}^{(i,j),0}\) is used to compute the Dice loss, enforcing shape alignment with the original ground truth to further improve model training.

\subsubsection{Dynamic Shifted Regional Transformer (DSRT)}

\begin{figure}[!t]
    \centering
    \includegraphics[scale=1]{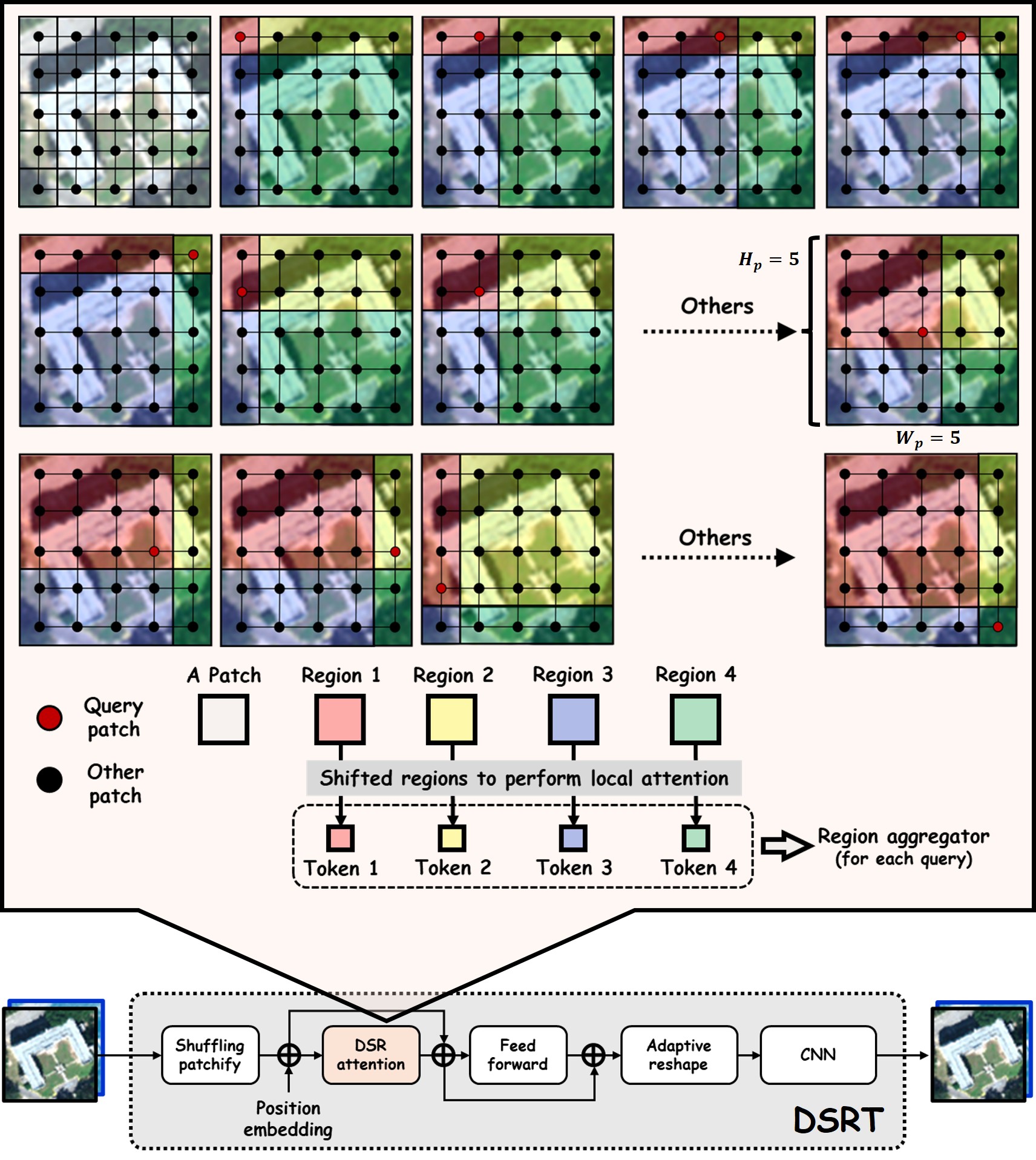}
    \caption{Illustration of DSRT and query-driven dynamic shifted regional attention. Using a \(5 \times 5\) grid of patches as an example, each query serves as the partition point, dividing the image into four position-aware regions: left-top, right-top, left-bottom, and right-bottom. After computing local window attention within each region, a region aggregator is introduced to integrate multi-region features, enhancing global dependency modeling.
 }
    \label{fig:DSRT}
\end{figure}

While our benchmark, Swin Transformer and its derivative, SwinUnet, have demonstrated impressive performance in general computer vision tasks, their direct application to HSI classification presents several potential issues. First, the fixed window partitioning strategy in them divides feature maps into uniform, non-overlapping regions, which may restrict the model’s flexibility in capturing spatial features of varying scales. This rigid structure could be problematic in HSI, where land-cover objects often exhibit irregular shapes and diverse spatial extents that do not align with fixed grids. Second, although the shifted window mechanism was designed to introduce cross-window interactions and mitigate boundary discontinuities, it remains suboptimal for modeling long-range dependencies, especially when complex spatial structures span multiple windows. Finally, in HSI classification, where discriminative features may be concentrated in specific, localized areas, this lack of query-aware adaptability can result in suboptimal feature representation. 

Thus, we propose the DSRT, to dynamically model shifted local and global contextual dependencies, as illustrated in Fig.~\ref{fig:DSRT}. This module focuses on adaptively aggregating and correlating multi-shape spatial regions in a position-aware manner. The design incorporates two key components:  (1) query-driven dynamic shifted region generation and (2) multi-region aggregation, which collectively enable region-specific interactions and adaptive integration using query-driven attention.

The dynamic region generation mechanism partitions the spatial feature map into four deformable regions based on the query position. Consider an embedded feature map \(\mathbf{E}\) as an example (omitting the head notation for simplicity). For each spatial query position \((h, w)\) after patchification, the surrounding feature map is divided into four shifted regions:
\begin{equation}
\begin{aligned}
    \mathbf{R}_{1} &=
    \begin{bmatrix}
    \mathbf{E}[0,0] & \mathbf{E}[0,1] & \dots & \mathbf{E}[0,w] \\
    \mathbf{E}[1,0] & \mathbf{E}[1,1] & \dots & \mathbf{E}[1,w] \\
    \vdots & \vdots & \ddots & \vdots \\
    \mathbf{E}[h,0] & \mathbf{E}[h,1] & \dots & \mathbf{E}[h,w]
    \end{bmatrix}, \\[8pt]
    \mathbf{R}_{2} &=
    \begin{bmatrix}
    \mathbf{E}[0,w] & \mathbf{E}[0,w+1] & \dots & \mathbf{E}[0,W_p-1] \\
    \mathbf{E}[1,w] & \mathbf{E}[1,w+1] & \dots & \mathbf{E}[1,W_p-1] \\
    \vdots & \vdots & \ddots & \vdots \\
    \mathbf{E}[h,w] & \mathbf{E}[h,w+1] & \dots & \mathbf{E}[h,W_p]
    \end{bmatrix}, \\[8pt]
    \mathbf{R}_{3} &=
    \begin{bmatrix}
    \mathbf{E}[h,0] & \mathbf{E}[h,1] & \dots & \mathbf{E}[h,w] \\
    \mathbf{E}[h+1,0] & \mathbf{E}[h+1,1] & \dots & \mathbf{E}[h+1,w] \\
    \vdots & \vdots & \ddots & \vdots \\
    \mathbf{E}[H_p-1,0] & \mathbf{E}[H_p-1,1] & \dots & \mathbf{E}[H_p-1,w]
    \end{bmatrix}, \\[8pt]
    \mathbf{R}_{4} &=
    \begin{bmatrix}
    \mathbf{E}[h,w] & \mathbf{E}[h,w+1] & \dots & \mathbf{E}[h,W_p-1] \\
    \mathbf{E}[h+1,w] & \mathbf{E}[h+1,w+1] & \dots & \mathbf{E}[h+1,W_p-1] \\
    \vdots & \vdots & \ddots & \vdots \\
    \mathbf{E}[H_p-1,w] & \mathbf{E}[H_p-1,w+1] & \dots & \mathbf{E}[H_p - 1,W_p-1]
    \end{bmatrix}.
\end{aligned}
\end{equation}
where \(H_p\) and \(W_p\) represent the number of partitioned patches in height and width, respectively, with \(0 \leq h \leq H_p\) and \(0 \leq w \leq W_p\). As shown in Fig.~\ref{fig:DSRT}, we assume \(H_p = W_p = 5\).

Each region is processed using Window Attention (WA), which transforms it into a fixed-size representation token. The WA mechanism is designed to handle regions of varying shapes and sizes:
\begin{equation}
\mathbf{\Hat{R}}_g = WA \big(\mathbf{R}_g \big),
\end{equation}
where \( g \) represents the region index, with \( g = 1,2,3,4 \).

The DWA operation follows the ViT framework \cite{vit0} but with a 1D-CNN, embedding a representation token before applying the standard self-attention mechanism:
\begin{equation}
    \mathbf{\Hat{R}}_g = ViT \Big( CNN \Big( Concat \Big[ \mathbf{R}^{cls}_g, \mathbf{R}_g[0], \mathbf{R}_g[1], \dots, \mathbf{R}_g[N_g] \Big] + \operatorname{Pos} \Big) \Big),
\end{equation}
where the learnable region token \( \mathbf{R}^{cls}_g \) is extracted to represent the entire region \( \mathbf{R}_g \). Here, \( N_g \) denotes the number of elements in the region \( \mathbf{R}_g \). $\operatorname{Pos}$ represents the positional embedding. This operation ensures that all regions contribute uniformly to the subsequent attention computation, regardless of their size.

After obtaining the four regional representation tokens (\(\mathbf{\Hat{R}}_1\), \(\mathbf{\Hat{R}}_2\), \(\mathbf{\Hat{R}}_3\), and \(\mathbf{\Hat{R}}_4\)), the Region Aggregator computes query-driven correlations between each position and its dynamically defined regions.

Unlike standard multi-head self-attention, we introduce a multi-region mechanism by defining the patchified feature map as:
\begin{equation}
    \mathbf{Q} \in \mathbb{R}^{head \times N \times dim}, \quad
    \mathbf{K} = \mathbf{V} = Concat \Big( [\mathbf{\Hat{R}}_1, \mathbf{\Hat{R}}_2, \mathbf{\Hat{R}}_3, \mathbf{\Hat{R}}_4] \Big).
\end{equation}

Thus, the key and value matrices are structured as:
\begin{equation}
    \mathbf{K} \in \mathbb{R}^{head \times N \times R \times dim}, \quad
    \mathbf{V} \in \mathbb{R}^{head \times N \times R \times dim},
\end{equation}
where \( head \) is the number of multi-heads, \( N \) is the total number of tokens, \( R = 4 \) is the number of multi-regions, and \( dim \) is the feature dimension.

The computational process follows these steps:
\begin{align}
    \mathbf{Q} &\leftarrow Linear(\mathbf{Q}),\\
    \mathbf{K} &\leftarrow Linear(\mathbf{K}),\\
    \mathbf{V} &\leftarrow Linear(\mathbf{V}),
\end{align}
where \(Linear\) represents a fully connected layer applied along the feature dimension \( dim \).

Next, the attention scores between the query and key vectors are computed using the scaled dot-product formula:
\begin{equation}
    \mathbf{A}_{n, r} = \frac{\mathbf{Q}_n \mathbf{K}_r^\top}{\sqrt{dim}},
\end{equation}
where \( n \in [1, N] \) corresponds to the sequence position, and \( r \in [1, R] \) denotes the region index. The key matrix transposed, \( \mathbf{K}^\top \), has dimensions \( \mathbb{R}^{head \times N \times dim \times R} \).

The attention scores are then normalized along the regional dimension (\( R \)) using the softmax function:
\begin{equation}
    \operatorname{Attn}_{n, r} = \frac{\exp(\mathbf{A}_{n, r})}{\sum_{r=1}^{R} \exp(\mathbf{A}_{n, r})},
\end{equation}
ensuring that the attention weights sum to 1 for each query.

Finally, the aggregated feature output is computed by applying the attention weights to the values (\(\mathbf{V}\)):
\begin{equation}
    \mathbf{R}_{out} = \operatorname{Attn} \cdot \mathbf{V},
\end{equation}
where \( \operatorname{Attn} \in \mathbb{R}^{head \times N \times R} \) and \( \mathbf{V} \in \mathbb{R}^{head \times N \times R \times dim} \).

This region-based attention mechanism aggregates information from multiple spatial regions and aligns it with query features, making the module effective for tasks requiring fine-grained spatial understanding.

\subsection{Multi-Branch Data Collaboration (Auxiliary Branch)\label{multibranch}}
\begin{figure}[!t]
    \centering
    \includegraphics[scale=0.32]{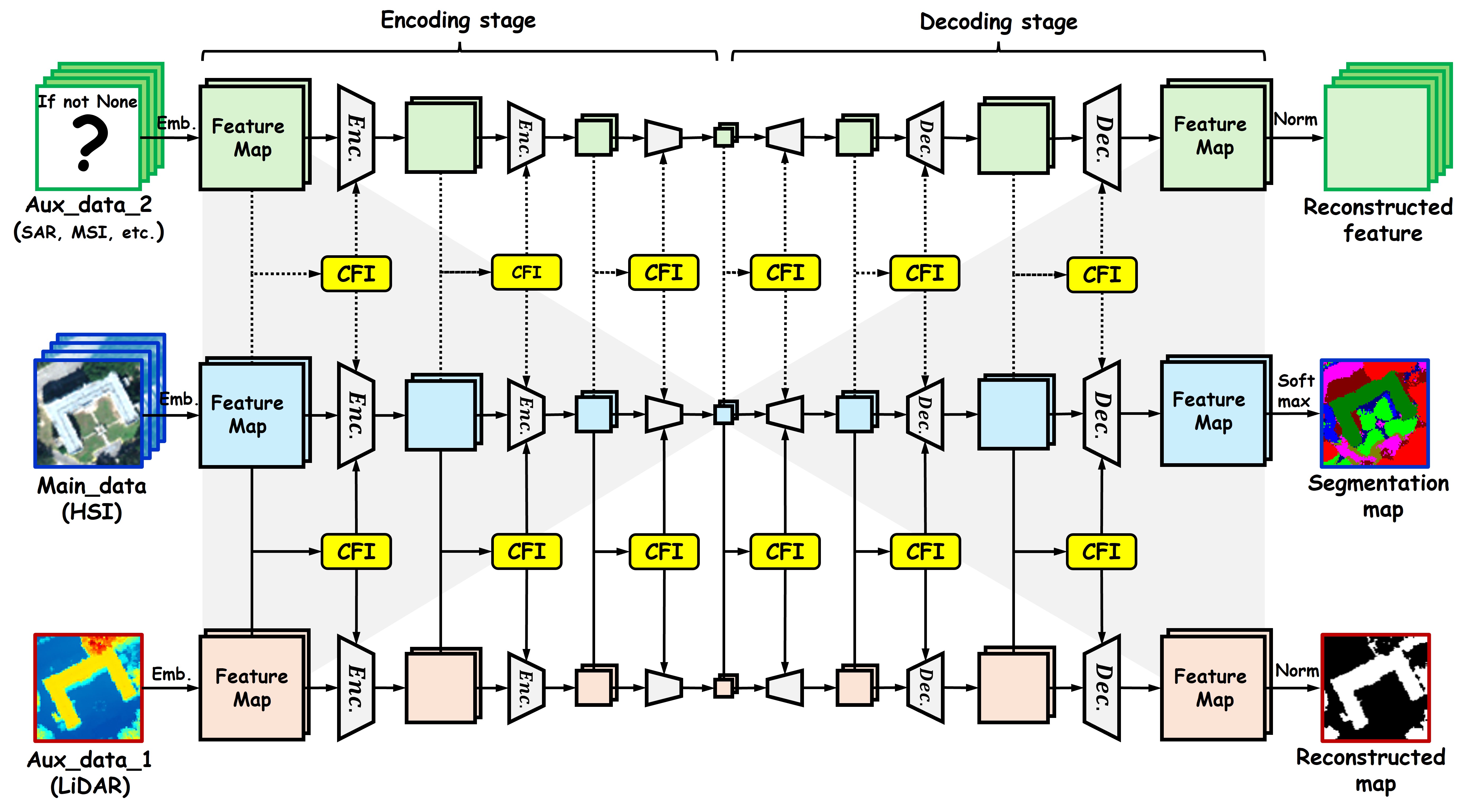}
    \caption{Illustration of multi-branch data collaboration for HSI segmentation. Auxiliary data serves as a feature enhancement for the main HSI branch, with feature interaction facilitated through the CFI module at each encoder and decoder stage. To effectively train the auxiliary branches, each is structured as an individual end-to-end model for reconstruction.}
    \label{fig:multi-branch}
\end{figure}
To integrate auxiliary data that provides complementary features for land-cover classification, we extend the single-branch HSI classification model into a multi-branch architecture, as shown in Fig.~\ref{fig:multi-branch}. The branches are interconnected via the previously introduced CFI module, enabling effective multi-modal feature fusion. This framework is scalable, supporting both dual-branch and triple-branch configurations when additional data sources are available.

Unlike the primary HSI branch, which is optimized for classification, the auxiliary branch functions as a supplementary feature extractor. Instead of performing classification, it is trained for reconstruction, preserving the structural and contextual properties of the auxiliary data.

In this setup, the model processes two inputs: the HSI data sample \(\mathbf{X}_{hsi}^{(i,j)}\) and an auxiliary data sample \(\mathbf{X}_{aux}^{(i,j)}\), which could be LiDAR (\(\mathbf{X}_{lidar}^{(i,j)}\)), SAR (\(\mathbf{X}_{sar}^{(i,j)}\)), or MSI (\(\mathbf{X}_{msi}^{(i,j)}\)).

The auxiliary data follows the same segmentation pipeline, but without the DFS module. Instead of classification at the final stage, it undergoes reconstruction via reconstruction loss by comparing the reconstructed feature to the original auxiliary data.

Feature interaction between the HSI and auxiliary branches is conducted using the CFI module, which ensures information exchange at each stage of the encoder:
\begin{equation}
    \mathbf{\hat{E}}_{hsi}^{(i,j), l}, \mathbf{\hat{E}}_{aux}^{(i,j), l} = CFI\Big(\mathbf{E}_{hsi}^{(i,j), l}, \mathbf{E}_{aux}^{(i,j), l}\Big).
\end{equation}

After feature interaction, the encoder updates both branches as:
\begin{equation}
    \mathbf{E}_{hsi}^{(i,j), l+1} = Encoder_{hsi} \Big(\mathbf{\hat{E}}_{hsi}^{(i,j), l} + \mathbf{E}_{hsi}^{(i,j), l}\Big),
\end{equation}
\begin{equation}
    \mathbf{E}_{aux}^{(i,j), l+1} = Encoder_{aux} \Big(\mathbf{\hat{E}}_{aux}^{(i,j), l} + \mathbf{E}_{aux}^{(i,j), l}\Big).
\end{equation}

Similar interactions occur in the decoder stage, where the feature maps from both branches are fused before up-sampling:
\begin{equation}
    \mathbf{\hat{D}}_{hsi}^{(i,j), l}, \mathbf{\hat{D}}_{aux}^{(i,j), l} = CFI\Big(\mathbf{D}_{hsi}^{(i,j), l}, \mathbf{D}_{aux}^{(i,j), l}\Big).
\end{equation}

Then, each branch updates its decoded features as:
\begin{equation}
    \mathbf{D}_{hsi}^{(i,j), l-1} = Decoder_{hsi} \Big(\mathbf{\hat{D}}_{hsi}^{(i,j), l} + \mathbf{D}_{hsi}^{(i,j), l}\Big),
\end{equation}
\begin{equation}
    \mathbf{D}_{aux}^{(i,j), l-1} = Decoder_{aux} \Big(\mathbf{\hat{D}}_{aux}^{(i,j), l} + \mathbf{D}_{aux}^{(i,j), l}\Big).
\end{equation}

At the final stage, the decoded HSI feature \( \mathbf{D}_{hsi}^{(i,j), 0} \in \mathbb{R}^{p \times p \times K} \) is used for segmentation via a cross-entropy loss with the corresponding patch labels \( \mathbf{Y}^{(i,j)}_{hsi} \). The decoded auxiliary feature \( \mathbf{D}_{aux}^{(i,j), 0} \in \mathbb{R}^{p \times p \times K_2} \) is compared with the original input \( \mathbf{X}_{aux}^{(i,j)} \) via mean square error (MSE) loss and perceptual loss (SSIM loss) for reconstruction, where \( K_2 \) is the number of channels in the auxiliary data.

The total loss function is defined as:
\begin{equation}
    \mathcal{L}_{total} = \alpha \times \mathcal{L}^{seg}_{hsi} + \beta \times \mathcal{L}^{dice}_{hsi} + \gamma \times \mathcal{L}^{rec}_{aux} + \delta \times \mathcal{L}^{ssim}_{aux},
\end{equation}
where \( \alpha, \beta, \gamma, \delta \) are learnable weights, normalized such that:
\begin{equation}
    \alpha + \beta + \gamma + \delta = 1.
\end{equation}

These weights balance the contributions of segmentation and reconstruction losses for optimal model training.

\subsection{Progressive learning with pseudo labeling\label{progressive learning}}
\begin{figure}[!t]%% placement specifier
\centering%% For centre alignment of image.
\includegraphics[scale=0.32]{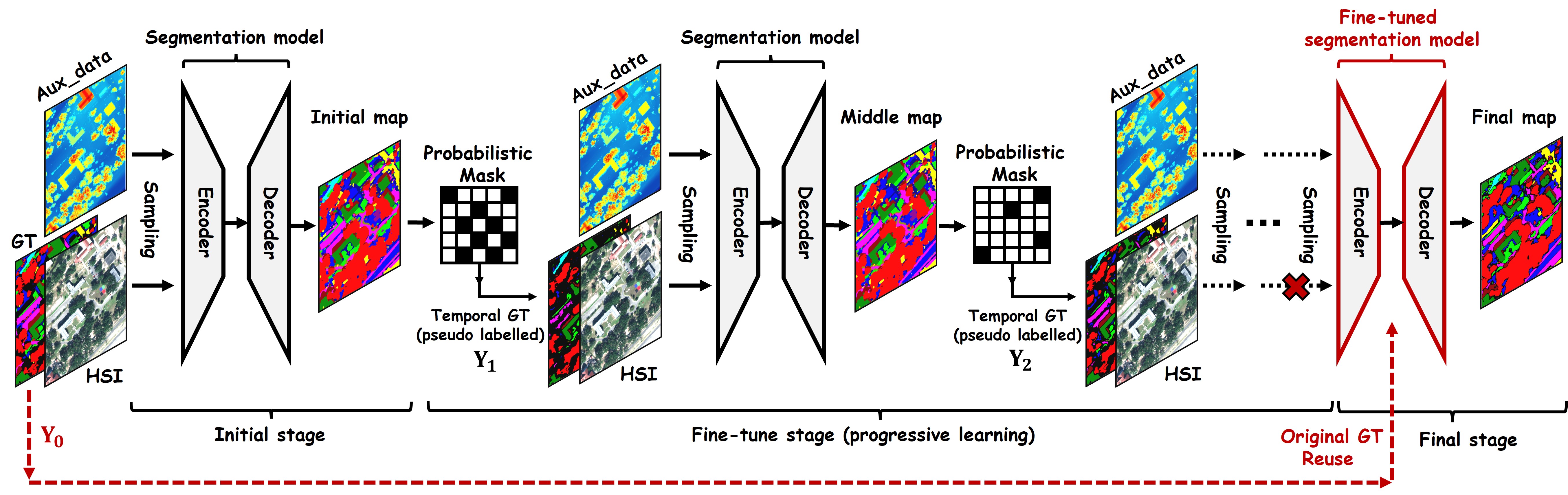}
\caption{The illustration of the progressive learning process with pseudo-labeling. At the initial stage, the original ground truth (\(\mathbf{Y}_{0}\)) is utilized for sample selection and evaluation. The segmentation results (initial map) are then processed through a probabilistic masking operation, which selects discriminatively classified pixels with high confidence scores. These selected pixels are used to generate a temporal pseudo-labeled ground truth, denoted as \(\mathbf{Y}_{temp}\), which is subsequently fed into the next fine-tuning stage.
This iterative process progressively refines the pseudo-labeled ground truth at each step. After multiple iterations, the fine-tuned model is finalized, while the original ground truth (\(\mathbf{Y}_{0}\)) is reused for the final evaluation to ensure consistency with the original labeled samples. }
\label{model_training}
\end{figure}

To effectively utilize unlabeled pixels from the original ground truth and support model training on unknown features, particularly in segmentation models, we propose a progressive learning framework with an adaptive pseudo-labeling strategy, as illustrated in Fig.~\ref{model_training}.

This probability-based adaptive pseudo-labeling allows the model to iteratively refine predictions by learning from pseudo-labeled pixels, which originate from the initially unlabeled regions. The process begins with the original ground truth (\(\mathbf{Y}_0\)), and temporal ground truths (\(\mathbf{Y}_{temp}\)) are generated at each iteration, where \( temp \in \mathbb{Z}, temp \geq 1 \). The number of iterations acts as a hyper-parameter, determining the level of refinement.

At the initial stage, a standard HSI classification task is conducted using a limited number of training samples. The model generates a classification map for the entire land-cover area, where each pixel is assigned a class label based on the highest probability in a one-hot-like vector.

To refine these predictions dynamically, low-confidence predictions are iteratively masked, enhancing the model's robustness and generalization. 

The classification probability output of the model at each iteration is defined as:
\begin{equation}
    \mathbf{O}_{temp} = Softmax \Big( Model \Big(\mathbf{X}_{hsi}, \mathbf{X}_{aux}, \mathbf{Y}_{temp} \Big) \Big),
\end{equation}
where \( \mathbf{O}_{temp} \in \mathbb{R}^{\mathcal{H} \times \mathcal{W} \times \mathcal{K}} \), with \( \mathcal{K} \) representing the number of classes.

The maximum probability across the class dimension is extracted for each spatial position:
\begin{equation}
    \mathbf{\hat{Y}}_{temp} = Maximum (\mathbf{O}_{temp}),
\end{equation}
where \( \mathbf{\hat{Y}}_{temp} \in \mathbb{R}^{\mathcal{H} \times \mathcal{W}} \), recording the maximum probability for each pixel.

To adaptively mask less discriminative regions, we compute the mean value and standard deviation of the probability map:
\begin{equation}
    \mu_2 = Mean (\mathbf{\hat{Y}}_{temp}), \quad
    \sigma_2 = Std (\mathbf{\hat{Y}}_{temp}).
\end{equation}

A dynamic threshold \( T_2 \) with an upper bound \( \tau_2 \) is defined as:
\begin{equation}
    \tau_2 \geq T_2 \geq \mu_2 + \sigma_2,
\end{equation}
where \( \tau_2 \in [0, 1] \) is a hyper-parameter investigated in the experiments. If \( \tau_2 \leq \mu_2 + \sigma_2 \), then \( T_2 \) is set as \( \mu_2 + \sigma_2 \). 

A binary mask is generated based on the computed threshold:
\begin{equation}
    \mathbf{M}_{temp} (h, w) =
    \begin{cases} 
    1, & \text{if } \mathbf{\hat{Y}}_{temp} (h, w) \geq T_2, \\
    0, & \text{otherwise}.
    \end{cases}
\end{equation}

The masked probability map is used to construct the next iteration of temporal ground truth \( \mathbf{Y}_{temp+1} \), ensuring that only high-confidence pixels contribute:
\begin{equation}
    \mathbf{Y}_{temp+1} (h, w) = argmax \big(\mathbf{O}_{temp} (h, w) \big) \times \mathbf{M}_{temp} (h, w),
\end{equation}
where \( (h, w) \) represents the spatial position in the feature map.

For the next step in the progressive learning process, the masking threshold is iteratively updated to enhance discriminative selection:
\begin{equation} 
\tau_2 \leftarrow \tau_2 + \zeta,
\end{equation}
where $\zeta$ is a tolerance value that incrementally adjusts the threshold. This adjustment accounts for the model’s increasing discriminative capability at each position, leading to higher confidence in classification probabilities over successive iterations. Empirically, we set $\zeta = 0.005$ to ensure a gradual refinement of the pseudo-labeling process while maintaining stability in training.

The pseudo-labeled temporal ground truth is iteratively updated for each fine-tuning stage. To assess the effectiveness of progressive learning with pseudo-labeling, the final model is re-evaluated using the original ground truth, HSI data, and auxiliary data, ensuring a comprehensive investigation of performance improvements.

\section{Experiments and Analysis\label{experiment}}
In this section, we presents the experiments and ablation studies for model evaluation and further analysis.
\subsection{Datasets}
\begin{figure}[!t]
    \centering
    \includegraphics[scale=0.58]{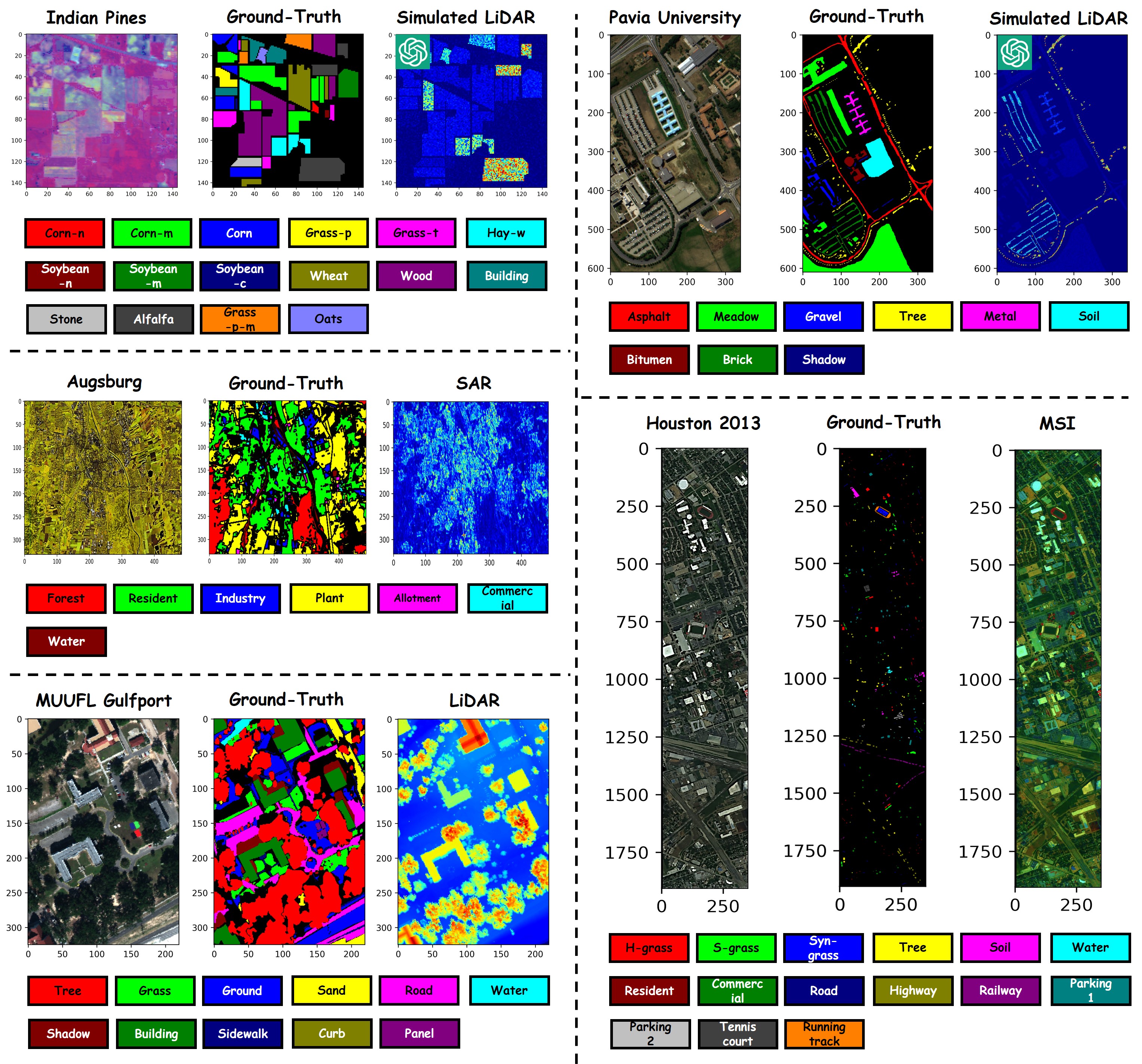}
    \caption{Five datasets are used for the experiments. Two well-known datasets, IP and PU, are complemented with simulated LiDAR images generated using ChatGPT. The AG dataset includes real SAR imagery, while the MG dataset is paired with real LiDAR data. Additionally, the HU dataset incorporates real MSI imagery.}
    \label{fig:datasets}
\end{figure}

To evaluate the proposed model, experiments were conducted on five datasets: Indian Pines (IP), Pavia University (PU), MUUFL Gulfport (MG), Augsburg (AG), and Houston 2013 (HU), as illustrated in Fig.~\ref{fig:datasets}. These datasets include both widely recognized benchmarks and custom datasets enriched with additional modalities, such as simulated or real LiDAR, SAR, or MSI. The characteristics of each dataset are detailed in Table~\ref{tab:datasets}, while the number of training and testing samples is listed in Table~\ref{tab:samples}. 

For training, we randomly select 10 samples per class, with the remaining samples used for testing. This small-sample training strategy is designed to assess the model's generalization capability and robustness, particularly in real-world scenarios where acquiring labeled HSI data is costly and labor-intensive. Moreover, this experimental setup effectively evaluates the proposed progressive pseudo-labeling strategy, which enables the model to leverage unlabeled data, bridging the gap between limited labeled pixels and the abundant unlabeled regions typically found in RS applications.

One difference of this study is the first-time evaluation of simulated LiDAR images generated by ChatGPT for HSI classification. This investigation explores the feasibility and effectiveness of using generative approaches to augment auxiliary modalities. By comparing classification results obtained with simulated LiDAR against those using real LiDAR or SAR data, we analyze the advantages and limitations of synthetic data for enhancing HSI classification performance in remote sensing.

\begin{table}[!t]
\scriptsize
\centering
\caption{Summary of the datasets' information used in the experiments.}
\label{tab:datasets}
\begin{tabular}{l|cccc}
\hline
\textbf{Dataset }     & \textbf{Spatial Size} & \textbf{Spectral Bands} & \textbf{Classes} & \textbf{Additional Modalities} \\ \hline
Indian Pines (IP)   & $145 \times 145$      & 200                     & 16                         & Simulated LiDAR (1 band)               \\
Pavia University (PU) & $610 \times 340$      & 103                     & 9                          & Simulated LiDAR (1 band)               \\
MUUFL Gulfport (MG)  & $325 \times 220$      & 72                      & 11                         & LiDAR (1 band)                    \\
Augsburg (AG) & $332 \times 485$     & 80                     & 7                       & SAR (4 bands)          \\ 
Houston 2013 (HU) & $1905 \times 349$ & 144 & 15 & MSI (8 bands) \\ \hline
\end{tabular}
\end{table}

\begin{table}[!t]
\centering
\scriptsize
\caption{Summary of the training and testing samples for each dataset.}
\label{tab:samples}
\begin{tabular}{c|cc|cc|cc|cc|cc}
\hline
      \textbf{Dataset} & \multicolumn{2}{c|}{\textbf{IP}} & \multicolumn{2}{c|}{\textbf{PU}} & \multicolumn{2}{c|}{\textbf{MG}} & \multicolumn{2}{c|}{\textbf{AG}} & \multicolumn{2}{c}{\textbf{HU}}  \\\hline
      No.   & Train  & Test    & Train  & Test    & Train  & Test    & Train  & Test   & Train  & Test\\\hline
\cellcolor{class-1}1           & 10     &36              & 10     &6621            & 10     &23236             & 10     &13497    &10 &1241    \\
\cellcolor{class-2}2           & 10     &1418           & 10     &18639            & 10     &4260              & 10     &30319     &10 &1244    \\
\cellcolor{class-3}\color{white}3         & 10     &820             & 10     &2089             & 10     &6872              & 10     &3841    &10 &687    \\
\cellcolor{class-4}4           & 10     &227            & 10     &3054              & 10     &1816             & 10     &26847     &10 &1234    \\
\cellcolor{class-5}5          & 10     &473            & 10     &1335             & 10     &6677             & 10     &565    &10 &1232     \\
\cellcolor{class-6}6          & 10     &720             & 10     &5019            & 10     &456             & 10     &1635     &10 &315    \\
\cellcolor{class-7}\color{white}7           & 10     &18           & 10     &1320             & 10     &2223          & 10     &1520    &10 &1258    \\
\cellcolor{class-8}\color{white}8           & 10     &468            & 10     &3672              & 10     &6230              &        &     &10 &1234   \\
\cellcolor{class-9}\color{white}9          & 10     &10             & 10     &937             & 10     &1375       &               &     &10 &1242    \\
\cellcolor{class-10}\color{white}10         & 10     &962       &       &                     & 10     &173             &           & &10  &1217    \\
\cellcolor{class-11}\color{white}11         & 10     &2445       &       &                    & 10     &259       &                   & &10 &1225    \\
\cellcolor{class-12}\color{white}12          & 10     &583       &               &              &        &              &        &      &10 &1223   \\
\cellcolor{class-13}\color{white}13          & 10     &195       &       &        &              &               &               &        &10 &459 \\
\cellcolor{class-14}\color{white}14        & 10     &1255       &       &        &              &               &               &        &10 &418 \\
\cellcolor{class-15}15         & 10     &376       &       &               &               &              &             &  & 10 &650   \\
\cellcolor{class-16}16         & 10     &83       &               &              &               &       &        &     & &    \\ \hline
Total        &160        &10089             &90        &42686              &110        &53577              &70        &78224     &150 &14879    \\ \hline
% Ratio        &1.56\%     &98.44\%              &0.21\%        &99.78\%              &0.2\%        &99.8\%              &0.09\%        &99.91\%    &0.1\% &99.9\%    \\ \hline
\end{tabular}
\end{table}

\subsection{Brief Description of Compared Methods}
To evaluate the effectiveness of the proposed method, we compare it against two categories of existing approaches:  
\begin{enumerate}
    \item Existing methods designed for HSI classification – These models are specifically developed to address HSI classification and serve as strong baselines for performance comparison.
    \item Segmentation-based models adapted for HSI classification – These methods originate from conventional image segmentation frameworks and we adapted to hyperspectral data to explore their feasibility in HSI classification.
\end{enumerate}
Below, we provide a brief overview of the methods included in each category.

\subsubsection{Existing Methods for HSI Classification}
 We have selected several existing works, including both single-source and multi-source HSI classification methods. The chosen methods are:
\begin{itemize}
    \item Single-source HSI classification methods: SSFTT~\cite{vit2}, TransHSI~\cite{transhsi}, and MiM~\cite{mim}.
    \item Multi-source HSI classification methods: DFI~\cite{dfi}, MFT~\cite{mft}, and DeepSFT~\cite{deepsft}.
\end{itemize}

The selection of these methods is based on their availability as open-source implementations on GitHub, ensuring reproducibility. 

\begin{itemize}
    \item \(\text{SSFTT}^1\): Combines 3D-CNNs and 2D-CNNs with a Gaussian-weighted feature tokenizer to prepare features for the Transformer encoder, enabling spectral-spatial fusion for HSI classification.

    \item \(\text{TransHSI}^1\): Uses PCA for dimensionality reduction, followed by spectral and spatial feature extraction. A tokenized spectral-spatial fusion module integrates 3D-CNNs, 2D-CNNs, and Transformers for classification. 

    \item \(\text{MiM}^1\): Introduces a centralized Mamba-cross-scan with a tokenized Mamba model. It employs a Gaussian decay mask, semantic token learner, and multi-scale loss design to enhance classification accuracy. 

    \item \(\text{DFI}^2\): A depth-wise feature interaction network for wetland classification, using cross-attention to extract self- and cross-correlation from multi-source feature pairs. It is optimized using consistency, discrimination, and classification losses. 
    
    \item \(\text{MFT}^2\): Implements multi-head cross-patch attention for HSI classification, incorporating auxiliary data sources in the Transformer encoder for better generalization.

    \item \(\text{DeepSFT}^2\): Uses local-global mixture blocks to extract hierarchical features. A local-global feature mixer and a symmetric fusion Transformer adaptively aggregate and fuse features from multiple modalities.
\end{itemize}

The head note indicates the number of input-sources used in each model.

\subsubsection{Segmentation-Based Models}
The application of segmentation models, particularly U-Net-based architectures, in HSI classification remains under-explored. There is a notable lack of literature investigating and comparing these models within the context of HSI tasks. 

Given the extensive variety of U-Net-based segmentation models, we select four representative architectures for investigation and comparison: original U-Net~\cite{unet}, TransUnet~\cite{transunet}, SwinUnet~\cite{swinunet}, and MambaUnet~\cite{mambaunet}. These models have been chosen for their prominence and diverse architectural designs, providing a comprehensive benchmark for evaluating their effectiveness in HSI classification. The brief descriptions of these models are as follows:
\begin{itemize}
    \item \(\text{U-Net}^1\): A fully convolutional network with an encoder-decoder structure and skip connections, widely used for pixel-wise prediction tasks. 

    \item \(\text{TransUnet}^1\): Combines a Transformer-based encoder with the U-Net framework, leveraging self-attention mechanisms to capture long-range dependencies. 

    \item \(\text{SwinUnet}^1\): Uses a Swin Transformer encoder, introducing hierarchical self-attention with shifted windows for improved feature extraction.
    
    \item \(\text{MambaUnet}^1\): Enhances U-Net by integrating the state-space model Mamba, originally developed for medical image segmentation.
\end{itemize}

By evaluating these U-Net-based architectures, we aim to provide insights into their applicability, adaptability, and performance in HSI classification.

\subsection{Experimental Settings}

\subsubsection{Evaluation Metrics}
To quantitatively assess the effectiveness of the proposed method and benchmark models, we employ four evaluation metrics: overall accuracy (OA), average accuracy (AA), kappa coefficient (Kappa), and classification accuracy for each land-cover category. A higher value for each metric indicates better classification performance.

\subsubsection{Experimental Environment}
All experiments were conducted using PyTorch on an Ubuntu 24.04 system. The hardware configuration includes an Intel Core i7-8700 CPU, 32GB RAM, and an NVIDIA GeForce RTX 3080 (11GB VRAM) GPU.

\subsubsection{Initial Parameter Settings}
\begin{table}[!t]
\scriptsize
\centering
\caption{The summary of hyper-parameter and experimental settings for five datasets.}
\label{tab:hyperparameter settings}
\begin{tabular}{c|c|c|c|c|c}
\hline
Definition & IP & PU & MG & AG & HU \\ \hline

Dimension after PCA: & 30 & 15 & 30 & 15 & 30 \\ 

Initial Patch Size: & $33 \times 33$  & $49 \times 49$ & $41 \times 41$ &$33 \times 33$ & $57 \times 57$ \\ 

Feature Dimension Setting: & 64 & 32 & 64 & 64 & 32 \\ 

Number of layers in encoder and decoder: & 4 & 2 & 4 & 4 & 3 \\ 

Number of Multi-Heads: & 2 & 2 & 4 & 4 & 2 \\ 

Masking upper bound $\tau_1$: & 0.8 & 0.9 & 0.8 & 0.6 & 0.9 \\ 

Masking upper bound $\tau_2$: & 0.7 & 0.9 & 0.7 & 0.7 & 0.8 \\ 

Epoch: & 300 & 200 & 300 & 300 & 200 \\ 

Batch Size: & 32 & 64 & 32 & 32 & 64 \\ 

Learning Rate:  & 0.0015 & 0.001 & 0.0015 & 0.0015 & 0.001\\ 

Feature Dropout: & 0.1 & 0.15 & 0.1 & 0.1 & 0.15 \\ 

Number of iterations: & 14 & 8 & 10 & 10 & 8 \\ \hline

\end{tabular}

\end{table}

In our method, all datasets undergo dimensionality reduction via PCA before processing. The cropped HSI and auxiliary data patches are resized to a fixed spatial dimension of \(144 \times 144\) pixels. The proposed segmentation model consists of five down-sampling encoder stages and five up-sampling decoder stages. The feature map sizes follow a hierarchical structure:  
\[
144 \,(12\times12) \rightarrow 100 \,(10\times10) \rightarrow 64 \,(8\times8) \rightarrow 36 \,(6\times6) \rightarrow 16 \,(4\times4).
\]

Accordingly, the patchifying sizes in DSRT for the encoder and decoder are set to 12, 10, 8, 6, and 4 for square patchifying.

Additional hyperparameter settings used in the experiments are summarized in Table~\ref{tab:hyperparameter settings}.

\subsection{Experimental Results}
\subsubsection{Comparison on Classification Results}

\begin{figure}[!t]
    \centering
    \includegraphics[scale=0.65]{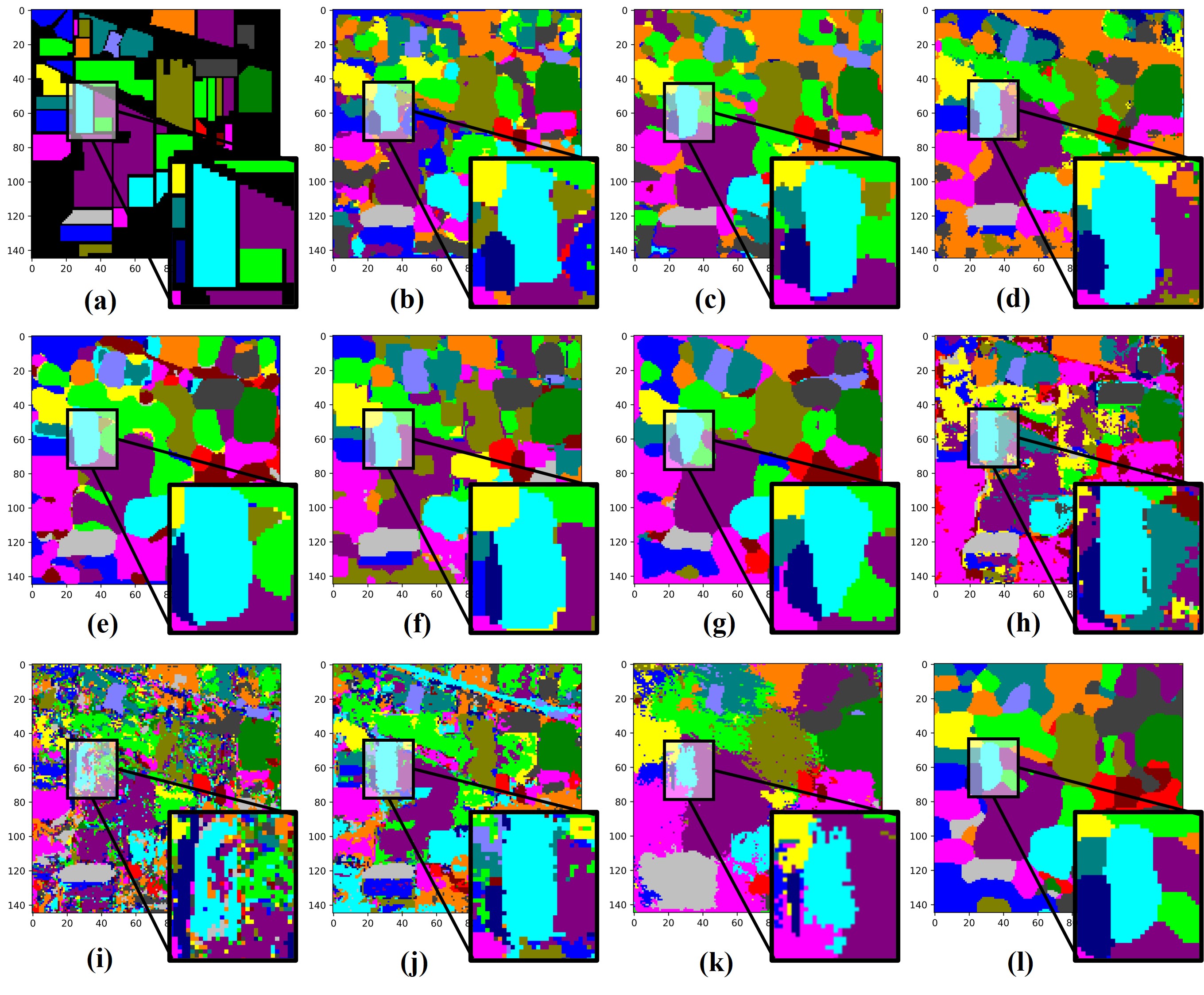}
    \caption{The comparison on IP dataset classification maps by different methods. (a) Ground-Truth. (b) SSFTT (OA=73.24\%). (c) TransHSI (OA=74.37\%). (d) MiM (OA=78.78\%). (e) DFI (OA=76.31\%). (f) MFT (OA=79.95\%). (g) DeepSFT (OA=83.07\%). (h) Unet (OA=64.87\%). (i) TransUnet (OA=57.87\%). (j) SwinUnet (OA=71.43\%). (k) MambaUnet (OA=76.36\%). (l) Ours (OA=85.28\%). }
    \label{fig:IP classification maps}
\end{figure}

\begin{figure}[!t]
    \centering
    \includegraphics[scale=0.65]{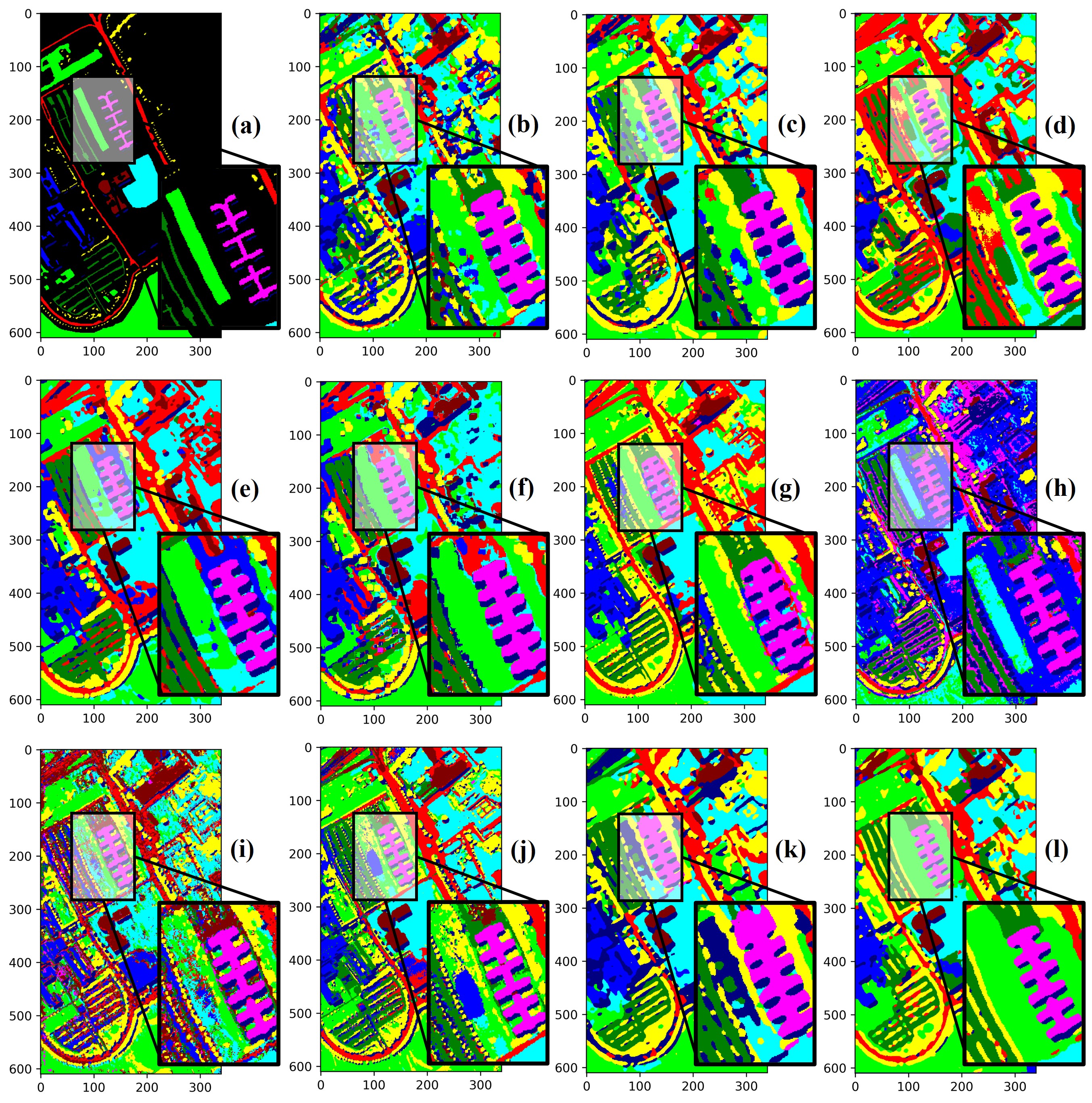}
    \caption{The comparison on PU dataset classification maps by different methods. (a) Ground-Truth. (b) SSFTT (OA=86.07\%). (c) TransHSI (OA=86.51\%). (d) MiM (OA=89.67\%). (e) DFI (OA=86.81\%). (f) MFT (OA=85.52\%). (g) DeepSFT (OA=90.86\%). (h) Unet (OA=66.24\%). (i) TransUnet (OA=74.09\%). (j) SwinUnet (OA=84.71\%). (k) MambaUnet (OA=85.37\%). (l) Ours (OA=93.37\%). }
    \label{fig:PU classification maps}
\end{figure}

\begin{figure}[!t]
    \centering
    \includegraphics[scale=0.69]{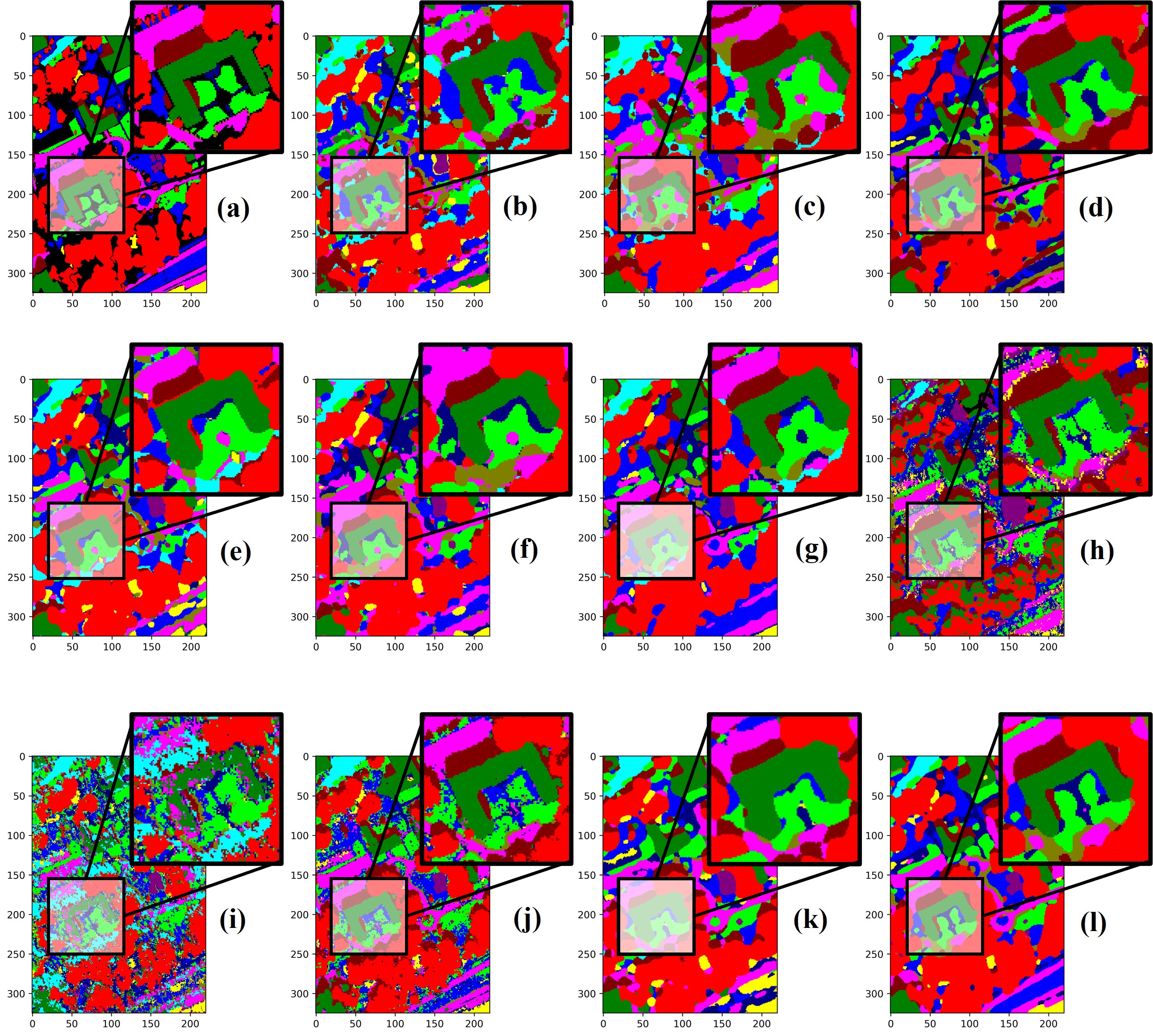}
    \caption{The comparison on MG dataset classification maps by different methods. (a) Ground-Truth. (b) SSFTT (OA=71.60\%). (c) TransHSI (OA=70.73\%). (d) MiM (OA=74.08\%). (e) DFI (OA=72.19\%). (f) MFT (OA=76.66\%). (g) DeepSFT (OA=79.25\%). (h) Unet (OA=53.87\%). (i) TransUnet (OA=59.18\%). (j) SwinUnet (OA=75.69\%). (k) MambaUnet (OA=75.86\%). (l) Ours (OA=81.99\%). }
    \label{fig:MG classification maps}
\end{figure}

\begin{figure}[!t]
    \centering
    \includegraphics[scale=0.55]{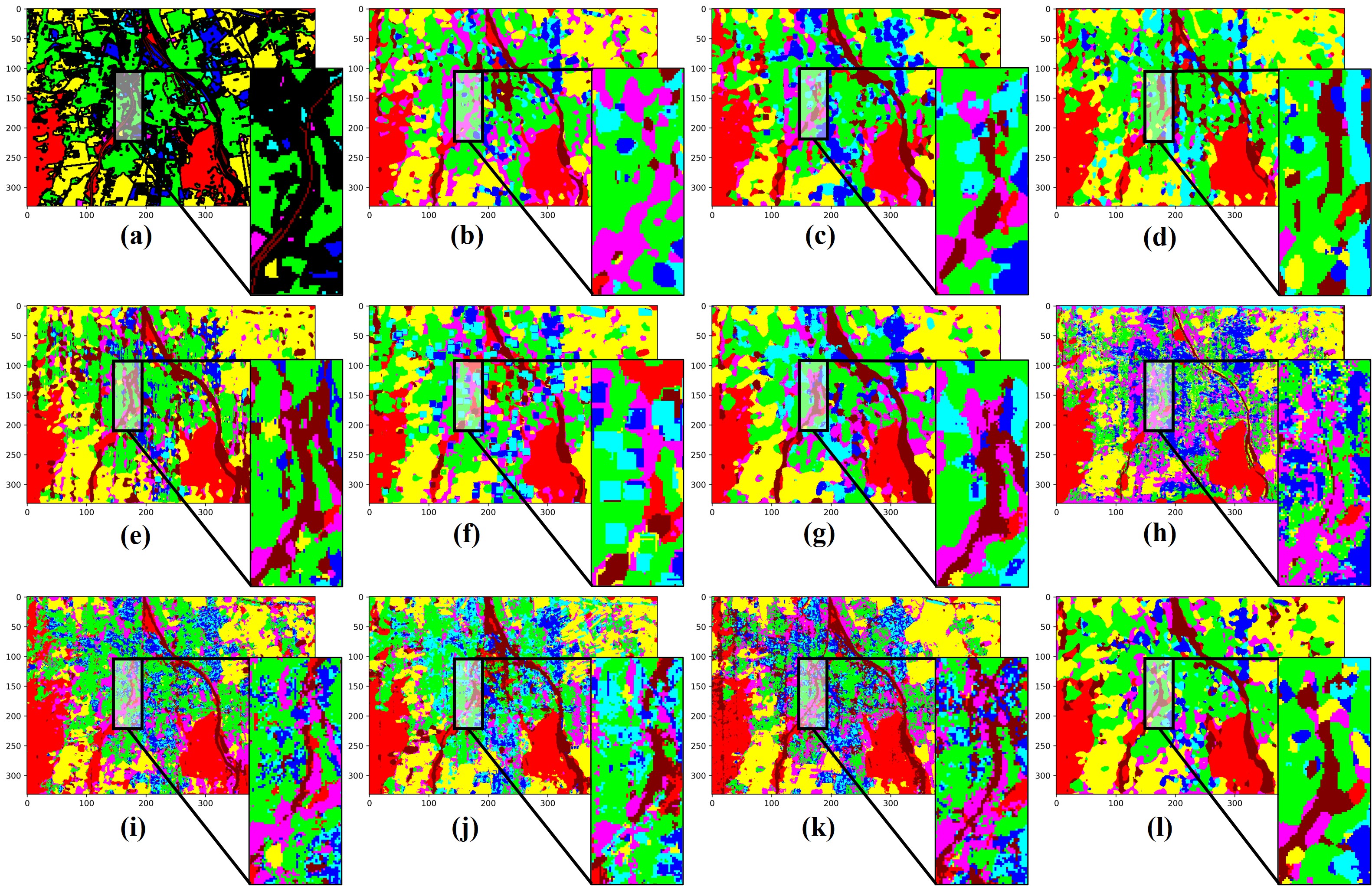}
    \caption{The comparison on AG dataset classification maps by different methods. (a) Ground-Truth. (b) SSFTT (OA=76.23\%). (c) TransHSI (OA=77.91\%). (d) MiM (OA=79.70\%). (e) DFI (OA=77.88\%). (f) MFT (OA=77.18\%). (g) DeepSFT (OA=80.80\%). (h) Unet (OA=63.05\%). (i) TransUnet (OA=71.13\%). (j) SwinUnet (OA=73.45\%). (k) MambaUnet (OA=74.44\%). (l) Ours (OA=82.64\%). }
    \label{fig:AG classification maps}
\end{figure}

\begin{figure}[!t]
    \centering
    \includegraphics[scale=0.6]{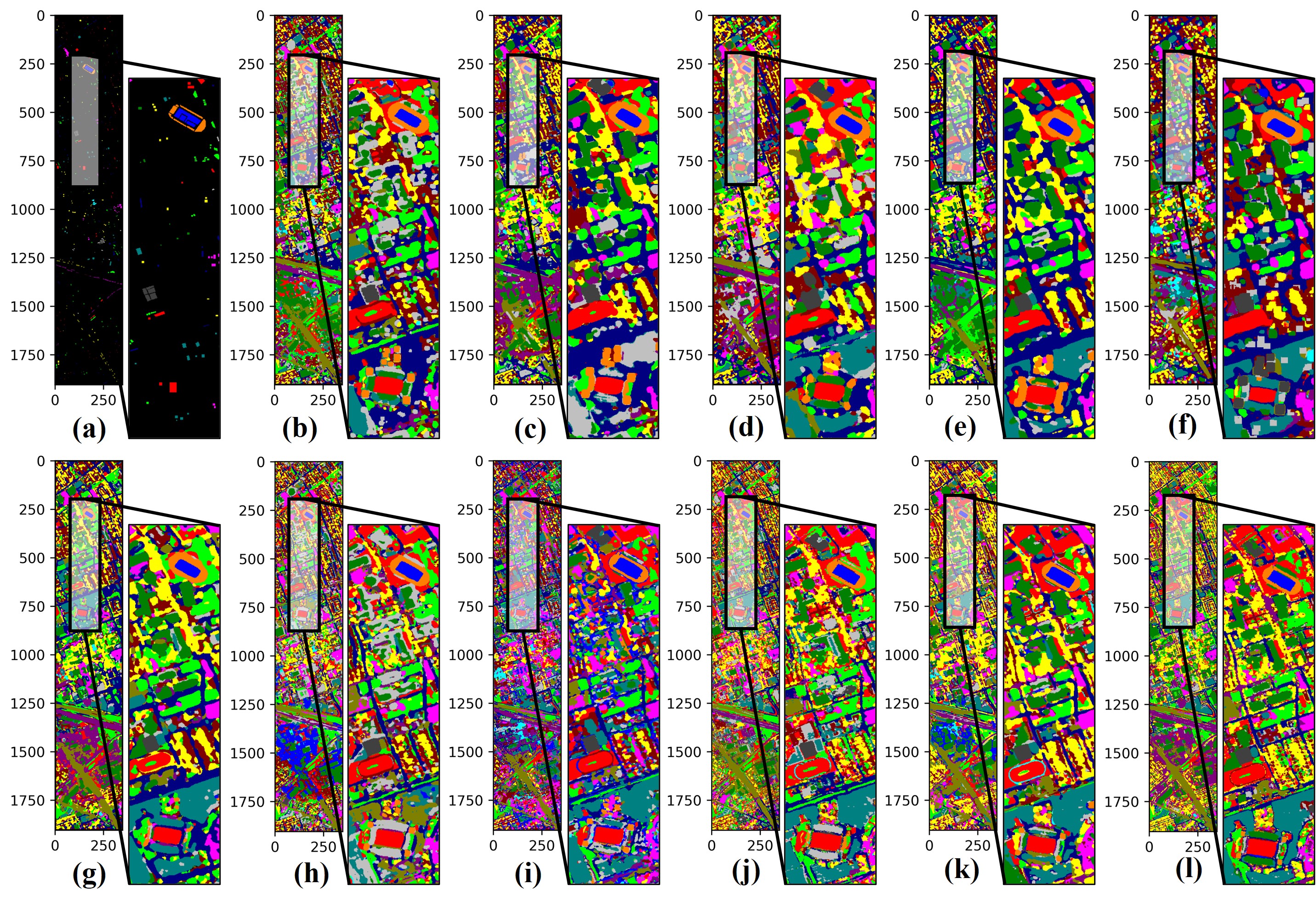}
    \caption{The comparison on HU dataset classification maps by different methods. (a) Ground-Truth. (b) SSFTT (OA=86.45\%). (c) TransHSI (OA=82.63\%). (d) MiM (OA=87.88\%). (e) DFI (OA=86.74\%). (f) MFT (OA=85.89\%). (g) DeepSFT (OA=89.56\%). (h) Unet (OA=80.93\%). (i) TransUnet (OA=81.05\%). (j) SwinUnet (OA=85.49\%). (k) MambaUnet (OA=84.73\%). (l) Ours (OA=92.31\%). }
    \label{fig:HU classification maps}
\end{figure}

The classification maps for the IP, PU, MG, AG, and HU datasets, obtained using different methods, are shown in Fig.s~\ref{fig:IP classification maps}, \ref{fig:PU classification maps}, \ref{fig:MG classification maps}, \ref{fig:AG classification maps}, and \ref{fig:HU classification maps}, respectively. Meanwhile, Tables~\ref{tab:IP table}, \ref{tab:PU table}, \ref{tab:MG table}, \ref{tab:AG table}, and \ref{tab:HU table} present the quantitative classification performance, including OA, AA, Kappa, and per-class accuracy. Certain regions in the classification maps are magnified for a more detailed comparison. Notably, all results are obtained from a single end-to-end training session without progressive learning (i.e., corresponding to the initial stage in Fig.~\ref{model_training}). Additionally, all results are averaged over five runs for reliability.

Among classification-based methods, SSFTT and TransHSI both incorporate CNNs for local feature extraction before self-attention in Transformers. While TransHSI slightly outperforms SSFTT due to its deeper feature tokenization layers, both methods suffer from noise-like misclassifications in large homogeneous regions and boundary misalignment with the ground truth. These noise artifacts often resemble the “salt-and-pepper effect,” where isolated misclassified pixels disrupt the spatial consistency in otherwise uniform areas, significantly degrading the visual quality of classification maps. MiM, which employs multi-scale tokenized Mamba networks with CNNs, improves classification accuracy over SSFTT and TransHSI, demonstrating the effectiveness of state-space models in low-sample scenarios. However, MiM still struggles with boundary misclassification, highlighting challenges in fine-grained spatial feature learning.

For multi-source methods, DFI, MFT, and DeepSFT integrate auxiliary data for improved classification. DFI, a CNN-based model, performs inconsistently, sometimes yielding lower accuracy than single-source models on IP and PU datasets, likely due to its incompatibility with the ChatGPT-generated LiDAR data. In contrast, MFT and DeepSFT, which combine CNNs and Transformers, achieve better feature fusion. DeepSFT, an enhanced version of MFT with a symmetric fusion Transformer, achieves the best results among multi-modal methods, even on IP and PU, validating the feasibility of simulated LiDAR. However, DeepSFT’s classification maps still exhibit noise artifacts in homogeneous regions and unclear object boundaries, indicating limitations in feature refinement. The presence of isolated misclassified pixels within uniform areas suggests insufficient contextual learning, similar to the noise-like misclassification patterns observed in single-source models.

Among segmentation-based models, U-Net and TransUnet exhibit the most severe noise-like and boundary misclassifications. Their classification maps are heavily affected by isolated errors in homogeneous regions and fragmented boundary delineation, indicating the need for substantial modifications to traditional segmentation architectures for effective HSI classification. Even with larger patch sizes, these models struggle to outperform small patch-based classifiers. SwinUnet, which applies shifted window self-attention, performs better but still struggles to produce smooth, uniform segmentations or clear boundaries, with residual misclassifications persisting in homogeneous areas. MambaUnet, which replaces Transformers with state-space models, shows moderate improvement over U-Net and TransUnet but is comparable to SwinUnet in both accuracy and visual clarity.

In contrast, our proposed method consistently outperforms all classification based and segmentation-based methods. It achieves the highest OA, AA, and Kappa across all datasets, improving OA by 2.21\%, 2.51\%, 2.74\%, 1.84\%, and 2.75\% over state-of-the-art classification models on IP, PU, MG, AG, and HU, respectively. Compared to segmentation-based models, it outperforms them by 8.92\%, 8.00\%, 6.13\%, 8.2\%, and 6.82\% in OA. Compared to SwinUnet, which employs a fixed shifted window for local self-attention, our method extends this to a query-driven dynamic window, achieving superior results.

\begin{table}[!t]
\centering
\tiny
\caption{Different methods in terms of OA, AA, and Kappa as well as the accuracies for each class on the \textbf{IP} dataset.}
\label{tab:IP table}
\begin{tabular}{c|cccccc|ccccc}
\hline 
      & \multicolumn{6}{c|}{Classification-based} & \multicolumn{5}{c}{Segmentation-based} \\ \hline
No.   &$\text{SSFTT}^1$       &$\text{TransHSI}^1$      &$\text{MiM}^1$      &$\text{DFI}^2$      &$\text{MFT}^2$      &$\text{DeepSFT}^2$      &$\text{Unet}^1$        &$\text{TransUnet}^1$       &$\text{SwinUnet}^1$       &$\text{MambaUnet}^1$       &$\text{Ours}^2$       \\ \hline
\cellcolor{class-1}1      &\textbf{100}       &\textbf{100}    &\textbf{100}  &97.22      &\textbf{100}    &\textbf{100}      &\textbf{100}        &\textbf{100}       &\textbf{100}       &97.22       &\textbf{100}       \\
\cellcolor{class-2}2      &65.66       &75.18      &69.81    &\textbf{80.39}    &46.19  &70.45     &26.23     &48.16    &74.11    &46.19     &73.41       \\
\cellcolor{class-3}\color{white}3     &70.37   &0.12  &64.02      &\textbf{73.65}  &73.04   &79.02 &41.21   &37.92   &51.09    &30.12       &61.58       \\
\cellcolor{class-4}4    &99.56       &96.48    &99.11    &87.22      &99.55  &96.91      &88.10    &57.70   &74.88       &\textbf{100}       &97.79       \\
\cellcolor{class-5}5      &90.06       &90.69   &92.17      &92.38      &77.58      &85.41      &87.31     &74.41       &91.75       &\textbf{93.44}       &79.06       \\
\cellcolor{class-6}6      &87.08       &93.33    &90.00   &91.80      &93.75      &94.86      &91.80      &61.25       &\textbf{95.27}       &87.36       &92.22       \\
\cellcolor{class-7}\color{white}7     &\textbf{100}       &\textbf{100}    &\textbf{100 }  &\textbf{100}      &\textbf{100}      &\textbf{100}      &\textbf{100}      &\textbf{100}       &\textbf{100}       &\textbf{100}       &\textbf{100}       \\
\cellcolor{class-8}\color{white}8      &98.50      &99.57    &\textbf{100}   &\textbf{100}      &97.43      &\textbf{100}      &96.79      &95.51       &99.35       &\textbf{100 }      &91.66       \\
\cellcolor{class-9}\color{white}9      &\textbf{100}       &\textbf{100}      &\textbf{100 }   &50.00      &\textbf{100}      &\textbf{100}      &\textbf{100}      &\textbf{100}       &\textbf{100}       &\textbf{100}       &\textbf{100}       \\
\cellcolor{class-10}\color{white}10    &77.44    &77.65      &82.95    &49.58   &77.33      &85.44      &26.50    &44.49       &61.12       &75.46       &\textbf{88.14}       \\
\cellcolor{class-11}\color{white}11   &44.70       &65.97   &66.66    &65.39   &72.14      &76.48      &60.20    &52.96       &63.84       &84.45       &\textbf{85.43}       \\
\cellcolor{class-12}\color{white}12    &62.44    &59.86      &67.40    &37.73   &66.89      &76.50      &70.15    &34.81       &69.81       &60.72       &\textbf{84.04}       \\
\cellcolor{class-13}\color{white}13    &\textbf{100}       &\textbf{100}      &99.48    &96.41      &\textbf{100}      &86.15      &\textbf{100}    &97.94       &98.97       &\textbf{100}       &\textbf{100}       \\
\cellcolor{class-14}\color{white}14    &87.89       &81.35      &90.75  &88.52      &94.58      &91.79      &99.92    &76.57       &91.31       &98.72       &\textbf{99.92} \\   
\cellcolor{class-15}15     &78.72       &\textbf{100}      &94.68      &81.91      &95.47      &95.21      &\textbf{100}        &66.48       &64.62       &82.71       &93.61       \\
\cellcolor{class-16}16    &95.18       &\textbf{100 }     &\textbf{100}    &34.93      &\textbf{100}      &95.18      &\textbf{100}        &98.79       &\textbf{100}       &\textbf{100}       &93.97       \\ \hline
OA    &73.24       &74.37      &78.78      &76.31      &79.95      &83.07      &64.87        &57.87       &71.43       &76.36       &\textbf{85.28}       \\
AA    &84.85       &83.76      &88.56      &82.70      &87.12      &89.58      &80.51        &71.69       &81.51       &84.77       &\textbf{90.02}       \\
Kappa &0.679       &0.684      &0.762      &0.709      &0.740      &0.809      &0.606        &0.528       &0.709       &0.732       &\textbf{0.833}       \\ \hline
% Train (sec.)    &21.11     &56.23     &165.9    &25.31   &35.02   &59.56   &57.19      &107.5       &250.9       &753.6       &1221.1     \\
% Test (sec.)    &2.59     &4.33   &139.63    &3.71    &3.48     &5.03     &10.44     &16.61      &45.36      &135.69      &155.19     \\ \hline
\end{tabular}
\end{table}

\begin{table}[!t]
\centering
\tiny
\caption{Different methods in terms of OA, AA, and Kappa as well as the accuracies for each class on the \textbf{PU} dataset.}
\label{tab:PU table}
\begin{tabular}{c|cccccc|ccccc}
\hline 
      & \multicolumn{6}{c|}{Classification-based} & \multicolumn{5}{c}{Segmentation-based} \\ \hline
No.   &$\text{SSFTT}^1$       &$\text{TransHSI}^1$      &$\text{MiM}^1$      &$\text{DFI}^2$      &$\text{MFT}^2$      &$\text{DeepSFT}^2$      &$\text{Unet}^1$        &$\text{TransUnet}^1$       &$\text{SwinUnet}^1$       &$\text{MambaUnet}^1$       &$\text{Ours}^2$       \\ \hline
\cellcolor{class-1}1      &47.06       &71.62      &94.26      &81.78      &65.39      &77.34      &51.83        &57.01       &83.12       &84.65       &\textbf{88.62}       \\
\cellcolor{class-2}2      &94.99       &84.23      &87.03      &85.35      &\textbf{96.94}      &96.45      &46.67        &77.08       &82.39       &84.45       &94.65       \\
\cellcolor{class-3}\color{white}3     &93.72       &85.20      &65.62      &87.79      &81.28      &90.61      &54.47        &93.01       &85.49       &\textbf{99.95}       &96.69       \\
\cellcolor{class-4}4    &94.79       &92.69      &91.61      &64.83      &49.44      &59.62      &96.66        &96.56       &\textbf{97.90}       &85.65       &93.61       \\
\cellcolor{class-5}5      &\textbf{100}       &99.25      &99.77      &79.40      &99.70      &99.02      &\textbf{100}        &\textbf{100}      &\textbf{100}       &92.65       &\textbf{100}       \\
\cellcolor{class-6}6      &79.17       &98.68      &86.90      &97.11      &88.88      &89.51      &99.58        &81.58       &99.98       &99.68       &\textbf{100}       \\
\cellcolor{class-7}\color{white}7     &99.09       &92.50      &94.77      &98.93      &98.63      &97.19      &84.84        &99.24       &100       &99.62       &\textbf{100}      \\
\cellcolor{class-8}\color{white}8      &60.26       &82.05      &91.09      &85.92      &57.81      &84.34      &79.86        &24.97       &87.09       &88.77       &\textbf{88.80}       \\
\cellcolor{class-9}\color{white}9      &97.71       &99.89      &98.93      &71.71      &88.58      &98.50      &\textbf{100}        &\textbf{100}       &99.89       &91.46       &\textbf{100}       \\ \hline
OA    &86.07       &86.51      &89.67      &86.81      &85.52      &90.86      &66.24        &74.09       &84.71       &85.37       &\textbf{93.37}       \\
AA    &88.31       &87.57      &90.01      &86.65      &82.73      &90.07      &81.54        &81.05       &92.87       &91.88       &\textbf{93.40}       \\
Kappa &0.828       &0.814      &0.853      &0.823      &0.801      &0.881      &0.594        &0.674       &0.823       &0.830       &\textbf{0.911}       \\ \hline
% Train (sec.)    &18.06     &32.12     &106.1    &22.54   &27.34   &39.57  &58.18      &83.52       &149.7       &651.6       &1346.2    \\
% Test (sec.)    &4.65     &8.58   &160.98    &5.87    &4.71     &8.04     &27.32     &20.22      &54.85      &124.4      &273.51    \\ \hline
\end{tabular}
\end{table}

\begin{table}[!t]
\centering
\tiny
\caption{Different methods in terms of OA, AA, and Kappa as well as the accuracies for each class on the \textbf{MG} dataset.}
\label{tab:MG table}
\begin{tabular}{c|cccccc|ccccc}
\hline 
      & \multicolumn{6}{c|}{Classification-based} & \multicolumn{5}{c}{Segmentation-based} \\ \hline
No.   &$\text{SSFTT}^1$       &$\text{TransHSI}^1$      &$\text{MiM}^1$      &$\text{DFI}^2$      &$\text{MFT}^2$      &$\text{DeepSFT}^2$      &$\text{Unet}^1$        &$\text{TransUnet}^1$       &$\text{SwinUnet}^1$       &$\text{MambaUnet}^1$       &$\text{Ours}^2$       \\ \hline
\cellcolor{class-1}1      &80.68       &73.29      &79.45      &81.67      &84.87      &\textbf{89.69}      &50.78        &67.66       &78.27       &80.31       &86.50       \\
\cellcolor{class-2}2      &57.51       &58.99      &77.91      &66.45      &61.31      &64.70      &61.33        &79.85       &83.59       &75.30       &\textbf{79.87}       \\
\cellcolor{class-3}\color{white}3     &73.18       &51.81      &48.64      &51.76      &43.91      &69.74      &26.86        &41.73       &54.17       &44.64       &\textbf{75.17}       \\
\cellcolor{class-4}4    &\textbf{97.24}       &86.61      &98.62      &84.80      &84.69      &80.67      &57.26        &83.14       &88.60       &95.98       &93.19       \\
\cellcolor{class-5}5      &51.14       &75.21      &59.27      &61.61      &79.07      &63.87      &42.57        &37.66       &84.48       &77.86       &\textbf{88.60}       \\
\cellcolor{class-6}6      &\textbf{100}       &\textbf{100}      &97.80      &\textbf{100}      &\textbf{100}      &\textbf{100}      &71.92        &98.90       &\textbf{100}       &\textbf{100}       &\textbf{100}       \\
\cellcolor{class-7}\color{white}7     &68.37       &78.22      &\textbf{86.45}      &48.62      &70.85      &78.36      &91.04        &29.32       &90.59       &65.00       &82.38       \\
\cellcolor{class-8}\color{white}8      &71.66       &81.70      &87.11      &84.36      &93.38      &82.72      &84.10        &56.99       &90.73       &\textbf{96.00}       &90.36       \\
\cellcolor{class-9}\color{white}9      &13.16       &40.00      &46.90      &37.89      &50.18      &43.78      &56.00        &49.67       &64.43       &41.81       &\textbf{72.22}       \\
\cellcolor{class-10}\color{white}10    &65.89       &63.00      &76.87      &52.60      &67.05      &53.17      &58.95        &69.36       &\textbf{99.42}       &30.05       &96.96       \\
\cellcolor{class-11}\color{white}11   &84.16       &98.45      &95.36      &99.22      &97.68      &98.06      &\textbf{100}        &91.11       &\textbf{100}       &99.61       &\textbf{100}       \\ \hline
OA    &71.60       &70.73      &74.08      &72.19      &76.66      &79.25      &53.87        &59.18       &75.69       &75.86       &\textbf{81.99}       \\
AA    &69.36       &73.39      &77.67      &69.91      &75.72      &75.07      &63.71        &64.13       &82.93       &73.32       &\textbf{85.99}       \\
Kappa &0.641       &0.635      &0.673      &0.644      &0.702      &0.730      &0.453        &0.505       &0.732       &0.705       &\textbf{0.777}       \\ \hline
% Train (sec.)    &18.24     &45.30     &141.5    &20.94   &27.78   &45.56    &78.16      &76.88       &137.4       &390.2       &1302.6      \\
% Test (sec.)    &5.41     &10.38   &119.38    &6.31    &5.29     &9.26     &41.31     &22.26      &64.66      &137.51      &253.48    \\\hline
\end{tabular}
\end{table}

\begin{table}[!t]
\centering
\tiny
\caption{Different methods in terms of OA, AA, and Kappa as well as the accuracies for each class on the \textbf{AG} dataset.}
\label{tab:AG table}
\begin{tabular}{c|cccccc|ccccc}
\hline 
      & \multicolumn{6}{c|}{Classification-based} & \multicolumn{5}{c}{Segmentation-based} \\ \hline
No.   &$\text{SSFTT}^1$       &$\text{TransHSI}^1$      &$\text{MiM}^1$      &$\text{DFI}^2$      &$\text{MFT}^2$      &$\text{DeepSFT}^2$      &$\text{Unet}^1$        &$\text{TransUnet}^1$       &$\text{SwinUnet}^1$       &$\text{MambaUnet}^1$       &$\text{Ours}^2$       \\ \hline
\cellcolor{class-1}1      &97.68       &91.49      &\textbf{98.31}     &93.48      &88.61      &90.47     &91.95        &94.09       &95.43       &89.81       &84.70       \\
\cellcolor{class-2}2      &71.92       &75.42      &81.66      &73.86      &71.89      &\textbf{90.09}     &43.38        &69.94       &63.67       &59.86       &87.99       \\
\cellcolor{class-3}\color{white}3     &35.17       &62.77      &40.90      &52.46      &39.78      &37.58     &\textbf{65.26}        &53.97       &53.81       &37.95       &45.11       \\
\cellcolor{class-4}4    &86.88       &82.06      &79.11      &82.47      &85.84      &79.15     &73.23        &66.24       &77.12       &88.74       &\textbf{88.89}       \\
\cellcolor{class-5}5      &92.03       &76.28      &72.90      &80.88      &60.17      &77.83     &98.05        &97.34       &89.02       &\textbf{98.58}       &87.61       \\
\cellcolor{class-6}6      &40.91       &41.46      &63.54      &15.04      &50.27      &\textbf{75.69}     &18.96        &32.59       &51.92       &52.41       &67.40       \\
\cellcolor{class-7}\color{white}7     &74.97       &63.15      &55.13      &69.40      &58.22       &56.82     &47.76        &52.36       &75.59       &\textbf{80.46}       &62.43      \\ \hline
OA    &76.23       &77.91      &79.70      &77.88      &77.18      &80.80      &63.05        &71.13       &73.45       &74.44       &\textbf{82.64}       \\
AA    &71.24       &70.37      &70.22      &66.80      &64.97      &72.52     &62.66        &66.62       &72.37       &72.56       &\textbf{74.85}       \\
Kappa &0.721       &0.713      &0.737      &0.700      &0.690      &0.765     &0.527        &0.622       &0.653       &0.667       &\textbf{0.787}       \\ \hline
% Train (sec.)    &16.61     &37.49     &127.9    &18.12   &28.56   &46.84  &30.75      &48.45       &102.6       &521.3       &1176.2      \\
% Test (sec.)    &6.12     &13.71   &135.15    &7.13    &6.25     &13.12     &22.64     &33.34      &84.90      &125.6      &344.56   \\\hline
\end{tabular}
\end{table}

\begin{table}[!t]
\centering
\tiny
\caption{Different methods in terms of OA, AA, and Kappa as well as the accuracies for each class on the \textbf{HU} dataset.}
\label{tab:HU table}
\begin{tabular}{c|cccccc|ccccc}
\hline 
      & \multicolumn{6}{c|}{Classification-based} & \multicolumn{5}{c}{Segmentation-based} \\ \hline
No.   &$\text{SSFTT}^1$       &$\text{TransHSI}^1$      &$\text{MiM}^1$      &$\text{DFI}^2$      &$\text{MFT}^2$      &$\text{DeepSFT}^2$      &$\text{Unet}^1$        &$\text{TransUnet}^1$       &$\text{SwinUnet}^1$       &$\text{MambaUnet}^1$       &$\text{Ours}^2$       \\ \hline
\cellcolor{class-1}1      &94.35       &95.24      &94.19      &85.49      &92.50      &86.78      &\textbf{97.09}        &95.64       &96.61       &\textbf{97.09}       &93.71       \\
\cellcolor{class-2}2      &98.07       &93.48      &97.58      &95.65      &90.51      &98.15      &\textbf{98.71}        &93.08       &96.30       &92.36       &97.99       \\
\cellcolor{class-3}\color{white}3     &98.68       &98.83      &99.70      &96.65      &99.27      &97.81      &\textbf{100}        &\textbf{100}       &\textbf{100}       &\textbf{100}       &\textbf{100}       \\
\cellcolor{class-4}4    &97.32       &94.97      &96.75      &91.81      &91.00      &87.43      &93.19        &95.29       &97.81       &95.86       &\textbf{99.91}       \\
\cellcolor{class-5}5      &99.91       &100      &99.10      &98.21      &98.13      &99.10      &\textbf{100}        &\textbf{100}       &\textbf{100}       &\textbf{100}       &\textbf{100}       \\
\cellcolor{class-6}6      &\textbf{100}       &99.68      &86.98      &95.55      &\textbf{100}      &97.77      &98.09        &\textbf{100}       &\textbf{100}       &\textbf{100}       &\textbf{100}       \\
\cellcolor{class-7}\color{white}7     &\textbf{92.44}       &84.73      &87.44      &86.48      &77.10      &85.21      &85.77        &91.89       &76.62       &79.96       &89.18       \\
\cellcolor{class-8}\color{white}8      &64.99       &65.96      &50.48      &79.49      &46.75      &65.39      &52.75        &40.27       &45.94       &57.05       &\textbf{83.79}       \\
\cellcolor{class-9}\color{white}9      &68.35       &80.59      &72.06      &\textbf{82.93}      &71.73      &77.37      &71.57        &77.29       &69.96       &68.27       &81.56       \\
\cellcolor{class-10}\color{white}10    &97.12       &55.29      &89.64      &59.90      &91.94      &98.84      &73.21        &94.08       &\textbf{98.76}       &91.70       &98.52       \\
\cellcolor{class-11}\color{white}11   &87.02       &92.24      &95.75      &82.85      &81.63      &96.89      &87.51        &91.75       &97.46       &89.14       &\textbf{99.91}       \\
\cellcolor{class-12}\color{white}12    &39.57       &3.67      &77.43      &83.31      &\textbf{90.59}      &82.82      &36.22        &56.25       &86.59       &70.64       &78.16       \\
\cellcolor{class-13}\color{white}13    &94.33       &\textbf{99.78}      &92.15      &91.06      &97.60      &95.42      &94.98        &90.84       &88.45       &94.98       &91.50       \\
\cellcolor{class-14}\color{white}14    &\textbf{100}       &\textbf{100}      &\textbf{100}      &\textbf{100}      &\textbf{100}      &\textbf{100}      &\textbf{100}        &\textbf{100}       &\textbf{100}       &\textbf{100}       &\textbf{100}       \\
\cellcolor{class-15}15     &\textbf{100}       &\textbf{100}      &\textbf{100}      &\textbf{100}      &99.53      &\textbf{100}      &\textbf{100}        &99.07       &\textbf{100}       &\textbf{100}       &\textbf{100}      \\ \hline
OA    &86.45       &82.63      &87.88      &86.74      &85.89      &89.56      &80.93        &81.05       &85.49       &84.73       &\textbf{92.31}       \\
AA    &88.81       &84.30      &89.28      &88.63      &88.55      &91.27      &85.94        &88.36       &90.30       &88.14       &\textbf{93.27}       \\
Kappa &0.853       &0.790      &0.869      &0.856      &0.847      &0.887      &0.816        &0.829       &0.847       &0.831       &\textbf{0.917}       \\ \hline
% Train (sec.)    &34.78     &73.95     &208.1    &30.32   &40.69   &73.09       &90.87      &128.1       &208.6       &715.6       &1313.2      \\
% Test (sec.)    &4.43     &6.36   &143.88    &5.11      &5.28     &7.96     &18.44     &12.12      &74.22      &102.6      &314.44   \\\hline
\end{tabular}
\end{table}

A closer examination of classification maps, especially in highlighted regions, confirms that HSIseg effectively reduces noise-like misclassifications and enhances boundary clarity, particularly in the IP, PU, and MG datasets. It maintains spatial coherence in uniform regions, significantly reducing isolated errors, and achieves sharp, well-aligned boundaries between land-cover types. These results support our hypothesis that larger patch sizes with spatial structural information improve both intra-object consistency and inter-object boundary delineation, demonstrating the potential of segmentation-based approaches for HSI classification when combined with multi-source data fusion.

\subsubsection{Comparison on T-SNE Visualization}
\begin{figure}[!t]
    \centering
    \includegraphics[scale=0.56]{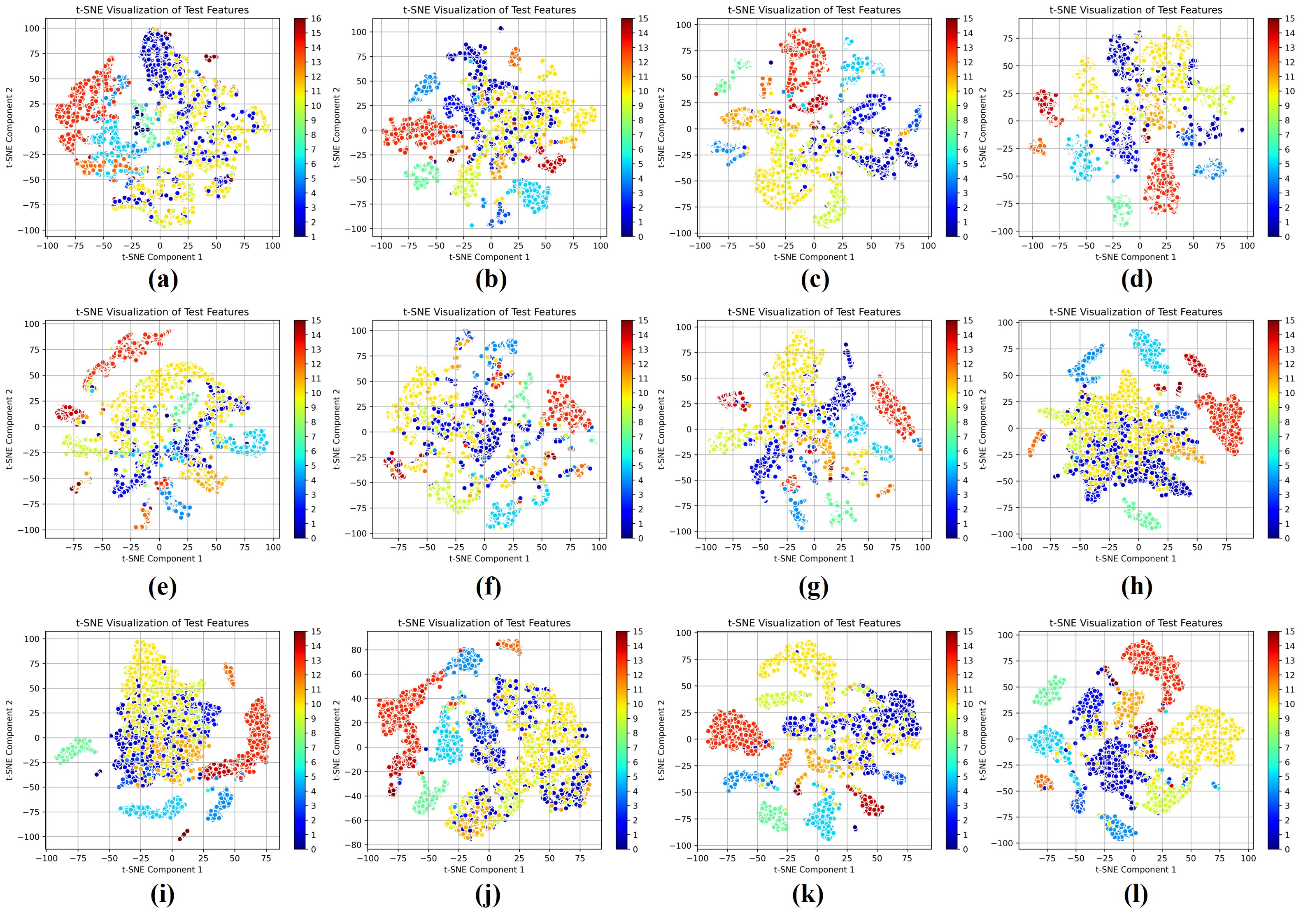}
    \caption{Comparison of T-SNE results on IP dataset by different methods. (a) T-SNE on original data. (b) SSFTT. (c) TransHSI. (d) MiM. (e) DFI . (f) MFT. (g) DeepSFT. (h) Unet. (i) TransUnet. (j) SwinUnet. (k) MambaUnet. (l) Ours. }
    \label{fig:IP TSNE}
\end{figure}
Deep learning-based methods excel in learning hierarchical feature representations, effectively transforming raw input data into more structured and task-relevant representations. These approaches typically train discriminative models to map input data into distinct feature spaces, which are subsequently leveraged for tasks such as HSI classification and clustering.

To evaluate the feature representational capabilities of different methods in this study, we employed t-distributed stochastic neighbor embedding (T-SNE) \cite{TSNE}, a widely used technique for visualizing high-dimensional data by projecting it into a lower-dimensional space while preserving local structure and cluster separability.

Fig.~\ref{fig:IP TSNE} illustrates the T-SNE visualization results for various methods using the IP dataset. The results demonstrate that our proposed model achieves superior feature discrimination compared to other approaches. The visualized feature clusters exhibit better intra-class cohesion and well-defined inter-class boundaries, indicating an improved ability to capture and preserve complex feature distributions. Notably, the clusters appear smoother and more compact, suggesting that our method effectively mitigates feature ambiguity and enhances separability, which is critical for accurate HSI classification.

\subsubsection{Comparison on Initial Patch size}
\begin{figure}[!t]
    \centering
    \includegraphics[scale=0.44]{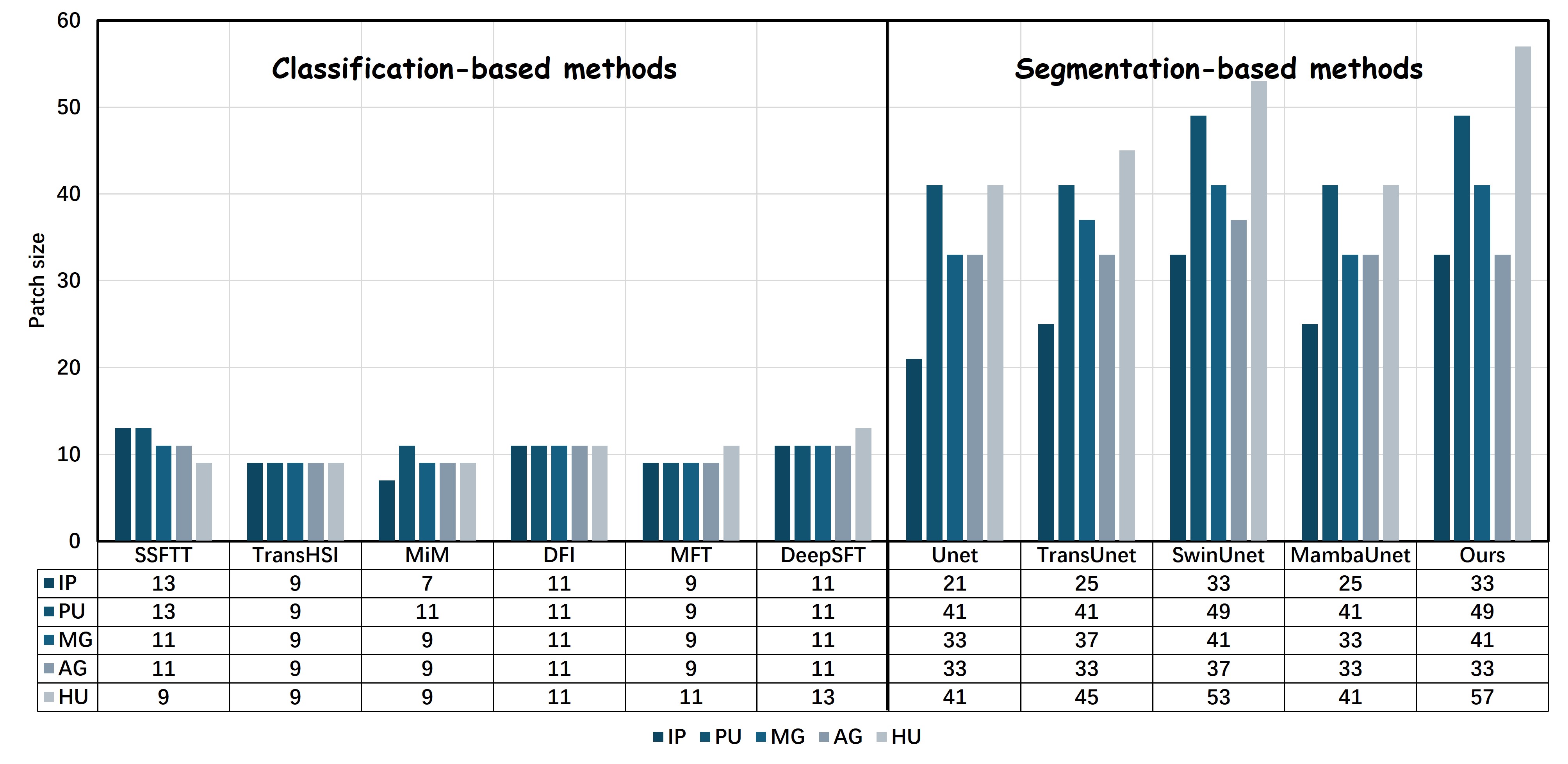}
    \caption{List of initial patch size used in different methods and different datasets.}
    \label{fig:PATCH SIZE COMPARISON}
\end{figure}
The primary distinction of this study lies in our pioneering use of a segmentation-based model for HSI classification. Unlike traditional classification methods that rely on small patch-wise inputs, segmentation models require significantly larger input sizes to effectively capture spatial context and structural information. The advantage of using larger patches is their ability to incorporate spatial patterns within HSI data, allowing the model to leverage both spectral and spatial contextual features. This is particularly beneficial for identifying structured objects and preserving spatial coherence in classification results.

To validate this segmentation-driven approach, we conduct experiments using substantially larger patch sizes compared to classification-based methods. Fig.~\ref{fig:PATCH SIZE COMPARISON} illustrates the differences in patch size selection across various methods and datasets.

The results confirm that segmentation-based models generally require larger patch sizes than classification-based approaches. Given the limited research on segmentation models for HSI classification, we investigate optimal patch sizes for these models. Our findings indicate that segmentation models, when assigned appropriately large patches, achieve comparable or superior performance to classification-based methods, as demonstrated by the quantitative results in performance tables.

Furthermore, classification map comparisons reveal that HSIseg, using larger patches, produces smoother intra-object classification in homogeneous land-cover regions. Additionally, boundaries between different land-cover classes are sharper and more precisely delineated, as observed in the magnified regions of the classification maps. This supports our hypothesis that larger patch sizes effectively preserve critical spatial structures, enhancing both object recognition and boundary refinement.

\subsubsection{Results on the \texorpdfstring{$\alpha$, $\beta$, $\gamma$, and $\delta$}{alpha, beta, gamma, and delta} for the loss balancing}
\begin{figure}[!t]
    \centering
    \includegraphics[scale=0.43]{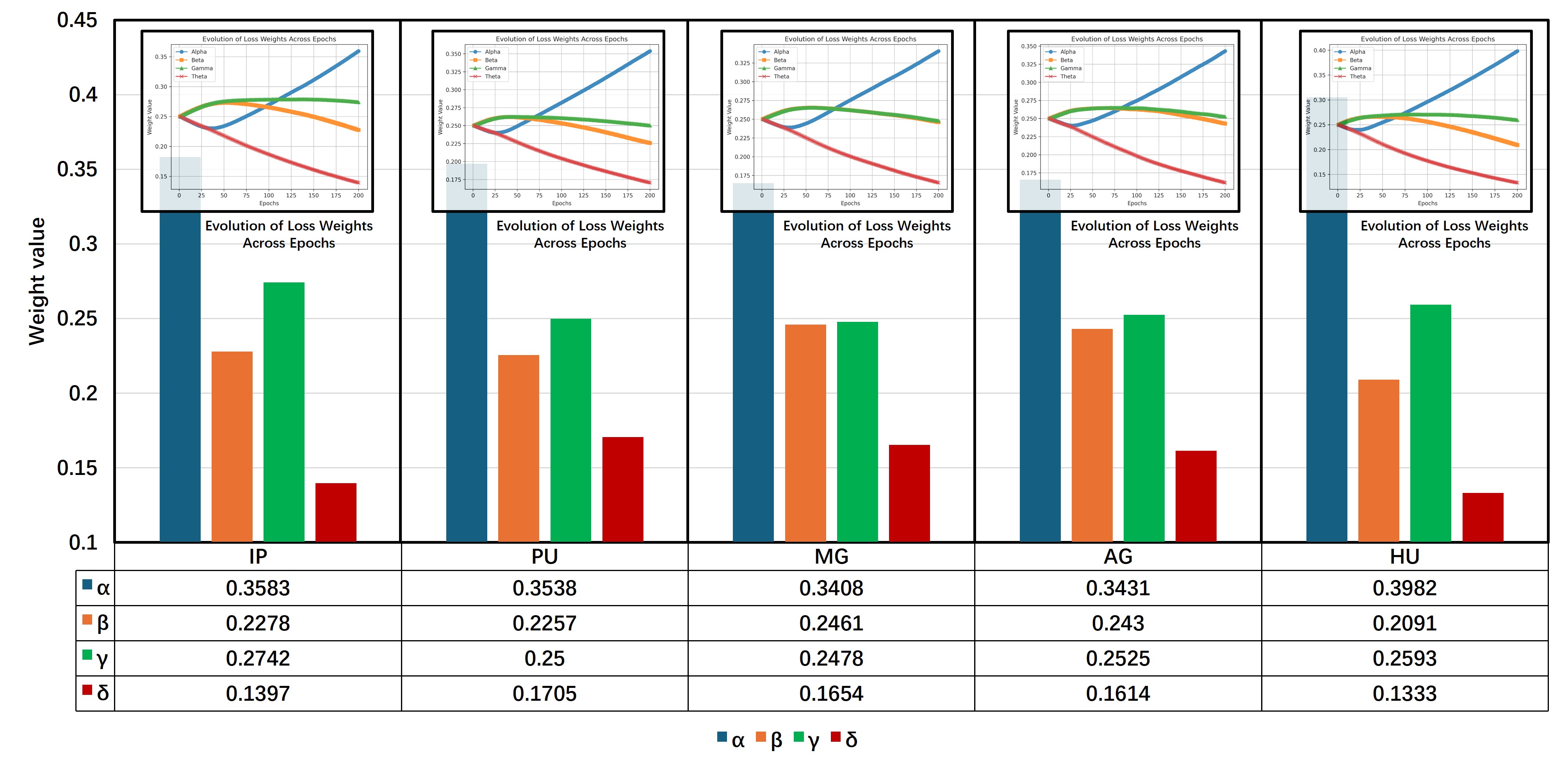}
    \caption{Investigation on the loss balancing weight and its evolution during training. }
    \label{fig:loss balancing}
\end{figure}
Since we introduce learnable weights \(\alpha, \beta, \gamma, \delta\) to balance the contributions of different loss components in multi-branch data collaboration, we record their values after completing the training process to analyze the influence of each loss term on model optimization.

Fig.~\ref{fig:loss balancing} presents the evolution of these weights across different datasets, highlighting their variations throughout the training process.

From the results, it is evident that the primary classification loss for HSI (\(\mathcal{L}_{hsi}^{cls}\)) consistently holds the highest weight (\(\alpha\)), demonstrating its dominant role in guiding model convergence. The second most influential component is the auxiliary data reconstruction loss (\(\mathcal{L}_{aux}^{rec}\)), with its associated weight \(\gamma\), indicating that auxiliary information contributes significantly to feature learning. Moreover, the Dice loss (\(\mathcal{L}_{hsi}^{dice}\)), introduced to enhance spatial structure alignment in segmentation, maintains a weight (\(\beta\)) comparable to \(\gamma\), reinforcing its importance in improving segmentation accuracy. Lastly, although we incorporate SSIM loss (\(\mathcal{L}_{aux}^{ssim}\)) to assess reconstruction quality, its weight (\(\delta\)) remains the lowest, suggesting that while perceptual consistency is relevant, its impact on the overall training process is relatively minor.

These findings confirm a reasonable and expected trend: the HSI classification task remains the primary objective, with auxiliary data acting as a supportive component to enhance the learning process.

\subsection{Ablation Studies}

\subsubsection{Effects on Initial Patch Size}
\begin{figure}[!t]
    \centering
    \includegraphics[scale=0.425]{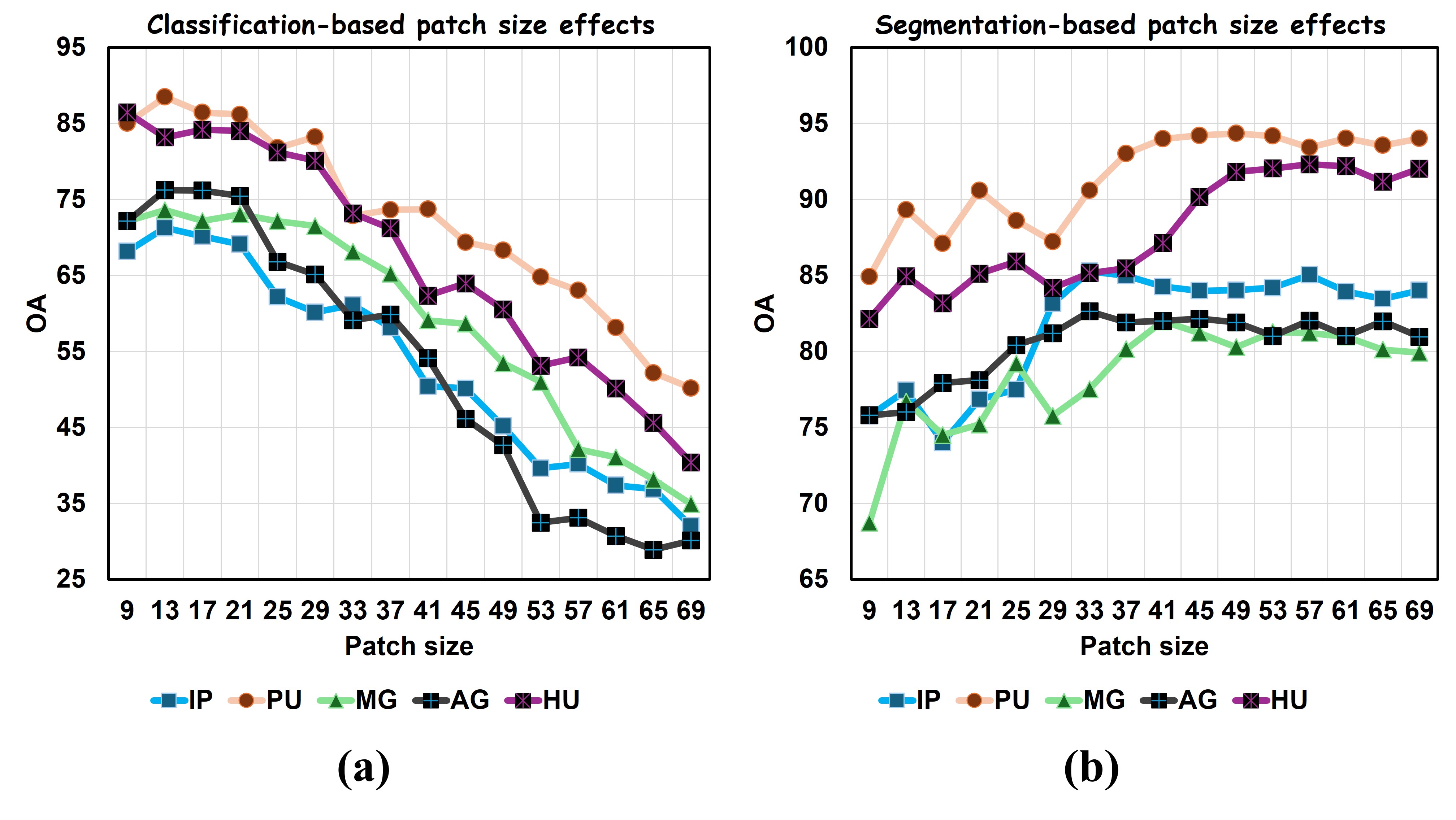}
    \caption{The comparison on the effects of initial patch size for the classification performance. (a) general classification-based model (SSFTT, etc.) where the performance tends to be degrading when patch size is increasing. (b) Our segmentation-based model, where performance tends to be upgrading until stable when patch size is increasing. }
    \label{fig:classification compare segmetation patch size}
\end{figure}
To determine the optimal initial patch size for segmentation-based methods, we conducted additional experiments to evaluate the influence of patch size on classification performance. We tested a range of patch sizes, varying from 9 to 69 with an interval of 4. The experimental results are illustrated in Fig.~\ref{fig:classification compare segmetation patch size}, where (a) shows the effects on classification-based methods (SSFTT) and (b) presents the results for our segmentation-based method.

Unlike classification-based methods, which often face challenges with larger patch sizes (e.g., the inclusion of more interfering pixels leading to degraded performance), segmentation-based methods appear to be less affected by larger patch sizes. As the patch size increases, the performance of segmentation-based methods tends to stabilize rather than decline. This is a significant finding, suggesting that segmentation-based approaches are more effective at mitigating the negative impact of interfering pixels, particularly near land-cover boundaries.

The saturation point, however, varies across datasets. We hypothesize that this variation depends on the ability to extract sufficient spatial contextual features at different scales present in the data. For instance, when comparing the PU, MG, and HU datasets, which represent urban scenes with distinct human-made objects, we observe noticeable differences in the saturation points due to variations in resolution and object size. This suggests that the saturation behavior is dataset-specific.

Unlike classification models that are constrained by the need to identify an optimal small patch size to avoid interference, segmentation-based models are less restricted by patch size due to their saturation behavior. Additionally, segmentation models provide fine-grained labels for each position within a patch, effectively mitigating the coarseness introduced by larger patch sizes. This capability makes segmentation-based methods particularly robust in scenarios involving varying spatial resolutions and complex boundary regions.

\subsubsection{Effects on Masking Ratio, \texorpdfstring{$\tau_1$}{tau1} and \texorpdfstring{$\tau_2$}{tau2}}
\begin{figure}[!t]
    \centering
    \includegraphics[scale=0.47]{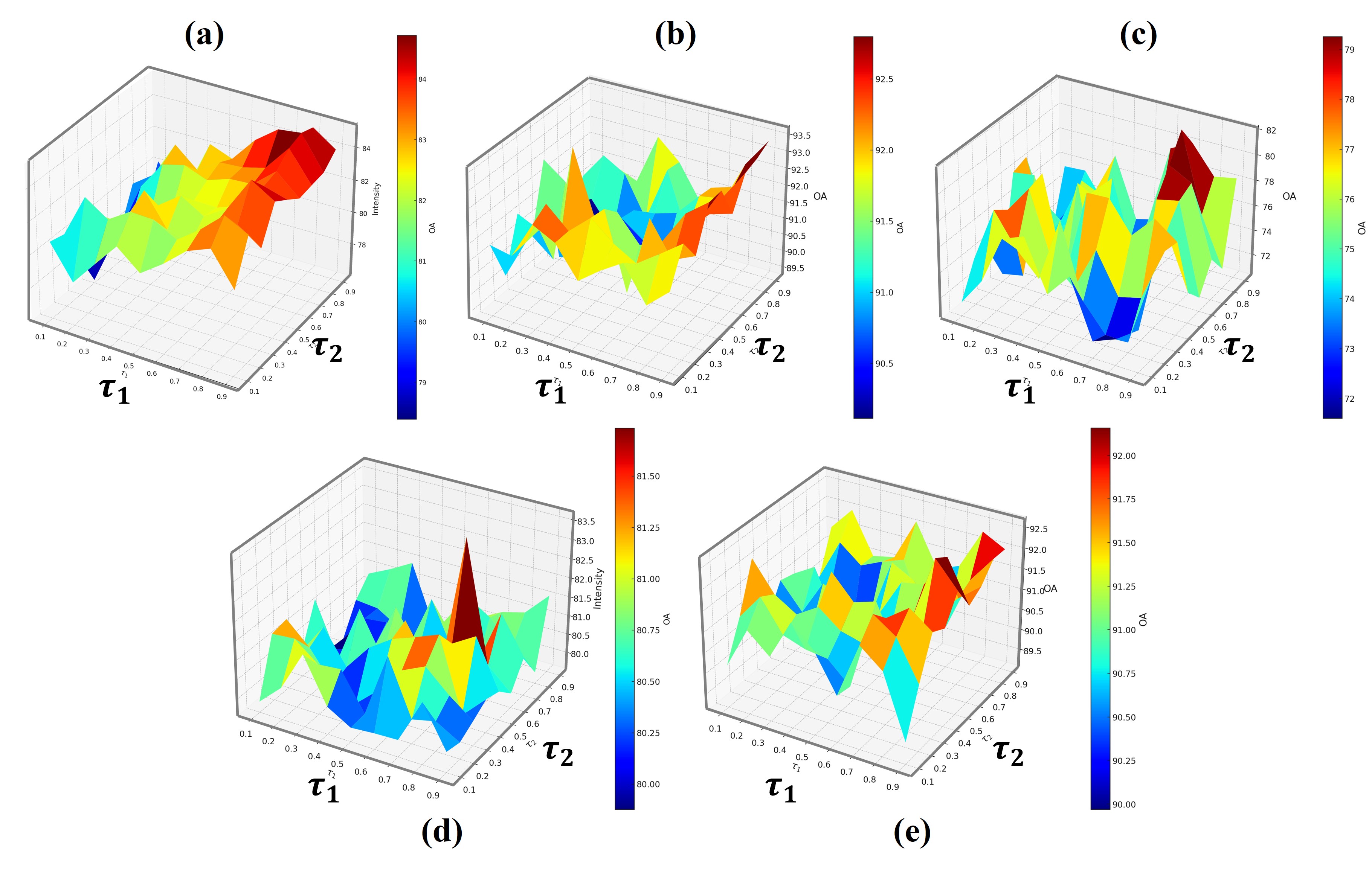}
    \caption{Investigation on the upper bound of masking value of $\tau_1$ and $\tau_2$ for different datasets. (a) IP ($\tau_1 = 0.8, \tau_2 = 0.7$). (b) PU ($\tau_1 = 0.9, \tau_2 = 0.9$). (c) MG ($\tau_1 = 0.8, \tau_2 = 0.7$). (d) AG ($\tau_1 = 0.6, \tau_2 = 0.7$). (e) HU ($\tau_1 = 0.9, \tau_2 = 0.8$).}
    \label{fig:tau1 and tau2}
\end{figure}
In this study, we introduce two masking upper bounds: $\tau_1$ in the DFS module and $\tau_2$ in the probabilistic pseudo-labeling process. The purpose of $\tau_1$ is to select a subset of discriminative positions with higher probability values, ensuring that the most informative features contribute to the decoding stage. Meanwhile, $\tau_2$ is used to select high-confidence classified pixels in the final inference stage. 

To determine the optimal thresholds, we conducted further experiments by varying $\tau_1$ and $\tau_2$ from 0.1 to 0.9. The experimental results, presented in Fig.~\ref{fig:tau1 and tau2}, illustrate the sensitivity of different datasets to these parameters.

From the results, we observe that the optimal values for $\tau_1$ and $\tau_2$ differ across datasets. For example, the PU and HU datasets achieve their best performance with higher thresholds ($\tau_1 = 0.9$, $\tau_2 = 0.9$ for PU and $\tau_1 = 0.9$, $\tau_2 = 0.8$ for HU), indicating that a more stringent selection of high-confidence regions is beneficial. Conversely, the AG dataset performs best with lower thresholds ($\tau_1 = 0.6$, $\tau_2 = 0.7$), suggesting that including a broader range of feature positions and a more flexible classification map contributes to better generalization.

However, it is difficult to establish a specific correlation between $\tau_1$ and $\tau_2$ across datasets. We hypothesize that this is due to the constraints imposed by the mean and standard deviation of probability distributions during the masking process. When $\tau_1$ and $\tau_2$ are set too low (e.g., 0.1 or 0.4), the difference in their effects becomes less significant, as they fall below the statistical threshold required for meaningful selection.

These findings emphasize the importance of dataset-specific tuning for the masking thresholds. Proper selection of $\tau_1$ and $\tau_2$ ensures a balance between selecting high-confidence features and maintaining sufficient diversity for effective feature learning.

% \subsubsection{Investigation on Tolerance $\zeta$ for Masking Threshold $\tau_2$}
% In the probabilistic masking in the progresive laearning, we design to make the masking threshold $\tau_2$ to be more discriminative for selection (e.g., higher is more discriminative), becaasue the model is fine-tuned step by step with further high-confidence on the pixels at each position. If the threshold $\tau_2$ is unchangable, it may shrink the model tranining somehow with those less discriminative pixels. Thus, we introdduce the tolerance value $\zeta$ to enlarge the $\tau_2$ gradually. 

% To evalute the effects on tolerance $\zeta$, we hvae conducted additional experiments on five datasets by changing the $\zeta$. The other parameters are unchanged especially the $\tau_2$. The experimental results are listed in Table~\ref{}

\subsubsection{Effects on Progressive Learning}
\begin{table}[!t]
\scriptsize
\caption{Investigation on the effects of number of iterations by the progressive learning with pseudo labeling. $\uparrow$ represents the OA improvements compared with the results from initial stage. }
\label{tab:iterations}
\centering
\begin{tabular}{c|cccccccccc}
\hline
     Dataset &\multicolumn{10}{c}{Number of iterations}    \\ \hline
 & 2 & 4 & 6 & 8 & 10 & 12 & 14 & 16 & 18 & 20  \\ \hline
IP &$1.87\uparrow$   &$2.01\uparrow$   &$2.14\uparrow$   &$2.20\uparrow$   &$2.31\uparrow$    &$2.53\uparrow$    &$\boldsymbol{2.96\uparrow}$    &$2.81\uparrow$    &$2.74\uparrow$    &$2.69\uparrow$        \\ \hline
PU &$1.91\uparrow$   &$2.21\uparrow$   &$2.41\uparrow$   &$\boldsymbol{2.52\uparrow}$   &$2.47\uparrow$    &$2.46\uparrow$    &$2.47\uparrow$    &$2.43\uparrow$    &$2.29\uparrow$    &$2.33\uparrow$        \\ \hline
MG &$1.81\uparrow$   &$2.21\uparrow$   &$2.58\uparrow$   &$3.01\uparrow$   &$\boldsymbol{3.15\uparrow}$    &$2.99\uparrow$    &$3.04\uparrow$    &$2.88\uparrow$    &$2.79\uparrow$    &$2.68\uparrow$        \\ \hline
AG &$1.52\uparrow$   &$1.68\uparrow$   &$2.01\uparrow$   &$2.29\uparrow$   &$\boldsymbol{2.35\uparrow}$    &$2.27\uparrow$    &$2.20\uparrow$    &$2.12\uparrow$    &$2.17\uparrow$    &$1.98\uparrow$        \\ \hline
HU &$1.80\uparrow$   &$2.10\uparrow$   &$2.27\uparrow$   &$\boldsymbol{2.53\uparrow}$   &$2.51\uparrow$    &$2.28\uparrow$    &$2.34\uparrow$    &$2.20\uparrow$    &$2.19\uparrow$    &$2.14\uparrow$        \\ \hline
\end{tabular}
\end{table}

\begin{figure}[!t]
    \centering
    \includegraphics[scale=0.42]{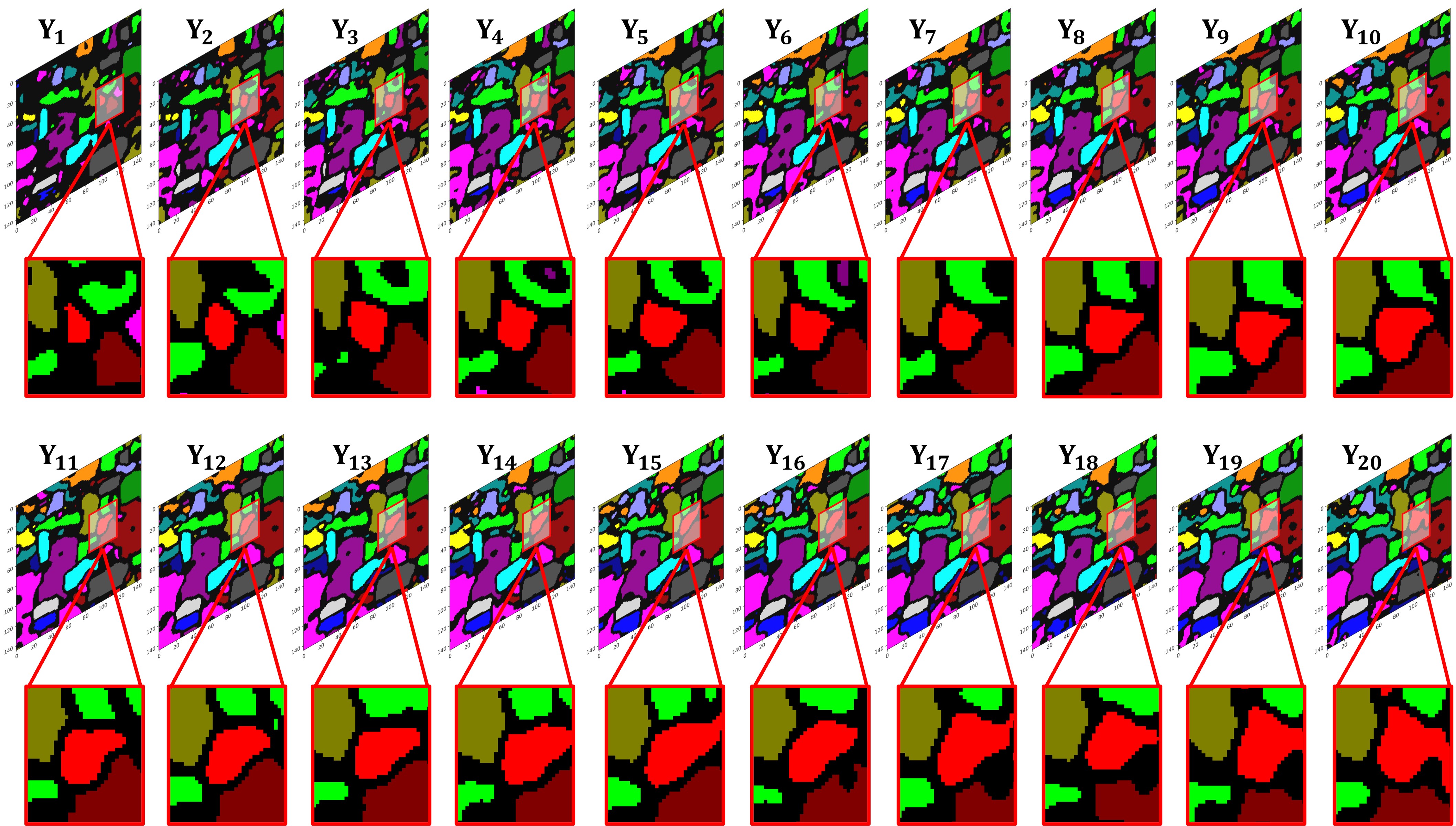}
    \caption{Example of progressive pseudo-labeling on the IP dataset with 20 iterations. By iteratively masking and labeling all pixels, including initially unknown regions, a temporal ground-truth is generated for the next training stage. As the iterations progress, the previously unknown areas are gradually labeled. Meanwhile, the differences between consecutive iterations progressively decrease, indicating stabilization of the pseudo-labeling process.}
    \label{fig:IP iterations}
\end{figure}
\begin{figure}[!t]
    \centering
    \includegraphics[scale=0.59]{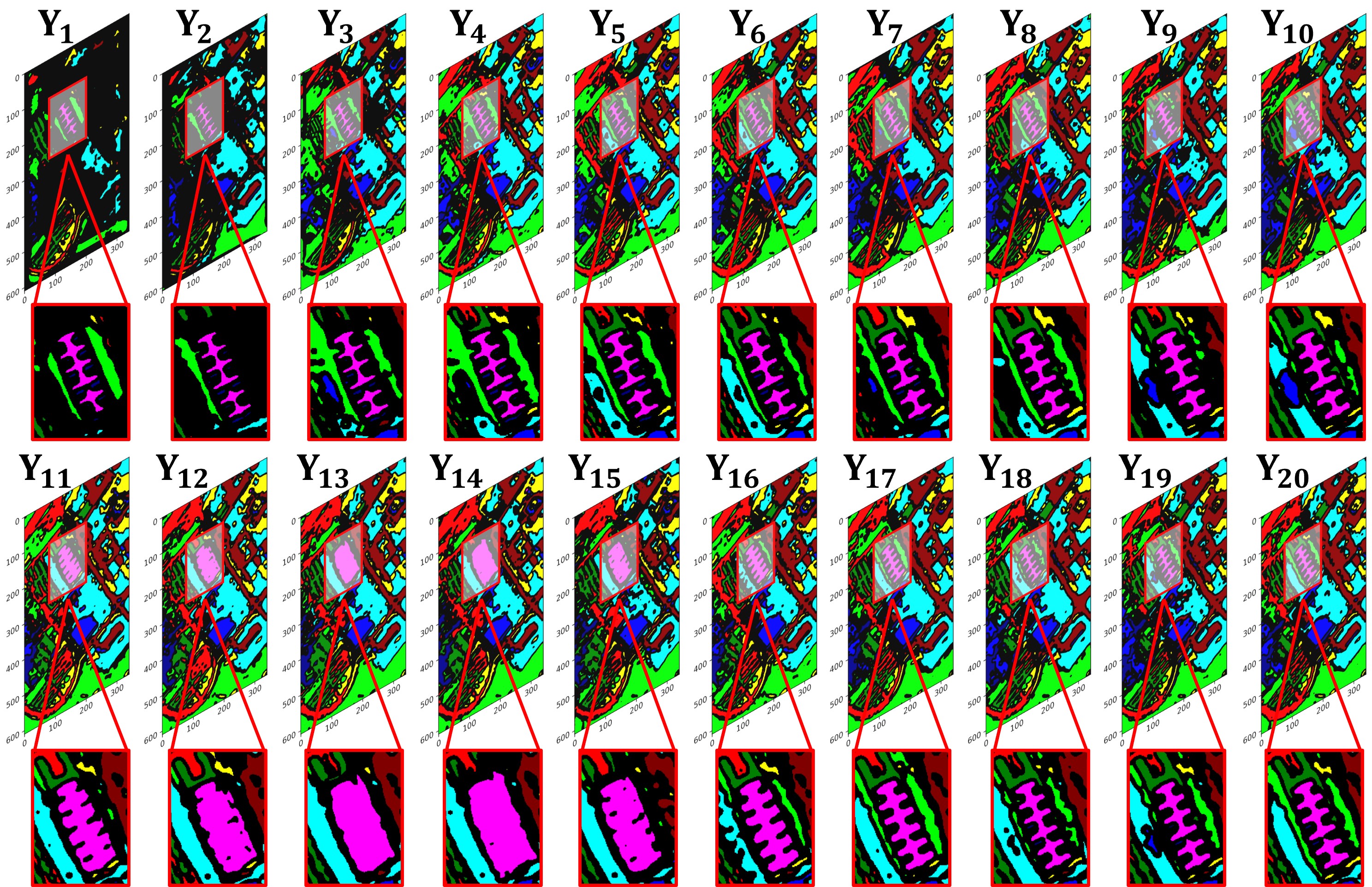}
    \caption{Example of progressive pseudo-labeling on the PU dataset with 20 iterations.}
    \label{fig:PU iterations}
\end{figure}
\begin{figure}[!t]
    \centering
    \includegraphics[scale=0.54]{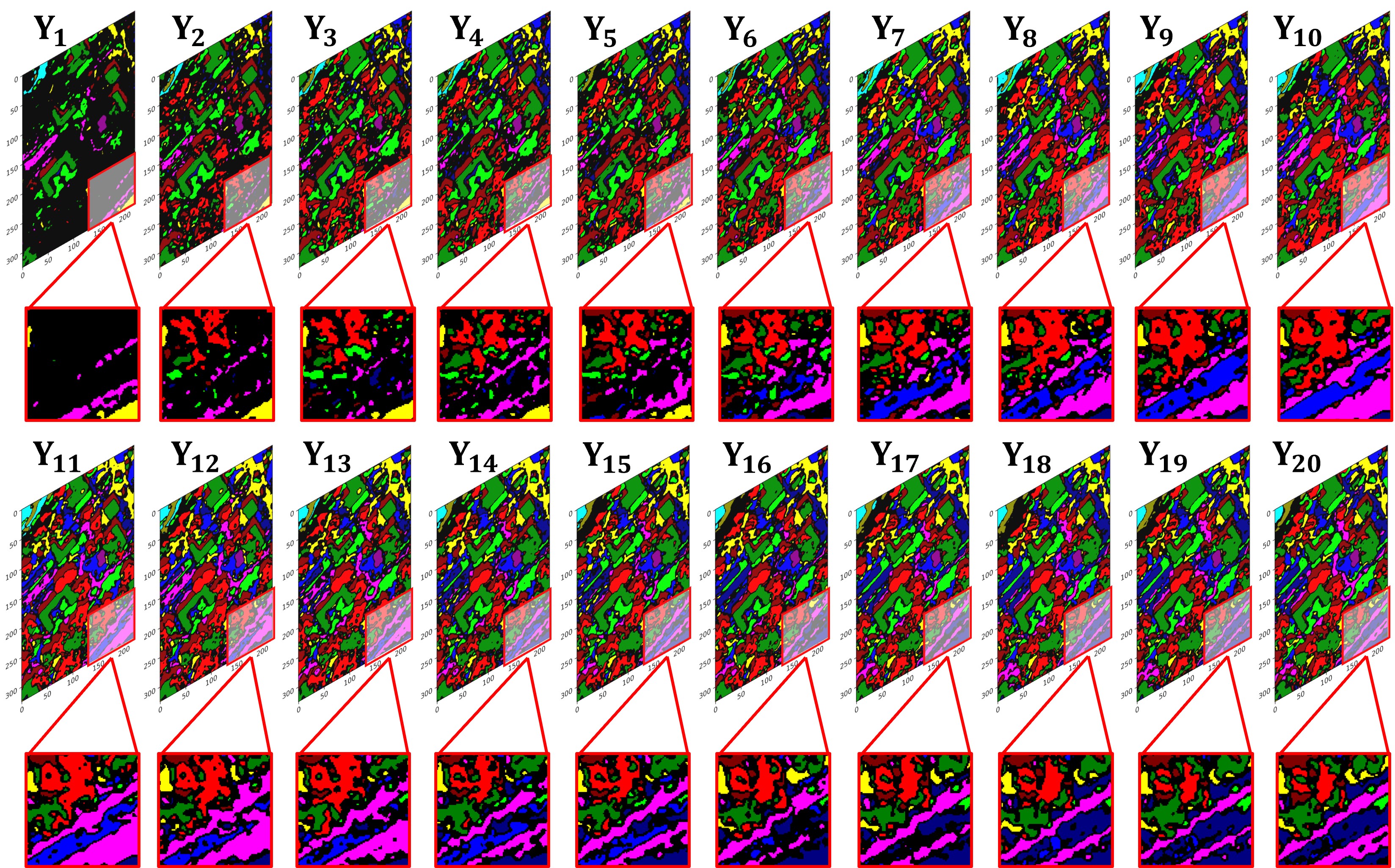}
    \caption{Example of progressive pseudo-labeling on the MG dataset with 20 iterations. }
    \label{fig:MG iterations}
\end{figure}
\begin{figure}[!t]
    \centering
    \includegraphics[scale=0.33]{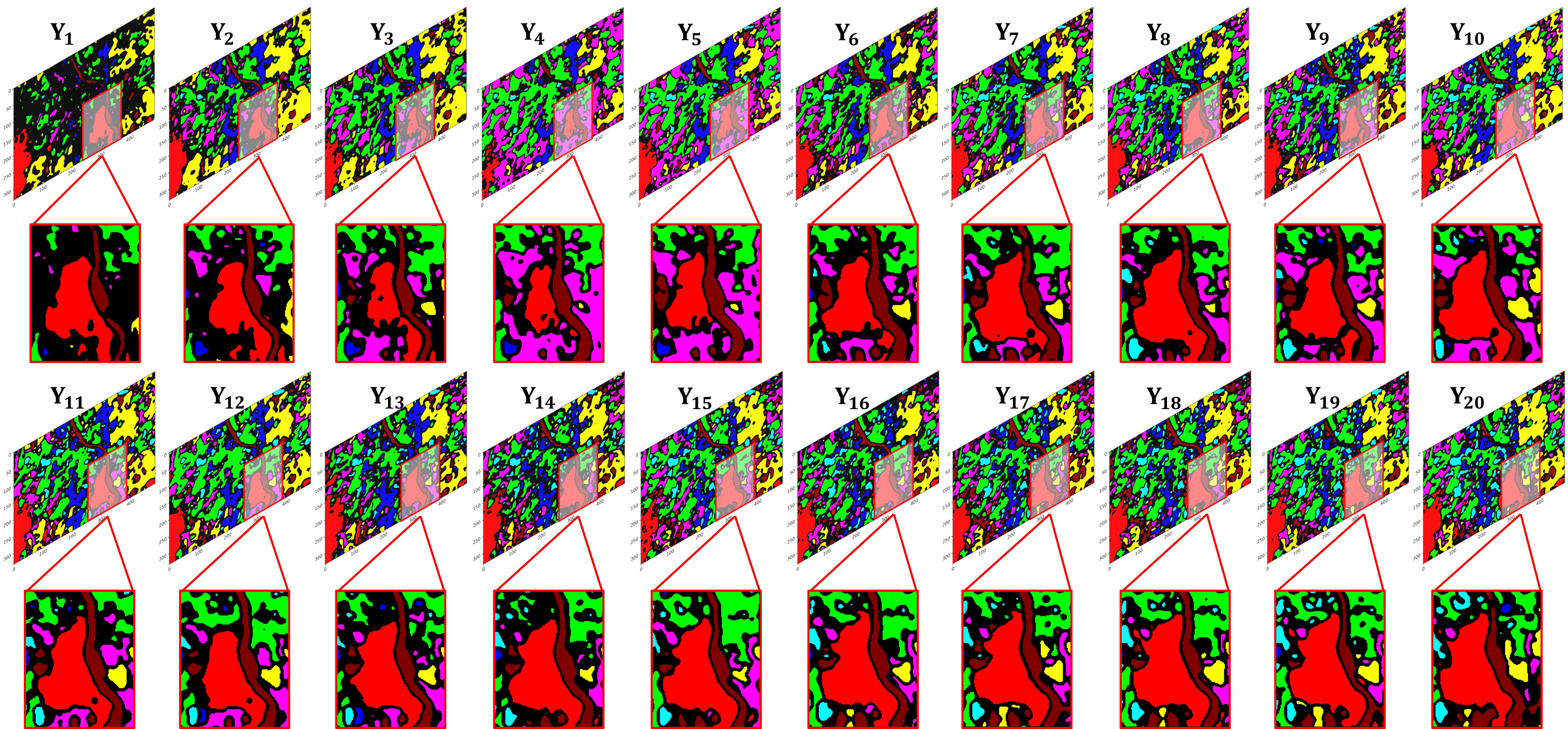}
    \caption{Example of progressive pseudo-labeling on the AG dataset with 20 iterations. }
    \label{fig:AG iterations}
\end{figure}
\begin{figure}[!t]
    \centering
    \includegraphics[scale=0.58]{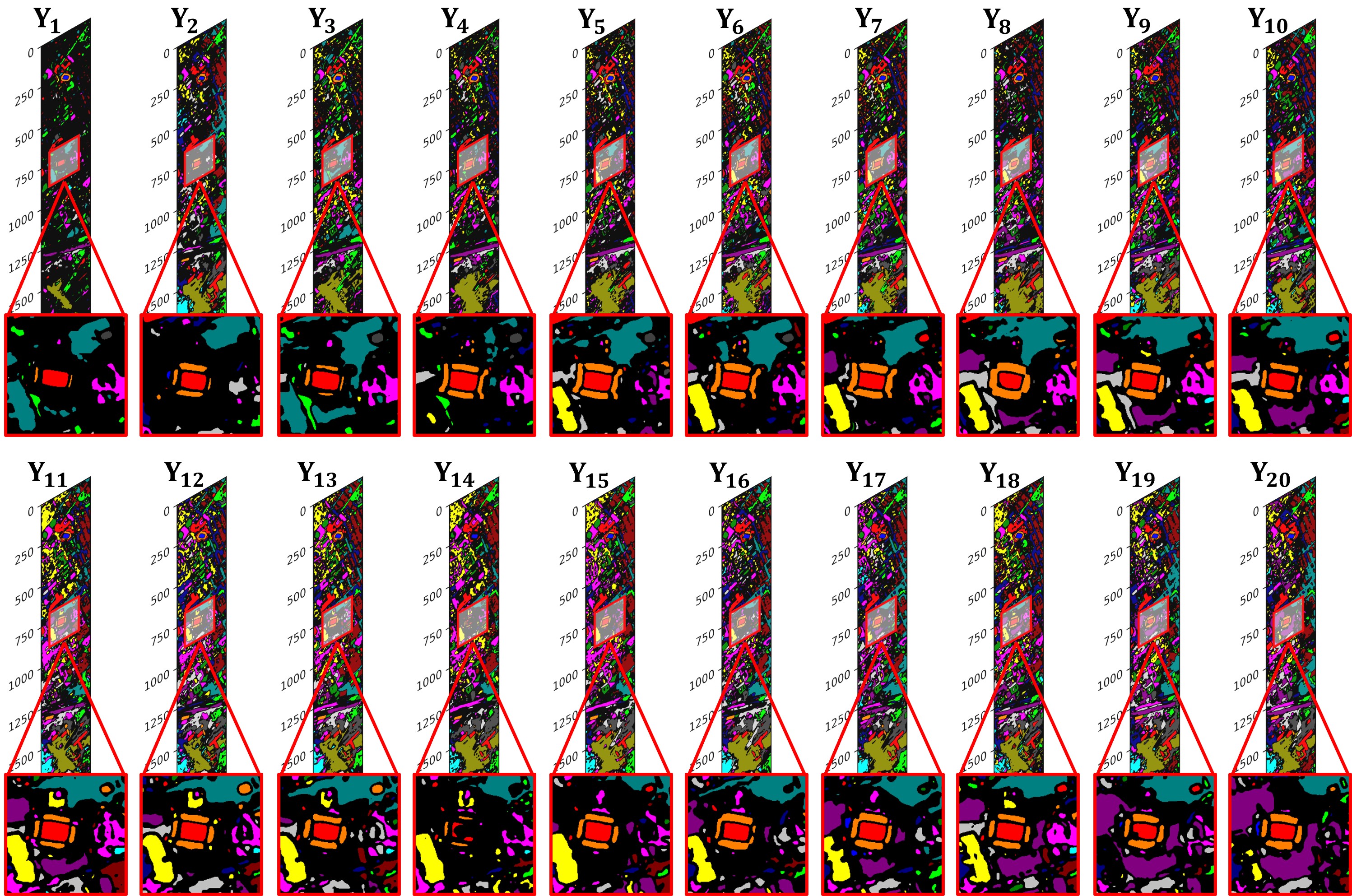}
    \caption{Example of progressive pseudo-labeling on the HU dataset with 20 iterations.}
    \label{fig:HU iterations}
\end{figure}
To validate the feasibility of the proposed progressive learning approach with iterative pseudo-labeling, we conducted experiments to determine the optimal number of iterations that enhance model performance. Specifically, we tested iteration numbers ranging from 2 to 30 at intervals of 2 (i.e., 2, 4, 6, 8, ..., 30). The experimental results are summarized in Table~\ref{tab:iterations}, while examples of pseudo-labeled temporal ground-truths are illustrated in Fig.~\ref{fig:IP iterations}, Fig.~\ref{fig:PU iterations}, Fig.~\ref{fig:MG iterations}, Fig.~\ref{fig:AG iterations}, and Fig.~\ref{fig:HU iterations}.

From the results, it is evident that increasing the number of iterations generally improves performance up to a saturation point, beyond which further iterations yield diminishing returns or even slight degradation in performance. For instance, the optimal iteration numbers were found to be 14, 8, 10, 10, and 8 for the IP, PU, MG, AG, and HU datasets, respectively. Beyond these values, additional iterations did not significantly enhance classification performance and, in some cases, led to performance degradation. This phenomenon can be attributed to the stabilization of the generated pseudo ground-truth, where differences gradually diminish as unknown regions become progressively labeled. When the model performs well in the early stages, the fine-tuning process refines the pseudo-labeling process, producing more reliable temporal ground-truth.

The visualization results further illustrate the progressive changes across iterations. As the number of iterations increases, previously unlabeled areas (represented as black regions) are progressively filled with high-confidence pseudo-labels. By actively generating pseudo-labels using discriminative maximum probabilities, our approach effectively expands the labeled dataset, providing more training samples for segmentation models. This strategy aligns with previous studies, such as \cite{seg_hsi}, which also addressed the challenge of labeling unknown regions. However, their approach relied on generating a complete simulated ground-truth dataset in advance, making it difficult to assess the reliability of the labeled areas. In contrast, our method incrementally labels discriminative regions in a stepwise manner, ensuring a more stable and adaptive pseudo-labeling process.

\begin{figure}[!t]
    \centering
    \includegraphics[scale=0.29]{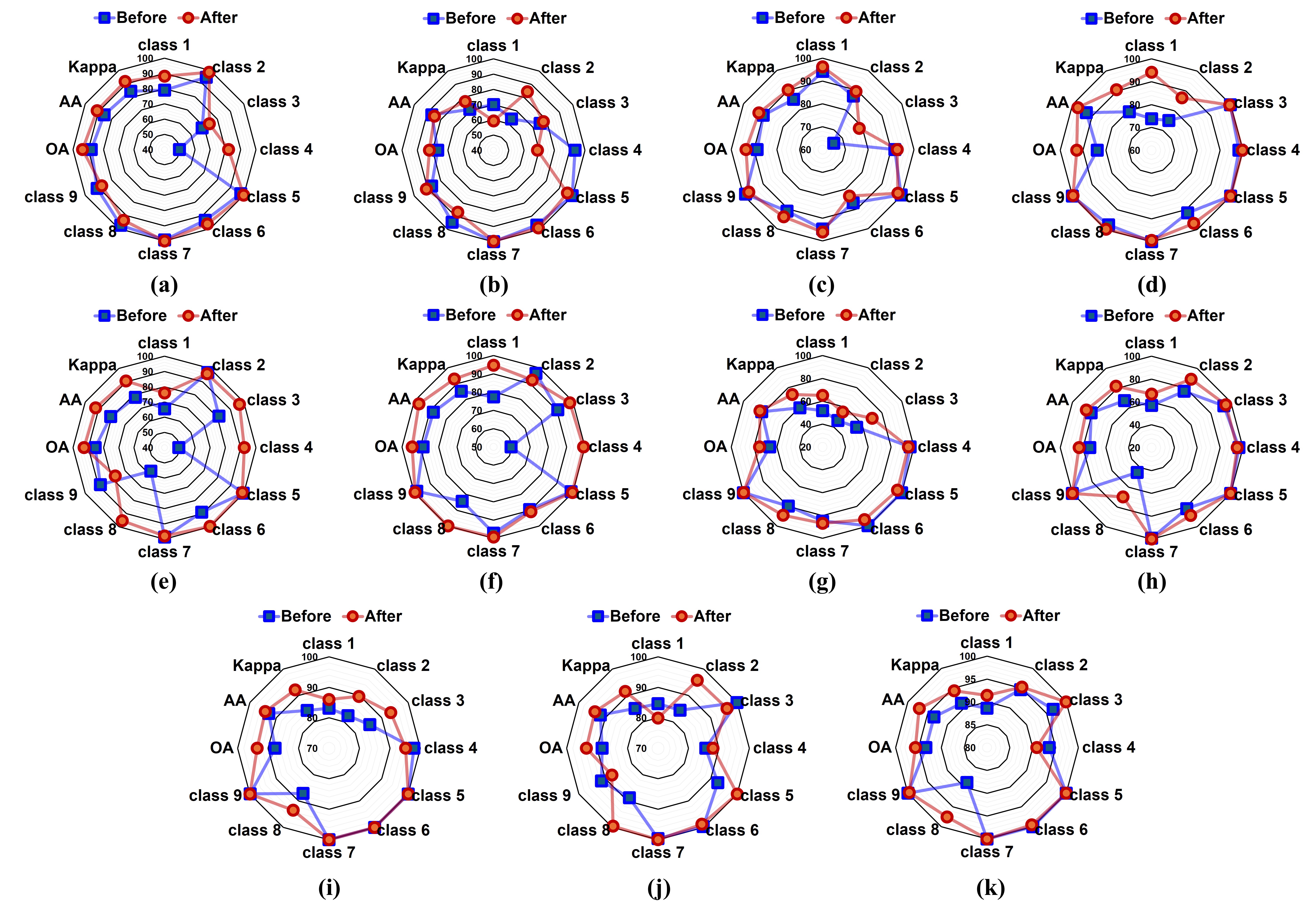}
    \caption{Example of the Effects of Progressive Learning on Different Methods for the PU Dataset: (a) SSFTT, (b) TransHSI, (c) MiM, (d) DFI, (e) MFT, (f) DeepSFT, (g) U-Net, (h) TransUnet, (i) SwinUnet, (j) MambaUnet, and (k) Ours.}
    \label{fig:effects of progressive learning}
\end{figure}

Additionally, to verify the adaptability of our progressive learning framework to other methods, we conducted further experiments on all compared approaches to evaluate their performance evolution. In these experiments, we controlled key variables by fixing the masking value $\tau_2$, tolerance $\zeta$, and the number of iterations across all methods. Taking the PU dataset as an example, we set $\tau_2 = 0.8$ and $\zeta = 0.005$ with 20 iterations. During the iterative process, $\tau_2$ gradually increases to 0.9. The results are illustrated in Fig.~\ref{fig:effects of progressive learning}. The results indicate that methods incorporating progressive learning generally achieve better performance compared to those without it. While some methods show significant improvements, others exhibit more modest gains. Interestingly, in certain cases, not all class-wise accuracies improved—some classes even experienced a decline in performance. We hypothesize that this is due to the selection of misclassified samples with high confidence in earlier iterations, which may introduce noise into the training process. Nevertheless, the overall performance trend demonstrates consistent improvement with the application of progressive learning.

In conclusion, the results validate that our proposed progressive learning framework with iterative pseudo-labeling is an effective strategy for HSI classification. By gradually incorporating pseudo-labeled areas into the training process, our method facilitates better utilization of segmentation models, particularly in scenarios where large portions of the dataset remain unlabeled.

\subsubsection{Effects on Multi-Source Data Collaboration}
To evaluate the impact of different modal inputs on classification performance, we conducted three sets of experiments: (1) using only HSI as input, (2) using only auxiliary data (LiDAR, SAR, or MSI), and (3) combining HSI with auxiliary data. The experimental results, summarized in Table~\ref{tab:multi-source}, demonstrate that multi-source inputs consistently outperform single-modality inputs across all five datasets in terms of OA, AA, and Kappa metrics. 

A notable aspect of this study is the use of simulated LiDAR data for the IP and PU datasets. Specifically, the simulated LiDAR images for IP and PU were generated using ChatGPT, which was prompted with the names of land-cover categories and the original ground-truth map. By leveraging its contextual knowledge of land-cover objects' typical heights, ChatGPT produced simulated LiDAR images corresponding to specific locations. Experimental results indicate that incorporating these contextual features—such as land-cover types and general height information—into the model training process improves classification accuracy.

However, using simulated LiDAR as the sole input resulted in significantly lower performance compared to datasets with real auxiliary data. This degradation is particularly evident when compared to datasets incorporating authentic LiDAR or other auxiliary modalities. We hypothesize that this decline is due to the limitations of the simulated LiDAR, which provides values only for originally labeled areas while neglecting unknown regions. The absence of comprehensive spatial coverage likely diminishes its effectiveness as a standalone input.

\begin{table}[!t]
\scriptsize
\caption{Investigation on effects of multi-source data collaboration}
\label{tab:multi-source}
\centering
\begin{tabular}{c|c|c|c|c|c|c}
\hline
                      &        & IP & PU & MG & AG & HU \\ \hline
\multirow{3}{*}{HSI only} & OA (\%)     &$84.42\pm1.51$    &$92.34\pm1.23$    &$79.21\pm2.01$    &$80.22\pm2.91$    &$91.07\pm1.01$    \\
                      & AA (\%)     &$87.15\pm1.19$    &$92.12\pm1.01$    &$82.14\pm1.59$    &$71.09\pm2.37$    &$92.41\pm0.48$    \\
                      & Kappa  &$0.812\pm0.04$    &$0.903\pm0.01$    &$0.721\pm0.02$    &$0.745\pm0.03$    &$0.904\pm0.02$    \\ \hline\hline
\multirow{3}{*}{Auxiliary data only}  & OA (\%)     &$53.05\pm8.12$    &$74.15\pm4.92$    &$75.45\pm0.68$    &$61.29\pm4.03$    &$83.12\pm1.64$    \\
                      & AA (\%)                   &$51.08\pm7.29$    &$71.15\pm3.43$    &$72.17\pm1.36$    &$56.81\pm3.91$    &$81.95\pm1.20$    \\
                      & Kappa                     &$0.478\pm0.16$    &$0.698\pm0.12$    &$0.699+0.14$    &$0.573\pm0.17$    &$0.804\pm0.08$    \\ \hline\hline
\multirow{3}{*}{HSI + Auxiliary data}  & OA (\%)     &$\boldsymbol{85.28\pm1.21}$    &$\boldsymbol{93.37\pm1.03}$    &$\boldsymbol{81.99\pm1.71}$    &$\boldsymbol{82.64\pm2.07}$    &$\boldsymbol{92.31\pm0.89}$    \\
                      & AA (\%)     &$\boldsymbol{90.02\pm0.87}$    &$\boldsymbol{93.40\pm0.28}$    &$\boldsymbol{85.99\pm0.41}$    &$\boldsymbol{74.86\pm1.13}$   &$\boldsymbol{93.27\pm0.30}$   \\
                      & Kappa  &$\boldsymbol{0.833\pm0.02}$    &$\boldsymbol{0.911\pm0.01}$    &$\boldsymbol{0.777\pm0.02}$    &$\boldsymbol{0.787\pm0.01}$    &$\boldsymbol{0.917\pm0.01}$    \\ \hline
\end{tabular}
\end{table}

\subsubsection{Effects on Feature Interaction in Decoding Stage}
Some classification-based studies \cite{deepsft}\cite{hsix1}\cite{hsix2} perform multi-source data fusion at the encoding stage, directly feeding the fused features into a decoder—typically a fully connected layer—for classification. In these approaches, feature collaboration in the decoding stage is often limited to simple summation, lacking sophisticated interactions. However, in segmentation-based models, the decoding stage follows an up-sampling process, rendering such straightforward fusion methods insufficient.

To address this limitation and extend multi-source feature fusion into the decoding stage, we draw inspiration from feature interaction techniques in medical image segmentation \cite{xnet1}\cite{xnet2}. We propose an iterative feature collaboration mechanism at each decoder stage, where features are fused via the CFI module before being processed by individual branches in subsequent decoding layers. Unlike classification-based methods that primarily focus on feature fusion at the encoding stage, our approach integrates feature interaction within the decoding stage.

\begin{figure}[!t]
    \centering
    \includegraphics[scale=0.43]{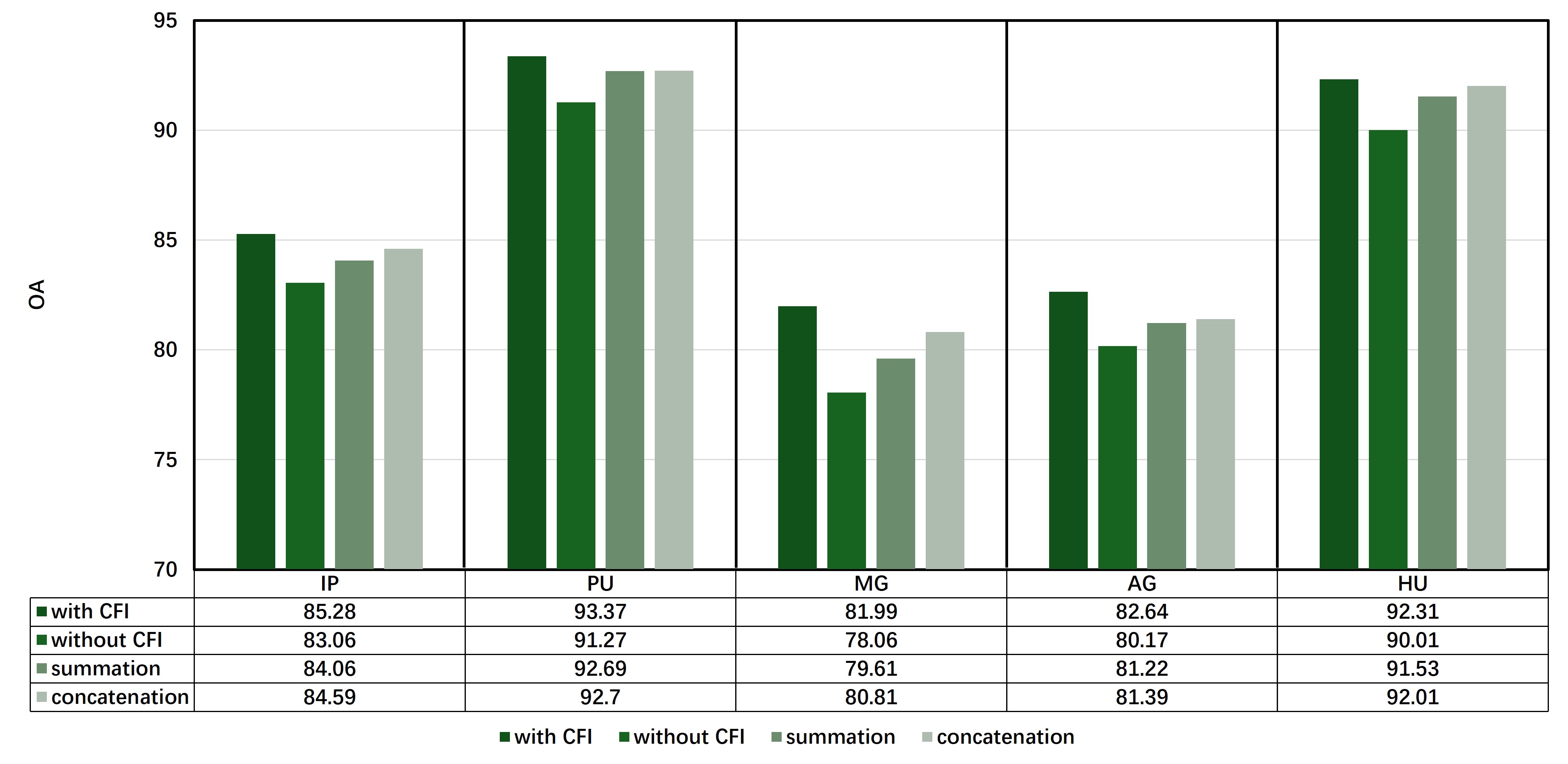}
    \caption{Effects of multi-source features interaction in the decoding stage.}
    \label{fig:decoding interaction}
\end{figure}

Fig.~\ref{fig:decoding interaction} presents ablation results illustrating the impact of employing CFI for feature interaction in the decoding stage within our segmentation framework. The results clearly indicate that incorporating feature interaction at this stage consistently improves HSI classification performance. Additionally, we extended our experiments to include simple fusion strategies, such as summation and concatenation of multi-branch features in the decoding stage. While these approaches also showed improvements, they failed to account for the correlations between features from different branches, resulting in inferior performance compared to the CFI-based approach.

\section{Further Discussion on Limitations and Future Research\label{discussion}}

While our proposed model achieves the highest classification accuracy across all five datasets, it is important to acknowledge its longer processing time. We attribute this computational cost primarily to the increased linear complexity introduced by the query-driven dynamic regional mechanism in the DSRT module. As the number of tokens in a patchified layer grows, the computational burden escalates, leading to prolonged training and inference times. Another contributing factor may be the limitations of our experimental hardware, which could have impacted the overall efficiency of the model during both training and testing phases.

To validate this assumption, we conducted an ablation study in which we replaced DSRT with a standard ViT model while keeping all other parameters unchanged. The results, summarized in Table~\ref{tab:DSRT and ViT}, include metrics such as OA, AA, Kappa, training time, and testing time. The findings confirm that substituting DSRT with a standard ViT significantly reduces computational overhead. However, this reduction in processing time comes at the cost of a slight decline in classification accuracy, though it remains higher than that of the compared methods. These results reinforce our hypothesis that the query-driven regional generation process in DSRT contributes substantially to computational complexity.

Since this study introduces a novel segmentation-based paradigm for HSI classification, the overall framework has been designed to be modular and extendable. This flexibility provides opportunities for future research to further refine and optimize both the encoder and decoder structures. For example, adopting a standard ViT-based architecture as the base encoder or decoder could serve as a more computationally efficient alternative, simplifying implementation while maintaining better performance than the compared state-of-the-art methods.

In conclusion, while DSRT significantly enhances classification performance, validating the feasibility of our proposed segmentation-based approach, its efficiency remains a challenge. Future research will focus on introducing sparsity into DSRT module and its selective query-driven mechanism to reduce computational cost while preserving its strong classification capabilities.

\begin{table}[!t]
\scriptsize
\caption{Investigation on proposed DSRT and general ViT for performance evaluation}
\label{tab:DSRT and ViT}
\centering
\begin{tabular}{c|c|c|c|c|c|c}
\hline
                      &               & IP & PU & MG & AG & HU \\ \hline
\multirow{6}{*}{DRST} & OA (\%)       &$85.28\pm1.21$    &$93.37\pm1.03$    &$81.99\pm1.71$          
                      &$82.64\pm2.07$    &$92.31\pm0.89$    \\

                      & AA (\%)       &$90.02\pm0.8$    &$93.40\pm0.28$    &$85.99\pm0.41$    &$74.86\pm1.13$     &$93.27\pm0.30$    \\
                      
                      & Kappa         &$0.833\pm0.02$    &$0.911\pm0.01$    &$0.777\pm0.02$    &$0.787\pm0.01$    &$0.917\pm0.01$     \\
                      
                      & Train (sec.)  &1221.1    &1346.2    &1302.6    &1176.2    &1313.2    \\
                      & Test (sec.)   &155.19    &273.51    &253.48    &344.56    &314.44    \\\hline\hline
                      
\multirow{6}{*}{Original ViT}  & OA (\%)       &$84.09\pm1.39$    &$92.58\pm0.14$     &$80.11\pm1.32$    &$81.38\pm1.27$    &$91.44\pm0.29$    \\
                      & AA (\%)       &$88.55\pm1.20$    &$92.31\pm0.82$     &$83.42\pm0.37$    &$73.04\pm0.93$    &$92.81\pm0.23$    \\
                      & Kappa         &$0.801\pm0.03$    &$0.902\pm0.01$     &$0.752\pm0.02$    &$0.732\pm0.02$    &$0.909\pm0.01$    \\
                      
                      & Train (sec.)  &203.15    &218.88    &195.69    &213.54    &209.17    \\
                      & Test (sec.)   &15.92    &22.16    &18.12    &35.18    &30.99    \\\hline
\end{tabular}
\end{table}

\section{Conclusion\label{conclusion}}

In this study, we introduce a novel paradigm and baseline for HSI classification by leveraging segmentation models as the foundational framework. Unlike traditional classification-based approaches that rely on small patch-wise learning, our method incorporates large-scale segmentation, enabling both single-source HSI classification and multi-source data collaboration for enhanced performance. Our segmentation model builds upon the U-Net architecture, integrating two key innovations: the DFS module and the CFI module. These enhancements ensure more effective feature alignment and interaction. Additionally, the encoder and decoder employ CNNs in conjunction with the proposed DSRT. DSRT introduces query-driven dynamic shifted regions, allowing the model to adaptively capture local-to-global dependencies for improved representation learning. To further enhance model generalization, we propose a progressive pseudo-labeling strategy. This strategy iteratively generates temporal pseudo ground-truth, progressively incorporating originally unlabeled pixels into the training process. By dynamically refining pseudo-labels, our method mitigates the challenges associated with limited labeled samples and improves overall classification accuracy. Extensive experiments conducted on five benchmark datasets validate the effectiveness and feasibility of our approach, achieving state-of-the-art results. Moreover, our findings demonstrate that segmentation-based models, when properly adapted, can overcome the limitations of small patch sizes in HSI classification, providing a promising direction for future research in this field.

\section{Acknowledgements}
M.S. Wong thanks the funding support from the General Research Fund (Grant No. 15603923 and 15609421), and the Collaborative Research Fund (Grant No. C5062-21GF) from the Research Grants Council, Hong Kong, China; and the funding support from the Research Institute for Sustainable Urban Development, The Hong Kong Polytechnic University, Hong Kong, China (Grant No. 1-BBG2).

\end{document}